%% file: main.tex
\documentclass[utf8]{FrontiersinHarvard}

\usepackage{fix-cm}
\usepackage{url}
\usepackage[hyperfootnotes=false]{hyperref}
\usepackage[onehalfspacing]{setspace}
\usepackage{framed,multirow}
\usepackage{booktabs}
\usepackage{longtable}
\usepackage{amssymb}
\usepackage{amsmath}
\usepackage{latexsym}
\usepackage{bm}
\usepackage[export]{adjustbox}
\usepackage{makecell}
\usepackage{placeins}
\usepackage{rotating}
\usepackage{tikz}

\input{revision_commands}
\ifdefined\showrevision
\makeatletter
\renewcommand{\revref}[2]{%
    \begingroup
    \@twosidefalse
    \normalmarginpar
    \marginnote{$R_{#1}C_{#2}$}%
    \endgroup
}
\renewcommand{\secondrevref}[2]{%
    \begingroup
    \@twosidefalse
    \normalmarginpar
    \marginnote{$R_{#1}C_{#2}$}[0.3cm]%
    \endgroup
}
\makeatother
\fi

\graphicspath{{figures/}}
\def\keyFont{\fontsize{8}{11}\helveticabold}
\def\firstAuthorLast{Wang {et~al.}}
\def\Authors{Junwen Wang\,$^{1,*}$, Oscar MacCormac\,$^{1,2}$,\ William Rochford\,$^{1,2}$, Aaron Kujawa\,$^{1}$,\ Jonathan Shapey\,$^{1,2}$ and Tom Vercauteren\,$^{1}$}
\def\Address{$^{1}$School of Biomedical Engineering \& Imaging Sciences,\\ King's College London, London, United Kingdom\\
$^{2}$Department of Neurosurgery, King's College Hospital,\\ London, United Kingdom}
\def\corrAuthor{Junwen Wang}
\def\corrEmail{junwen.wang@kcl.ac.uk}

\definecolor{newcolor}{rgb}{.8,.349,.1}

\newcommand{\figref}[1]{Figure~\ref{#1}}
\newcommand{\tabref}[1]{Table~\ref{#1}}
\newcommand{\secref}[1]{Section~\ref{#1}}

\DeclareMathOperator{\TPR}{TPR}

\DeclareMathOperator{\Dice}{Dice}
\DeclareMathOperator{\NSD}{NSD}
\DeclareMathOperator{\SDice}{Dice_{small}}
\DeclareMathOperator{\SNSD}{NSD_{small}}

\DeclareMathOperator{\BACC}{BACC}
\DeclareMathOperator{\Fone}{F1}

\DeclareMathOperator{\loss}{\mathcal{L}}

\begin{document}
\renewcommand{\thefootnote}{\fnsymbol{footnote}}
\onecolumn
\firstpage{1}

\title[Tree-Based Semantic Losses]{Label tree semantic losses for rich\\ multi-class medical image segmentation}

\author[\firstAuthorLast]{\Authors}
\address{}
\correspondance{}

\extraAuth{}

\maketitle

\begin{abstract}\sloppy\noindent
Rich and accurate medical image segmentation is poised to underpin the next generation of AI-defined clinical practice by delineating critical anatomy for pre-operative planning, guiding real-time intra-operative navigation, and supporting precise post-operative assessment. However, commonly used learning methods for medical and surgical imaging segmentation tasks penalise all errors equivalently and thus fail to exploit any inter-class semantics in the label space. This becomes particularly problematic as the cardinality and richness of labels increases to include subtly different classes. In this work, we propose two tree-based semantic loss functions which take advantage of a hierarchical organisation of the labels. We further incorporate our losses in a recently proposed approach for training with sparse, background-free annotations to extend the applicability of our proposed losses. Extensive experiments are reported on two medical and surgical imaging segmentation tasks, namely head MRI for whole brain parcellation with full supervision and neurosurgical hyperspectral imaging for scene understanding with sparse annotations. \revdel{Results demonstrate that our proposed method reaches state-of-the-art performance in both cases.}\revmod{Results demonstrate consistent improvements over the evaluated task-specific baselines, with the strongest support for the Wasserstein-based compound loss in whole-brain parcellation and for hierarchy-weighted top-level supervision in the sparse HSI setting.}\revref{2}{3}\secondrevref{1}{4}

\tiny
\keyFont{\section{Keywords:} semantic segmentation, hyperspectral imaging, label hierarchy, sparse annotations, whole brain parcellation}
\end{abstract}

\section{Introduction}\label{sec:introduction}
\begin{figure}[tbp]
    \centering
    \includegraphics[width=0.98\linewidth]{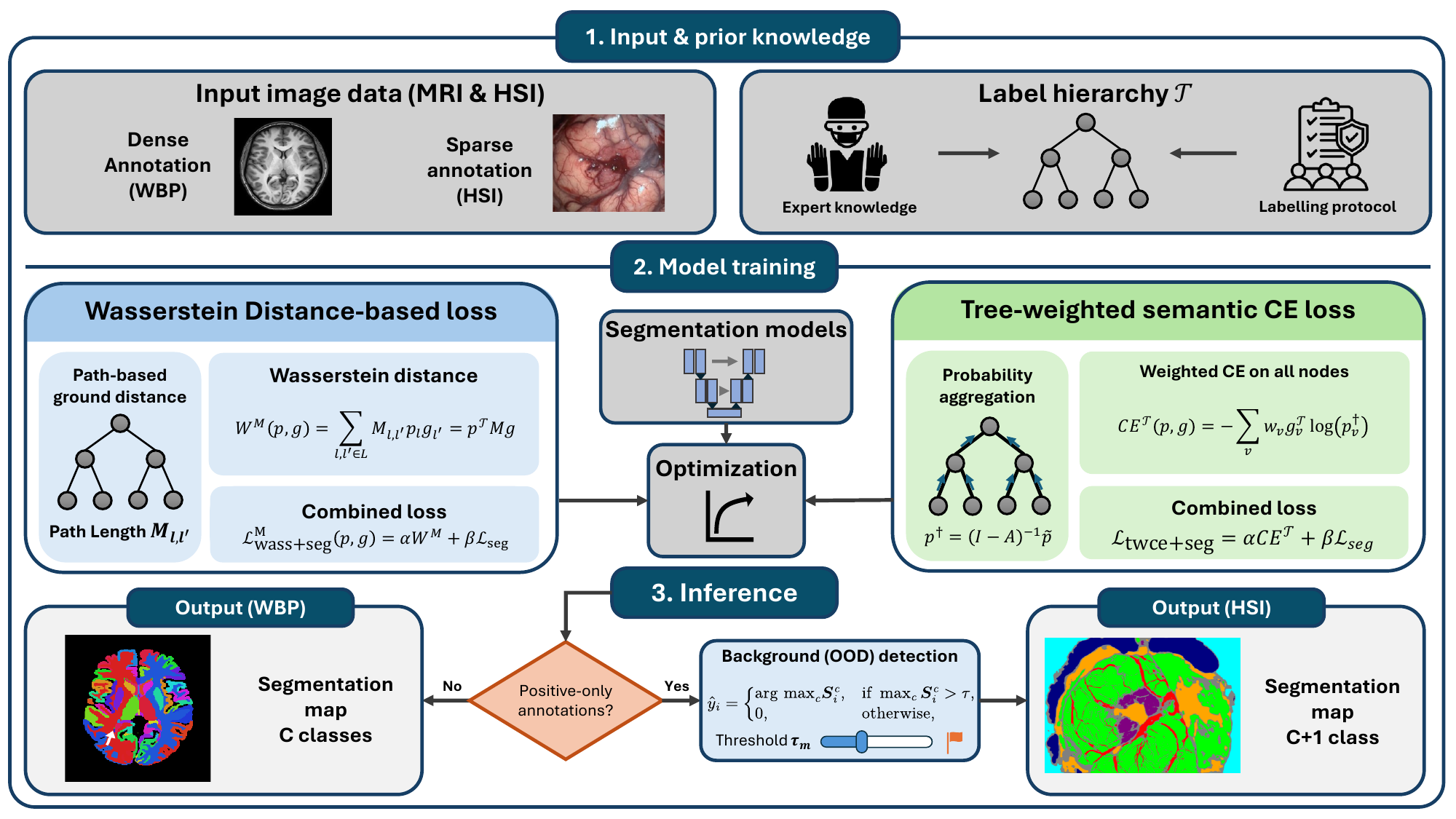}
    \caption{Overview of the proposed tree-semantic loss framework for medical image segmentation. \label{fig:overview_tree_semantic_loss}}
\end{figure}
\begin{figure}[tbp]
    \centering
    \includegraphics[width=0.72\linewidth]{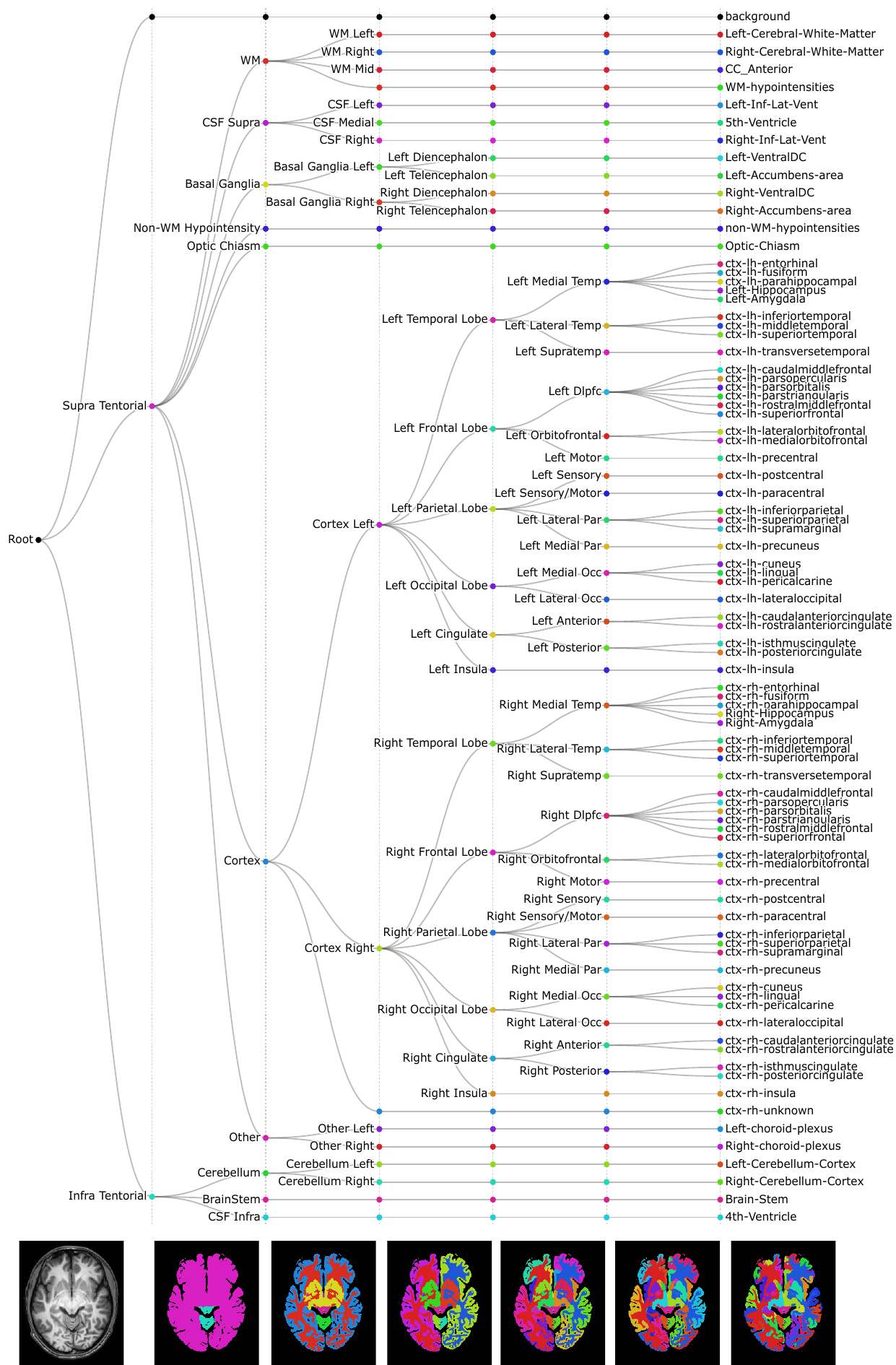}
    \caption{The neuro-anatomical label hierarchy of Mindboggle dataset. From left to right, the hierarchy progresses from coarse object categories to specific classes. Rich annotations correspond to leaf node classes.
    The colour coding matches the ground-truth mask at each level. \label{fig:hierarchy_mindboggle}}

\end{figure}
Segmentation plays a crucial role in medical and surgical image analysis by locating and precisely outlining regions of interest such as organs, lesions, and tissues across a variety of imaging modalities. Two particularly important brain imaging applications that rely on rich and accurate segmentation are head MRI for whole brain parcellation (WBP) and interventional hyperspectral imaging (iHSI) for scene understanding and tissue characterisation. WBP divides an MRI volume into spatially coherent, anatomically or functionally meaningful brain regions~\citep{eickhoffImagingbasedParcellationsHuman2018}. Hyperspectral imaging (HSI) captures wide-field views across dozens to hundreds of optical spectral bands, and can be used intra-operatively to reveal biochemical contrasts invisible to the naked eye \citep{shapeyIntraoperativeMultispectralHyperspectral2019}.

A major challenge with such rich segmentation tasks relates to the granularity at which the data is annotated and that at which segmentation models should operate. Recent works have examined the impact of training models with various levels of labelling granularity such as pixels, patches, and entire images~\citep{seidlitzRobustDeepLearningbased2022,garciaperazaherreraHyperspectralImageSegmentation2023}. Other studies have provided comparative analyses of different pixel-level algorithms for brain tissue differentiation \citep{martin-perezMachineLearningPerformance2024}. Underpinning these questions is the drive for a holistic and refined understanding of the images and surgical scenes. Annotation efforts are ongoing to provide training data across large number of potentially subtly varying classes. Combined with sparse annotations processes that may be employed to label data at scale, many of these classes may only have small amounts of training samples. It is thus important to take advantage of the semantics of the labels and realise that some types of errors are more acceptable than others. However, there is only limited previous work in medical imaging that has leveraged the structure of the label space as a source of information. In surgical imaging, to our knowledge, no such efforts have been published. By contrast, the importance of label semantics is getting recognised in the general field of computer vision~\citep{frognerLearningWassersteinLoss2015,leTreeSlicedVariantsWasserstein2019a,bertinettoMakingBetterMistakes2020}. Adopting these concepts to WBP and iHSI is expected to produce more robust models for both dense and sparse annotation setting.

Additional challenges arises when labelling needs to be performed at scale. Leveraging sparsely annotated datasets becomes an effective strategy that has been widely adopted already in iHSI segmentation and surgical imaging~\citep{studier-fischerHeiPorSPECTRALHeidelbergPorcine2023,hyttinenOralDentalSpectral2020,carstensDresdenSurgicalAnatomy2023}. For instance, a recent general surgery iHSI dataset adopted a sparse annotation protocol by labelling only representative image regions, omitting marginal areas, superficial blood vessels, adipose tissue, and other artefacts~\citep{studier-fischerHeiPorSPECTRALHeidelbergPorcine2023}. Within the neurosurgical HSI dataset used in this study, representative examples of this labelling strategy are illustrated in the second column of \figref{fig:qualitative_results_all_classes}. In sparsely annotated medical image segmentation, the absence of a label cannot be taken as evidence that a region is negative. A truly positive pixel may go unmarked for two reasons: (i) the annotator finds the region ambiguous or (ii) it is skipped due to time constraints. The most straightforward, albeit wrong, approach would be to presume that every unlabelled pixel belongs to the negative background class. To appropriately address such partial supervision, \citet{wang2024oodsegoutofdistributiondetectionimage} recently proposed a framework that learns from background-free, positive-only sparse label masks. Pixel-wise out-of-distribution (OOD) detection methodology is used at inference time to flag as background any tissue or object type that has not been annotated in the training data.

\revdel{In this work, we propose two tree-based semantic loss functions for supervised segmentation that incorporate prior knowledge of label structure.} \revmod{In this work, we exploit semantic relationships in the label structure through a general tree-based semantic supervision framework, instantiated by two complementary loss functions for supervised segmentation.}\revref{2}{1}
\figref{fig:overview_tree_semantic_loss} provides an overview of the proposed framework for medical image segmentation. A hierarchical label tree encodes the semantic relationships among classes and serves as the basis for two \revdel{complementaryloss}\revmod{complementary loss} functions. For WBP, the hierarchy is derived from established anatomical guidelines based on the Desikan--Killiany--Tourville (DKT) protocol~\citep{klein101LabeledBrain2012} (\figref{fig:hierarchy_mindboggle}, with a larger version available in the supplementary material and online via \href{https://observablehq.com/@junwens-project/mindboggle-label-hierarchy}{an interactive hierarchy visualisation}); for surgical HSI, the hierarchy is defined by expert consensus, with the full structure provided online via \href{https://observablehq.com/@junwens-project/ihsi-hierarchy}{an interactive hierarchy visualisation} and in supplementary material \figref{fig:hierarchy_phasetwo}. We encode this structure directly into optimisation via a \emph{Wasserstein distance-based segmentation loss}, which penalises errors according to path length in the label tree, and a \emph{tree-weighted semantic cross-entropy} loss, which extends weighted cross-entropy to all nodes in the hierarchy.
\revnew{Both losses incorporate semantic information into training, but they use the hierarchy differently: the Wasserstein formulation changes the cost of leaf-label mistakes through pairwise tree distances, whereas tree-weighted cross-entropy supervises aggregated probabilities at internal as well as leaf nodes.}\revref{2}{1}
We further integrate these losses into the positive-only sparse supervision framework of~\citep{wang2024oodsegoutofdistributiondetectionimage} to support background detection as OOD without degrading performance on positive (ID) classes.
\revdel{Across rich segmentation tasks in brain MRI and neurosurgical iHSI, the proposed losses achieve state-of-the-art performance.}\revmod{Across rich segmentation tasks in brain MRI and neurosurgical iHSI, the proposed losses improve over the evaluated baselines in several clinically relevant settings.}\revref{2}{3}\secondrevref{1}{4}
\revdel{In particular, experiments on three MRI WBP datasets show that the Wasserstein-based loss consistently surpasses the CE + Dice nnU-Net baseline, with especially marked gains for small structures with limited annotations, while experiments spanning two tasks and four datasets demonstrate that the proposed framework can be incorporated into standard segmentation models without architectural modification and remains effective under both sparse and dense annotation regimes.}
\revmod{In particular, experiments on three MRI WBP datasets show that the Wasserstein-based loss provides the strongest support among the evaluated losses, especially for NSD and small-structure metrics, while experiments spanning two tasks and four datasets demonstrate that the proposed framework can be incorporated into standard segmentation models without architectural modification in both sparse and dense annotation settings.}\revref{2}{3}\secondrevref{1}{4}
The code is available at \href{https://github.com/cai4cai/nnunet-tree-semantic-extension}{https://github.com/cai4cai/nnunet-tree-semantic-extension}.

The main contributions of this paper can be summarised as follows:
\begin{itemize}
    \item We introduce two tree-based semantic losses for medical image segmentation, namely a Wasserstein distance-based segmentation loss and a tree-weighted semantic cross-entropy loss, which explicitly encode hierarchical relationships between labels during training.
    \item We show how these losses can be applied to both densely supervised whole-brain parcellation and sparsely supervised neurosurgical HSI segmentation, including integration with a positive-only OOD detection framework for background identification.
    \item We demonstrate that incorporating expert-defined or anatomically derived label hierarchies leads to more semantically meaningful supervision and improved segmentation performance relative to standard baselines.
    \item \revdel{We establish strong empirical performance across two segmentation tasks and four datasets, with the Wasserstein-based loss consistently improving over the CE + Dice nnU-Net baseline for WBP,}\revmod{We provide empirical evidence across two segmentation tasks and four datasets, with the Wasserstein-based loss consistently improving over the CE + Dice nnU-Net baseline for WBP,}\revref{2}{3}\secondrevref{1}{4} \revnew{evaluated using manual references on MB42 and automated GIF-derived pseudo-ground-truth masks on AOMIC/IXI,} particularly on small structures with limited annotations.\revref{1}{3}
\end{itemize}

\section{Related work}\label{sec:related_work}
\subsection{Semantic segmentation in WBP}
Although the desired granularity varies across protocols, whole-brain parcellation (WBP) typically entails segmenting hundreds of distinct classes~\citep{klein101LabeledBrain2012}. Classical tools such as FreeSurfer~\citep{fischlWholeBrainSegmentation2002} and GIF (Geodesic Information Flows)~\citep{cardosoGeodesicInformationFlows2015a} automatically deliver robust and accurate results, but processing a single scan can take from several hours to an entire day. 
\revnew{In particular, GIF is a well-established multi-atlas segmentation and label-fusion method for brain parcellation that has been used successfully in recent clinical neuroimaging studies.~\citep{bocchettaStructuralMRIPredicts2023,youngCharacterizingClinicalFeatures2021}.}\revref{1}{3}
Deep-learning approaches have been proposed to shorten inference time~\citep{henschelFastSurferFastAccurate2020,guharoyQuickNATFullyConvolutional2019,royErrorCorrectiveBoosting2017} at the cost of increased GPU memory requirements. Recently, \citet{kujawaLabelMergeandSplitGraphColouring2024} introduced a label merge-and-split framework that clusters spatially disjoint regions with a greedy graph-colouring algorithm, allowing the network to predict a much smaller set of merged labels and then restores the original labels at inference via atlas-derived influence regions. However, with the exception of \citep{grahamHierarchicalBrainParcellation2020a} discussed in \secref{sec:hierarchlosss}, previous learning-based WBP approaches have not exploited the semantic relationships within the label classes.

\subsection{Semantic segmentation in hyperspectral imaging}
Semantic segmentation works have initially relied on classical machine learning pipelines \citep{raviManifoldEmbeddingSemantic2017,fabeloSpatiospectralClassificationHyperspectral2018,mocciaUncertaintyAwareOrganClassification2018a} and more recently adopted deep learning \citep{khanTrendsDeepLearning2021}. Many studies \citep{seidlitzRobustDeepLearningbased2022,trajanovskiTongueTumorDetection2021} adopt a U-Net type architecture \citep{ronnebergerUNetConvolutionalNetworks2015,jegouOneHundredLayers2017}. Previous work also examined the impact of training models with various levels of input granularity such as spectral pixels (1D CNNs), image patches, and entire HSI images (2D CNNs) \citep{seidlitzRobustDeepLearningbased2022,garciaperazaherreraHyperspectralImageSegmentation2023,martin-perezMachineLearningPerformance2024} concluding that providing both spectral and spatial context is beneficial. \citet{seidlitzRobustDeepLearningbased2022} segmented 20 organ types from 506 hypercubes taken from 20 pigs, demonstrating best performance achieved when models were trained on full images rather than on pixels or patches~\citep{seidlitzRobustDeepLearningbased2022}. Similar results were obtained in \citep{garciaperazaherreraHyperspectralImageSegmentation2023} for HSI-based segmentation of dental tissues from the ODSI-DB dataset \citep{hyttinenOralDentalSpectral2020}, which includes data from 30 human subjects annotated with 35 tissue types. These works have employed sparse annotations but have not exploited dedicated learning approaches for this, leading to suboptimal results, as discussed in \secref{sec:sparseannot}. Furthermore, only small label sets were used with no insight into the semantics relationships between labels.

\subsection{Hierarchical loss functions}\label{sec:hierarchlosss}
Hierarchical loss functions encode class-taxonomy information directly in optimisation, so prediction errors are penalised by semantic severity rather than uniformly. Early work in large-scale visual recognition established this principle: hierarchy-aware objectives based on WordNet and lowest-common-ancestor structure~\citep{dengWhatDoesClassifying2010}, class-similarity-aware multiclass learning with shared regularisation~\citep{zhaoLargeScaleCategoryStructure2011}, and taxonomy-conditioned metric learning~\citep{vermaLearningHierarchicalSimilarity2012}.

Outside medical imaging, deep-learning formulations made hierarchy-aware supervision more practical and expressive. ``Better-mistake'' cross-entropy variants in~\citep{bertinettoMakingBetterMistakes2020} redistributed penalties toward semantically nearby classes, while optimal-transport formulations~\citep{frognerLearningWassersteinLoss2015} and later tree-based Wasserstein variants~\citep{leTreeSlicedVariantsWasserstein2019a,takezawaSupervisedTreeWassersteinDistance2021} provided principled distance-based objectives. For dense prediction,~\citep{liSemanticHierarchyAwareSegmentation2024} further showed that hierarchy-constrained learning can be applied to semantic segmentation through pixel-wise multi-label formulations.

Within medical imaging, earlier hierarchy-aware work was comparatively limited. A brain-parcellation approach in~\citep{grahamHierarchicalBrainParcellation2020a} predicted along the hierarchy to model uncertainty, but with increased output complexity and without clear segmentation-accuracy gains over leaf-only training. Wasserstein-based Dice objectives~\citep{fidonGeneralisedWassersteinDice2018,fidonGeneralizedWassersteinDice2022} introduced semantically meaningful penalties for multi-class segmentation, though evaluations were mostly in lower-cardinality BraTS-style settings~\citep{menzeMultimodalBrainTumor2015}. In~\citep{dengHATsHierarchicalAdaptive2024}, hierarchical supervision is achieved by encoding anatomical taxonomy and scale relations in a unified matrix and optimising a taxonomy-consistent loss. By contrast,~\citep{luH2ASegHierarchicalAdaptive2024} introduces multi-level cross-modal interaction with target-aware modality weighting to hierarchically refine PET/CT tumour features.

Taken together, these studies support hierarchy-aware supervision, but gaps remain for our target setting: rich medical label spaces, compatibility with standard segmentation architectures, and training under sparse positive-only annotations where background labels are absent.

\subsection{Medical image segmentation with sparse annotation}\label{sec:sparseannot}
Pixel- or voxel-level annotation of medical images is time-consuming and costly. Early work showed that accurate segmentation can be achieved from only a handful of labelled regions when unlabelled pixels or voxels are ignored in the loss~\citep{cicek3DUNetLearning2016}. The finding inspired much of the subsequent weakly supervised learning (WSL) literature.

Existing WSL methods utilise various modes of sparse annotations, including image-level annotation~\citep{kuangWeaklySupervisedLearning2024}, bounding box~\citep{wangBoundingBoxTightness2021,wangInteractiveMedicalImage2018}\revnew{~\citep{weiWeakPolypYouOnly2023}}\revref{2}{4}, scribbles~\citep{wangDeepIGeoSDeepInteractive2019}\revnew{~\citep{wangFewMoreScribblebased2025}}\revref{2}{4}, points~\citep{glockerVertebraeLocalizationPathological2013,dorentInterExtremePoints2021}\revnew{~\citep{enAnnotationClicksPointSupervised2022}}, and 2D slices within a 3D structure~\citep{cai3DMedicalImage2023}.
These methods all focus on using the sparse annotations at training time to produce full segmentation mask at test time. As an example, in \citep{glockerVertebraeLocalizationPathological2013}, the authors introduced a semi-automatic labelling strategy that transforms sparse point-wise annotations into dense probabilistic labels for vertebrae localisation and identification. In \citep{xuWeaklySupervisedHistopathology2014}, the authors propose to segment both healthy and cancerous tissue from colorectal histopathological biopsies using bounding boxes. In \citep{wangInteractiveMedicalImage2018}, the authors reported improved CNN performance on sparse annotated input through image-specific fine-tuning.
\revdel{Finally, }In \citep{wangDeepIGeoSDeepInteractive2019}, the authors combined sparsely annotated input with a CNN through geodesic distance transforms, followed by a resolution-preserving network resulting in better dense prediction.
However, all of these methods largely ignore the harder setting, which we face here, in which the model does not have guidance on what should be treated as non-object context. To address this limitation, \citet{wang2024oodsegoutofdistributiondetectionimage}, which we detail in \secref{sec:ood_detection}, proposed a segmentation framework that allows for positive-only learning by exploiting out-of-distribution (OOD) detection mechanisms.

\subsection{Out-of-distribution detection and positive-only learning}\label{sec:ood_detection}
Several studies have explored OOD detection within the context of image classification~\citep{hendrycksBaselineDetectingMisclassified2017,liangEnhancingReliabilityOutofdistribution2018,leeSimpleUnifiedFramework2018,hsuGeneralizedODINDetecting2020}. As an early example exploiting deep learning, \citep{hendrycksBaselineDetectingMisclassified2017} proposed using the maximum softmax score as a baseline for OOD detection based on an observation that correctly classified images tend to have higher softmax probabilities than erroneously classified examples. \citet{liangEnhancingReliabilityOutofdistribution2018} found that applying confidence calibration through temperature scaling~\citep{guoCalibrationModernNeural2017a} effectively separates ID and OOD images. \citet{leeSimpleUnifiedFramework2018} suggested measuring the Mahalanobis distance between test image features and the training distribution from the penultimate convolutional layer of the model. \citet{hsuGeneralizedODINDetecting2020} proposed decomposing the confidence score to adapt the temperature during training.

Despite methodological advances and positive demonstration for image classification purposes, application of OOD detection in medical image segmentation is uncommon. Some studies hypothesize that this may be due to the lack of OOD-based evaluation protocols and the difficulty in gathering relevant data for it~\citep{lambertTrustworthyClinicalAI2024,bulusuAnomalousExampleDetection2020}. Recent research has attempted to address this issue by using other datasets as OOD examples. \citet{karimiImprovingCalibrationOutofDistribution2023} used two separate datasets: one for training the neural network and evaluating its performance on ID data, and another for testing specifically for OOD detection. \citet{gonzalezDistancebasedDetectionOutofdistribution2022} collected four types of OOD datasets to account for different distribution shifts from ID data for COVID-19 lung lesion segmentation task. However, acquiring an additional dataset that can be considered OOD is a difficult and time-consuming process. Therefore, a more scalable approach would be to establish both training and evaluation within a single dataset.

Recently, \citep{wang2024oodsegoutofdistributiondetectionimage} introduced a framework that addresses sparse multi-class positive-only segmentation learning by employing pixel-level out-of-distribution (OOD) detection to detect regions that do not correspond to any of the annotated classes in the training set. The model output probabilities are treated as a reliable signal for pixel-level out-of-distribution (OOD) detection. The network is trained solely on positive (in-distribution) classes, and a pixel is flagged as background (OOD) when the maximum predicted probability across these classes falls below a threshold. To remedy the absence of dedicated sparse positive-only segmentation benchmarks, the authors further devised a two-level cross-validation scheme. By iterating not only over subjects but also over subsets of the label space, this enables a rigorous and comprehensive evaluation using existing annotations only.

\section{Methodology}
This section provides detailed methodology for our proposed loss functions (\secref{sec:wasserstein_based_loss} and \secref{sec:tree_weighted_semantic_loss}). Furthermore, we demonstrate how these can be integrated in the approach of \citep{wang2024oodsegoutofdistributiondetectionimage} for segmenting background pixels from positive-only annotations (\secref{sec:ood_seg}).

\subsection{Wasserstein distance in label space}\label{sec:wasserstein_based_loss}
Let $\mathbf{L}$ be the label space with $C$ leaf nodes, where $\mathbf{L} = \{1,2,\dots,C\}$. Let $p,q \in P(\mathbf{L})$ be probability vectors on $\mathbf{L}$. The Wasserstein distance between $p$ and $q$ is the minimal cost to transform $p$ into $q$ given the ground distances $M_{l,l'}\in\mathbb{R}^{+}$ between any two labels $l$ and $l'$. The ground distance is represented as a matrix $M$ and the associated Wasserstein distance $W^M(p,q)$ is defined through an optimal transport problem:
\begin{equation}\label{eq:wassdist}
\begin{split}
W^M(p,q) &= \min_{T_{l,l'}}\sum_{l,l'\in\mathbf{L}} T_{l,l'}M_{l,l'} \\
\textrm{subject to } \forall l\in\mathbf{L},\sum_{l'\in\mathbf{L}} T_{l,l'} &= p_l \\
\textrm{and } \forall l'\in\mathbf{L},\sum_{l\in\mathbf{L}} T_{l,l'} &= q_{l'}
\end{split}
\end{equation}
By leveraging the distance matrix $M$ on $\mathbf{L}$, the Wasserstein distance yields a semantically-meaningful way of comparing two label probability vectors. Given a tree structure with weights associated to the edges, a semantic ground distance can be induced by the path lengths between the leaf nodes. If $q=g$ is a crisp ground truth, a closed-form expression of \eqref{eq:wassdist} is given in \citep{fidonGeneralisedWassersteinDice2018}:
\begin{equation}\label{eq:wassdistcrisp}
W^{M}(p,g) = \sum_{l,l'\in\mathbf{L}} M_{l,l'}p_l g_{l'} = p^T M g
\end{equation}

While \eqref{eq:wassdistcrisp} can be used directly as the loss for training a segmentation model, prior work has shown benefits in combining generic and task-specific losses \citep{maLossOdysseyMedical2021}. Segmentation frameworks such as the nnU-Net \citep{isenseeNnUNetSelfconfiguringMethod2021} have also been optimised for working with combined losses such a weighted sum of Dice and CE. We generalise the formulation in our preliminary work \citep{wangTreebasedSemanticLosses2025} and propose combining \eqref{eq:wassdistcrisp} with a generic segmentation loss $\loss_{\text{seg}}$ to obtain a compound Wasserstein distance based segmentation loss:
\begin{equation}\label{eq:wdsl}
\loss_{\text{wass+seg}}^M(p,g) = \alpha W^M + \beta \loss_{\text{seg}}
\end{equation}

\subsection{Tree-weighted semantic cross-entropy loss}\label{sec:tree_weighted_semantic_loss}
We also propose another approach to building semantic loss functions by computing the aggregated probabilities across all the nodes in the tree hierarchy, not just the leaf nodes. A segmentation loss such as CE can then be evaluated across all node probabilities.

\revdel{Let the label tree $\mathcal{T}$ be composed of $K$ level with $0$ corresponding to the deepest level (leaf nodes). Let $A$ be the adjacency matrix associated with $\mathcal{T}$.} \revmod{ Let the label tree $\mathcal{T}$ be composed of $K$ levels, with level $0$ corresponding to the deepest level, i.e. the leaf nodes. Let $A$ be the adjacency matrix associated with $\mathcal{T}$, which encodes the parent--child links in the tree: $A_{u,v}=1$ if node $u$ is the parent of node $v$, and $A_{u,v}=0$ otherwise.} \revref{1}{7} Let $\tilde{p}$ be a zero-padding of $p$ to initially associate non-leaf nodes with a zero mass, and $p^{\dagger}$ be the vector collecting all the probabilities:
\begin{equation}\label{eq:collected_prob}
p^{\dagger} = \left(\sum_{k \geq 0} A^k\right)\tilde{p} = (I-A)^{-1}\tilde{p}
\end{equation}
\revdel{where $A^k = 0$ for $k > K$.}
\revmod{For a finite tree, $A$ is nilpotent: if the longest leaf-to-root path has $K$ edges, then $A^k=0$ for $k>K$. Hence $(I-A)^{-1}=I+A+\cdots+A^K$ is a finite sum.}\revref{1}{7}
\revnew{For a three-level tree ordered as $(r,a,b,c,d,e,f)$, with $p=(0.1,0.1,0.3,0.5)^T$ on the leaves, $\tilde{p}=(0,0,0,0.1,0.1,0.3,0.5)^T$ and $p^\dagger=(1,0.2,0.8,0.1,0.1,0.3,0.5)^T$. \figref{fig:tree_matrix_example} shows the tree and the corresponding adjacency matrix.}\revref{1}{7}
\input{figures/tree_matrix_example}
Here, we focus on an extended CE weighted according to domain specific insight:
\begin{equation}\label{eq:cetree}
CE^{\mathcal{T}}(p,g) = - \sum_v w_v g^{\dagger}_v\log(p^{\dagger}_v)
\end{equation}
where $w_v$ is the weight of the edge associated with $v$ as a child node. We note that if $w_v=1$ for all leaf nodes and $w_v=0$ otherwise, equation \eqref{eq:cetree} reduces to the standard CE.

Similar to the Wasserstein case in \secref{sec:wasserstein_based_loss}, equation~\eqref{eq:cetree} is combined with a generic segmentation loss $\loss_{\text{seg}}$. We refer to our semantically-informed variant of the segmentation loss as the tree-weighted semantic segmentation loss:
\begin{equation}\label{twce+seg}
\loss_{\text{twce+seg}} = \alpha CE^{\mathcal{T}} + \beta \loss_{\text{seg}}
\end{equation}
\revnew{The coefficients $\alpha$ and $\beta$ in \eqref{eq:wdsl} and \eqref{twce+seg} are loss-specific hyperparameters; we reuse the symbols only to indicate their shared role in balancing the semantic and segmentation terms.}\revref{2}{5}

\subsection{Learning from sparse positive-only annotations}\label{sec:ood_seg}
We build on the recent work \citep{wang2024oodsegoutofdistributiondetectionimage} to learn segmentation from sparse multi-class positive-only annotations. We extend their approach based on pixel-wise OOD detection methodology to benefit from our tree-based semantic losses. Given an image $x$, each spatial location $i$ is associated with a class label $y_i$. An annotated pixel $i$ in the sparse positive-only training set is such that $y_i \in \{c\} = \{1,2,\dots,C\}$ and $C$ is the number of positive classes. $c=0$ is retained to denote background pixels. By construction, no background annotation is available at training time but OOD detection can be used to differentiate positive classes from the background at inference time.

The framework starts by training a segmentation model using only the positive classes. For clarity, the background class is not a possible output of the network but OOD-tailored training may be used to improve the performance of the next step. The framework then employs a confidence score from an OOD detection mechanism with a threshold $\tau$ to flag background pixels at inference time. The model prediction becomes:
\begin{equation}
\hat{y}_{i} =
\begin{cases}
\operatorname*{arg\,max}_c \boldsymbol{S}_{i}^c, & \text{if } \max_c \boldsymbol{S}_{i}^c > \tau, \\
0, & \text{otherwise},
\end{cases}
\label{eqa:ood_decision}
\end{equation}
where $\boldsymbol{S}_{i}^c$ is a scoring function that captures the probability of pixel $i$ belonging to the positive class $c$, while acknowledging the possibility of it being background / OOD. $\boldsymbol{S}_{i}^c$ can be selected from various OOD detection methods used in image classification tasks \citep{hendrycksBaselineDetectingMisclassified2017,liangEnhancingReliabilityOutofdistribution2018,leeSimpleUnifiedFramework2018,hsuGeneralizedODINDetecting2020}.

To use the hierarchy for OOD detection, we aggregate probabilities using \eqref{eq:collected_prob} and threshold scores at a chosen hierarchy level $h$ instead of at the leaf level. Let $\mathcal{V}_h$ be the nodes at level $h$, with $h=0$ for leaves. We use the coarsest non-root level, $h=K-1$. For pixel $i$ and node $c_h\in\mathcal{V}_h$, the score in \eqref{eqa:ood_decision} is $S_i^{c_h}=p^\dagger_{i,c_h}$; the maximum and argmax are then taken over $c_h\in\mathcal{V}_h$.

\section{Experimental setup}
This section details the configuration used to evaluate our tree-based loss functions. We consider two segmentation tasks: 1) 3D MRI based WBP with full supervision; and 2) 2D hyperspectral surgical scene segmentation with sparse positive-only annotations. All methods share identical hyper-parameters unless otherwise noted, ensuring that performance differences arise solely from the loss functions themselves.

\subsection{Dataset}\label{sec:dataset}
\subsubsection{\texorpdfstring{\revdel{WBP with full annotations}\revmod{WBP datasets with dense reference masks}\robustrevref{1}{3}}{WBP datasets with dense reference masks}}
We evaluated our approach on 3D T1-weighted MR images drawn from three openly available datasets. The \textbf{Mindboggle101} collection provides 101 manually annotated scans aggregated from multiple public sources \citep{klein101LabeledBrain2012}. In Mindboggle101, two constituent subsets (NKI-RS-22 and NKI-TRT-20) were combined to create a 42-image test cohort, hereafter termed Mindboggle42 or \textbf{MB42} for short. The remaining 59 scans, referred to as Mindboggle59 or \textbf{MB59}, served as multi-atlas data for the classical GIF algorithm \citep{cardosoGeodesicInformationFlows2015}. 
\revnew{Because manual WBP annotation is often difficult due to single manual annotations may contain local inconsistencies for fine anatomical structures. Automated GIF-derived masks can therefore be qualitatively more anatomically coherent in some regions.}\revref{1}{3} 

\revdel{GIF with Mindboggle59 was used to generate pseudo-ground-truth WBP masks for two other datasets which otherwise do not provide WBP annotations.}
\revmod{GIF with Mindboggle59 was used to generate GIF-derived pseudo-ground-truth WBP masks for two other datasets which otherwise do not provide manual WBP annotations. Accordingly, AOMIC and IXI metrics quantify agreement with automated GIF-derived pseudo-ground-truth masks rather than direct accuracy against manual annotations, and these experiments cannot quantify GIF label error without additional manual AOMIC/IXI WBP labels.}
\revref{1}{3} The \textbf{AOMIC} PIOP2 dataset \citep{snoekAmsterdamOpenMRI2021} contributes 226 MRI scans (180 training, 46 testing) acquired from healthy participants aged 18--25 years. Finally, the \textbf{IXI} dataset\footnote{\url{https://brain-development.org/ixi-dataset/}} comprises 581 scans (464 training, 117 testing) collected from healthy individuals aged 20--86 years. By construction, all three datasets share the label space arising from the DKT protocol~\citep{klein101LabeledBrain2012}. The DKT protocol does not directly provide a label hierarchy but its label space is very similar to that used in the original GIF implementation for which a label hierarchy is provided in \citep{grahamHierarchicalBrainParcellation2020a}. Manual matching was performed to adapt the hierarchy in \citep{grahamHierarchicalBrainParcellation2020a} to the DKT label set. The resulting hierarchy is shown in \figref{fig:hierarchy_mindboggle} and \href{https://observablehq.com/@junwens-project/mindboggle-label-hierarchy}{an interactive hierarchy visualisation}.

\subsubsection{Surgical HSI dataset with sparse positive-only annotations}
The data was obtained from patients undergoing microscopic cranial neurosurgery as part of an ethically approved single-centre, prospective clinical observational investigation employing a prototype hyperspectral imaging system (NeuroHSI study: REC reference 22/LO/0046, ClinicalTrials.gov ID NCT05294185). Informed consent was obtained from all participants. The primary objective was to evaluate the intra-operative utility of a \(4 \times 4\), 16-band visible-range snapshot mosaic camera (IMEC CMV2K-SSM4X4-VIS) mounted on a surgical microscope.

The dataset comprises 22,829 annotated frames derived from 45 distinct patients, encompassing both neuro-oncological and neurovascular pathologies. Multiple videos were acquired throughout each case, with each recording representing a specific surgical phase intended to capture relevant intra-operative details. The data includes varying visual perspectives due to changes in surgical microscope positioning. Training snapshot data are first processed with a demosaicking pipeline \citep{liDeepLearningApproach2022,Li_2024_BMVC}, resulting in 1080p frames with 16 channels (hypercubes). As in \citep{Li_2024_BMVC}, these hypercubes can be converted to synthetic standard RGB (sRGB) for visualisation purposes.

The sparse, background-free annotations encompass 107 subclasses organised into a hierarchical structure defined by neurosurgeons. Due to space constraints, the full label hierarchy is provided in the supplementary material and online via \href{https://observablehq.com/@junwens-project/ihsi-hierarchy}{an interactive hierarchy visualisation}, with corresponding colour references displayed at each node in the label hierarchy. For each video a representative subset of frames was selected by an experienced neurosurgeon to minimise motion blur, maximise the number of tissue classes included and ensure key surgical phases were represented. Selected frames were manually annotated by two neurosurgeons who had also been present during the surgical procedure. In cases where tissue class was ambiguous based on HSI-derived sRGB images alone, corresponding high resolution snapshot pictures taken using the integrated surgical microscope camera were correlated with hyperspectral data to determine tissue class, and if necessary discussed with the operating surgeon. Where definitive identification of tissue class was not possible, the area was left unlabelled. Where feasible, these manual annotations were then propagated across subsequent frames algorithmically using the registration-based propagation feature of the ImFusion Labels software. Each propagated annotation was verified and corrected (when needed) by a neurosurgeon prior to final submission.

We also note that the coupling of the camera onto the surgical microscope induces a partial masking of the sensor on the outside of the circular field of view of the microscope. In addition to human-labelled categories, an additional label was generated by a content area estimation algorithm \citep{buddRapidRobustEndoscopic2023}. This algorithm provides a robust estimation of the location of the non-informative areas on the sensor. By treating these areas as part of our label hierarchy, the model can more effectively discriminate content regions. \figref{fig:qualitative_results_all_classes} presents example sRGB image plus annotation overlays.

\subsection{Implementation details}\label{sec:implementation}
There are two distinct experimental configurations, depending on the dataset and task. Across all experiments, when training with our tree semantic losses, we used the same hyperparameters as those used in the baseline approach to ensure a fair comparison. Similar to \citep{wangTreebasedSemanticLosses2025}, for all compound losses, we set $\alpha = \beta = 0.5$.
\revnew{Note that this shared numerical setting does not imply that the two losses share same coefficients.}
\revref{2}{5} All experiments ran on an NVidia DGX cluster with V100 (32GB) and A100 (40GB) GPUs.

\subsubsection{WBP}
For the WBP task, we use the nnU-Net framework~\citep{isenseeNnUNetSelfconfiguringMethod2021} and only modify the loss function. This framework employs a built-in empirical rule to automatically decide all hyperparameters based on statistics extracted from the training set. As per the nnU-Net default, the generic segmentation loss is set to $\loss_{\text{seg}} = Dice + CE$.

\subsubsection{HSI}
For the neurosurgical HSI dataset, we adopted a similar training pipeline as described in \citep{wang2024oodsegoutofdistributiondetectionimage}. Specifically, we used a U-Net architecture with an EfficientNet-b5 encoder~\citep{tanEfficientNetRethinkingModel2019}, pre-trained on ImageNet \citep{dengImageNetLargescaleHierarchical2009} for all experiments\footnote{\href{https://github.com/qubvel/segmentation_models.pytorch}{\texttt{github:qubvel/segmentation\_models.pytorch}}}. Given the sparsely annotated nature of the training data, the Dice loss is not applicable. We thus opt for $\loss_{\text{seg}} = CE$ for Wasserstein based loss and $\loss_{\text{seg}} = 0$ for tree-weighted CE loss to prevent computing CE on leaf node classes twice. We employed the Adam optimiser \citep{kingmaAdamMethodStochastic2017} with $\beta_1 = 0.9$ and $\beta_2 = 0.999$, together with an exponential learning rate scheme ($\gamma = 0.999$). We set the initial learning rate to $0.001$, used a mini-batch size of $5$, and trained for a total of $50$ epochs. For data augmentation, we adopted a similar setup to that reported in \citep{wang2024oodsegoutofdistributiondetectionimage}: random rotation (rotation angle limit: $45^\circ$), random flipping, random scaling (scaling factor limit: $0.1$), and random shifting (shift factor limit: $0.0625$). All transformations were applied with a probability of $0.5$. In addition, we apply $\ell^1$-normalisation at each spatial location to account for the non-uniform illumination of the tissue surface. This is routinely applied in HSI because of the dependency of the signal on the distance between the camera and the tissue \citep{bahlSyntheticWhiteBalancing2023,studier-fischerHeiPorSPECTRALHeidelbergPorcine2023}.

\subsubsection{Choice of tree weights / ground distances}
Edge weights can be defined across hierarchical levels, producing distinct cost matrices $M$ that affect overall performance. We consider four edge weight setups. A first simple case is $M_t$ that assigns a weight of 1 only to edges at the top-level and 0 elsewhere. A second simple case is $M_{\ell}$ that assigns a weight of 1 only to leaf nodes and 0 elsewhere. Neither $M_t$ nor $M_{\ell}$ effectively take advantage of the label hierarchy and these cases therefore represent baselines. By contrast, $M_e$ and $M_h$ place non-zero weights on every edge. $M_e$ sets all edge weights to 1, whereas $M_h$ imposes a scaling parameter $\kappa$ such that a parent-level edge weight is $\kappa$ times larger than its children.

To focus the number of experiments reported here, for the WBP task, we reuse the best configuration identified in~\citep{wangTreebasedSemanticLosses2025}. We thus only report results with the $M_h$ configuration combined with $\kappa = 10$.

For the HSI experiments, we report all four edge weight configuration. Through preliminary experiment not reported here, we found that the following choice of $\kappa$ yields appropriate performance for the $M_h$ configuration: $\kappa = 10$ for $\loss_{\text{wass+seg}}$ based experiments and $\kappa = 2$ for $\loss_{\text{twce+seg}}$ based experiments.

\section{Results}\label{sec:results}
This section presents quantitative (\secref{sec:brain_parcellation_result} and \secref{sec:hsi_result}) and qualitative (\secref{sec:qualitative_result}) results. In addition, we conducted error analysis by plotting confusion matrix to investigate the effectiveness which model can better infer relationships among different tissue types (\secref{sec:error_analysis}).

\subsection{\texorpdfstring{\revdel{WBP with full annotations}\revmod{WBP with dense reference masks}\robustrevref{1}{3}}{WBP with dense reference masks}}\label{sec:brain_parcellation_result}
\begin{table*}[tb]
\centering
\small
\setlength\tabcolsep{0.5pt}
\renewcommand{\arraystretch}{0.95}
\caption{Cross-dataset performance on whole brain parcellation. Rows correspond to the training split, and columns correspond to the test split. $\Dice$ and $\NSD$ metrics are averaged over all 108 classes, whereas $\SDice$ and $\SNSD$ are averaged over the 10 classes which have smallest annotated region. The best performance among all losses for each training dataset is highlighted in bold. \revnew{Asterisks mark per-dataset proposed-loss comparisons against $\loss_{\text{seg}}$ that remain significant after Holm-Bonferroni correction ($p<0.05$) in matched-subject paired tests. The Avg. columns summarise the three test datasets.}\robustrevref{1}{2} For both $\loss_{twce+seg}$ and $\loss_{wass+seg}$, $M_h$ is chosen for ground distance matrix configuration.\label{tab:brain_parcellation_result}}
\begin{adjustbox}{width=\textwidth}
\input{table/wbp_results_tabular.tex}
\end{adjustbox}
\end{table*}
\tabref{tab:brain_parcellation_result} summarises the cross-dataset evaluation on the three WBP sub-datasets (MB42, AOMIC, and IXI). \revnew{The MB42 test set supports evaluation against manual WBP annotations, while the AOMIC and IXI test sets support fair relative comparison against the same automated GIF-derived pseudo-ground-truth masks for all losses.}\revref{1}{3} Our baseline is the standard nnU-Net, trained with its default loss $\loss_{\text{seg}}$ and hyper-parameters. \revnew{We also include the Generalised Wasserstein Dice loss~\citep{fidonGeneralisedWassersteinDice2018} as an additional baseline.} \revref{1}{1}\revdel{Replacing this loss with one of our proposed compound losses ($\loss_{\text{twce+seg}}$ and $\loss_{\text{wass+seg}}$) yields consistent gains}\revmod{Replacing these baseline losses with one of our proposed compound losses ($\loss_{\text{wass+seg}}$) yields consistent gains}\revref{1}{4}\secondrevref{2}{3} \revnew{in manual-reference performance on MB42 and in pseudo-ground-truth agreement on AOMIC/IXI}. We report the mean \textit{Dice} score and the mean \textit{Normalised Surface Dice} (NSD) metric~\citep{seidlitzRobustDeepLearningbased2022}, where the latter measures the overlap of two volume surface. The surface element is counted as overlapping when the closest distance to other surface is less to 3mm tolerance. The benefits are particularly pronounced for classes with fewer annotations. We performed pixel-level counting across all classes and selected the 10 classes with the fewest annotated pixels. We report the mean Dice and NSD scores of these classes as $\SDice$ and $\SNSD$, respectively.

\revnew{For WBP comparisons, subject-level Dice and NSD values were generated from retained predictions and labels for identical test cases. Statistical comparisons used two-sided paired-sample tests on matched subjects for each predefined train--test dataset pair, metric, and proposed loss against $\loss_{\text{seg}}$. \tabref{tab:wbp_paired_sample_tests} reports the number of paired subjects, paired differences, and raw and Holm-Bonferroni-adjusted significance markers across the 72 predefined tests. Statistical support is interpreted using the adjusted p-values. Under this criterion, $\loss_{\text{wass+seg}}$ shows the strongest support, with adjusted significance for NSD and small-structure metrics in nearly all train--test settings, while Dice gains are not uniformly significant for every dataset pair.}\revref{1}{2}

\revnew{We further quantify the GIF performance for the dataset with manual labels. We applied GIF to the MB42 images using the MB59 manual labels as the multi-atlas set, and compared the resulting pseudo-labels with the manual MB42 annotations. This yielded a mean label-wise $\Dice$ of $78.6 \pm 15.4$. However, direct quantification of GIF error on AOMIC and IXI is not possible because manual WBP labels are not available for these datasets.}\revref{1}{3}

\subsection{Surgical HSI with sparse positive-only annotations}\label{sec:hsi_result}
\begin{table*}[tb]
\centering
\setlength\tabcolsep{5pt}
\renewcommand{\arraystretch}{1.1}
\caption{Cross-validation results on top-level classes of the HSI dataset. For each method and metric, performance is reported at thresholds $\tau_0=0$ and $\tau_m$. $\tau_m$ is chosen from the optimal threshold for foreground classes in the validation set. The best performance among all losses is highlighted in bold. Rows shaded in grey represent the baseline results, which are equivalent to the standard CE or Wasserstein+CE training on leaf node classes only (that is, without class semantics). The asterisk indicates a strong baseline result which is equivalent to standard CE training on top-level nodes only.}
\begin{tabular}{c c cc cc cc}
\toprule
\multirow{2}{*}{\textbf{Loss}} &
&
\multicolumn{2}{c}{$\boldsymbol{\TPR}\uparrow$} &
\multicolumn{2}{c}{$\boldsymbol{\BACC}\uparrow$} &
\multicolumn{2}{c}{$\boldsymbol{\Fone}\uparrow$} \\
\cmidrule(lr){3-4}
\cmidrule(lr){5-6}
\cmidrule(lr){7-8}
\multicolumn{2}{c}{} & $\boldsymbol{\tau_0=0}$ & $\boldsymbol{\tau_m}$
& $\boldsymbol{\tau_0=0}$ & $\boldsymbol{\tau_m}$
& $\boldsymbol{\tau_0=0}$ & $\boldsymbol{\tau_m}$ \\
\midrule
$\loss_{\text{wass}}$ & $M_t$ & $0.51{\pm0.03}$ & $0.51{\pm0.03}$ & $0.74{\pm0.01}$ & $0.74{\pm0.01}$ & $0.47{\pm0.05}$ & $0.47{\pm0.05}$ \\
\midrule
\rowcolor{gray!30} \multirow{4}{*}{$\loss_{twce}$} & $M_{\ell}$ & $0.61{\pm0.04}$ & $0.65{\pm0.05}$ & $0.79{\pm0.02}$ & $0.82{\pm0.02}$ & $0.60{\pm0.02}$ & $0.65{\pm0.03}$ \\
& $M_e$ & $0.64{\pm0.04}$ & $0.76{\pm0.03}$ & $0.80{\pm0.02}$ & $0.88{\pm0.01}$ & $0.62{\pm0.05}$ & $0.74{\pm0.05}$ \\
& $^*M_t$ & $0.65{\pm0.05}$ & $0.73{\pm0.12}$ & $0.81{\pm0.02}$ & $0.86{\pm0.06}$ & $0.63{\pm0.04}$ & $0.72{\pm0.10}$ \\
& $M_h$ & $0.65{\pm0.03}$ & $0.75{\pm0.03}$ & $0.81{\pm0.02}$ & $0.87{\pm0.02}$ & $0.64{\pm0.04}$ & $0.76{\pm0.05}$ \\
\midrule
\rowcolor{gray!30} \multirow{4}{*}{$\loss_{wass+seg}$} & $M_{\ell}$ & $0.61{\pm0.06}$ & $0.70{\pm0.05}$ & $0.79{\pm0.03}$ & $0.85{\pm0.03}$ & $0.62{\pm0.04}$ & $0.72{\pm0.04}$ \\
& $M_e$ & $0.65{\pm0.04}$ & $0.72{\pm0.09}$ & $0.81{\pm0.02}$ & $0.85{\pm0.04}$ & $0.64{\pm0.01}$ & $0.73{\pm0.07}$ \\
& $M_t$ & $0.66{\pm0.04}$ & $0.77{\pm0.07}$ & $0.81{\pm0.02}$ & $0.88{\pm0.04}$ & $0.63{\pm0.03}$ & $0.78{\pm0.09}$ \\
& $M_h$ & $\boldsymbol{0.68}{\pm0.02}$ & $\boldsymbol{0.80}{\pm0.03}$ & $\boldsymbol{0.83}{\pm0.01}$ & $\boldsymbol{0.90}{\pm0.02}$ & $\boldsymbol{0.66}{\pm0.02}$ & $\boldsymbol{0.82}{\pm0.06}$ \\
\bottomrule
\end{tabular}
\label{tab:cross_validation}
\end{table*}
\tabref{tab:cross_validation} presents the cross-validation results on top-level classes by selecting the output probability at the top-level only (that is, $p^{\dagger,K-1}$). We report the results with different confidence thresholds $\tau$. Where $\tau_0=0$ represents no background (OOD) detection and $\tau_m$ represents the threshold which maximises scores across the positive annotations (ID data). To better capture performance on the background, $\tau_m$ could be also computed by incorporating held-out classes for background / OOD performance monitoring during validation in the two-level cross-validation \citep{wang2024oodsegoutofdistributiondetectionimage}. However, this is not presented here due to time constraints.

For all loss functions we evaluate the \textit{\revdel{Truth}\revmod{True} Positive Rate (TPR)}, \textit{Balanced Accuracy (BACC)}, and \textit{F1 scores}. Results are reported by averaging across classes under one-vs-rest strategy, where the positives are pixels of such class, and the negatives are the pixels of all the other classes. We do not report IoU for the HSI task because the training and evaluation setting is based on sparse positive-only annotations, for which unlabelled pixels cannot be interpreted as reliable negatives and IoU therefore becomes difficult to interpret as a primary metric. We also do not report FPR separately, as it is already reflected in BACC through the class-wise true negative rate / false positive behaviour. \revnew{Paired-sample t-tests on test-image F1 scores showed that the final tree-based configurations using $M_h$ significantly improved over their corresponding $M_{\ell}$ baselines for both $\loss_{twce}$ and $\loss_{wass+seg}$ ($p<0.0001$).}\revref{1}{2} For model performance on leaf node classes, we report F1 scores at $\tau_m$ for both losses. For $\loss_{wass+seg}$, the mean of F1 scores at $\tau_m$ based on $M_{\ell}$ and $M_h$ are $0.069$ and $0.073$, respectively. For $\loss_{twce}$, the results are $0.068$ and $0.037$, respectively.

While both loss outperforms the baseline, our results shows that $M_t$ performs similarly to $M_e$, suggesting that top-level edge weights have the greatest impact on performance when evaluating accumulated probabilities on corresponding nodes. For $\loss_{wass+seg}$, employing $M_h$ yields the best performance, surpassing the strong baseline that adapts CE loss to train only on the top-level node. \revdel{Demonstrating that an appropriate choice of $M$ can achieve state-of-the-art results on both top-level and leaf nodes.}\revmod{These results indicate that an appropriate choice of $M$ can improve performance over the evaluated baselines on both top-level and leaf nodes.}\revref{2}{3}\secondrevref{1}{4} For background / OOD detection, our findings exhibit trends similar to those reported in previous work on three medical image datasets \citep{wang2024oodsegoutofdistributiondetectionimage}. By removing pixels considered outliers, all methods gain further improvements.

Furthermore, it is possible to train the model using the pure Wasserstein distance defined in \eqref{eq:wassdistcrisp}. We refer to this loss as the Wasserstein loss, denoted by $\loss_{wass}$. The results for $\loss_{wass}$ based on $M_t$ are reported in \tabref{tab:cross_validation}, where we observe that it underperforms the baseline. Based on this observation, further investigation of the distance matrix $M$ was not conducted for $\loss_{wass}$.

\subsection{Qualitative results}\label{sec:qualitative_result}
\begin{figure}[htbp]
\centering
\setlength{\tabcolsep}{0pt}
\footnotesize
\begin{tabular}{c@{\hspace{2pt}} cccccc}
\vspace{2pt}
\rotatebox[origin=c]{90}{GT} &
\includegraphics[width=.15\linewidth,valign=m]{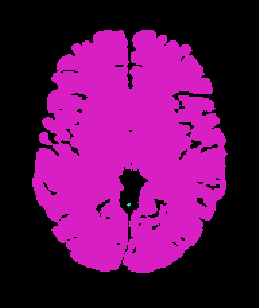} &
\includegraphics[width=.15\linewidth,valign=m]{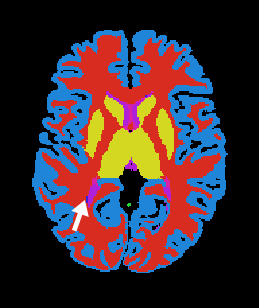} &
\includegraphics[width=.15\linewidth,valign=m]{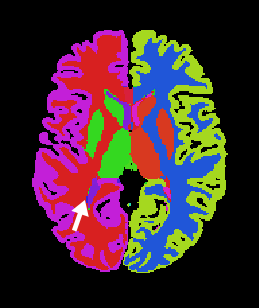} &
\includegraphics[width=.15\linewidth,valign=m]{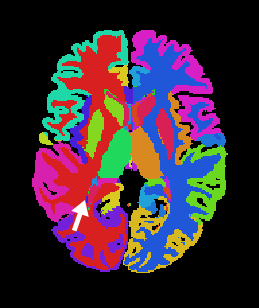} &
\includegraphics[width=.15\linewidth,valign=m]{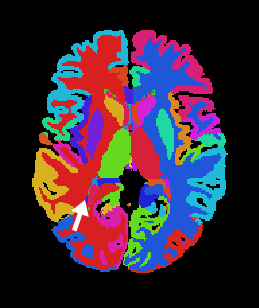} &
\includegraphics[width=.15\linewidth,valign=m]{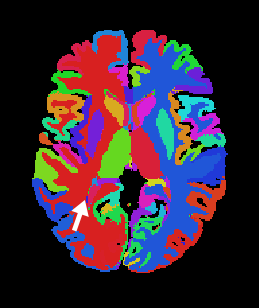}\\
\vspace{2pt}
\rotatebox[origin=c]{90}{$\loss_{seg}$} &
\includegraphics[width=.15\linewidth,valign=m]{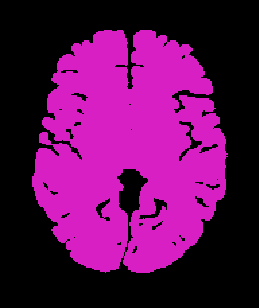} &
\includegraphics[width=.15\linewidth,valign=m]{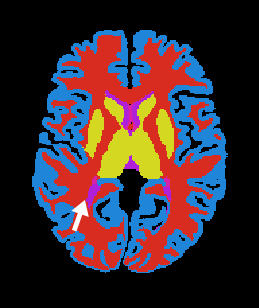} &
\includegraphics[width=.15\linewidth,valign=m]{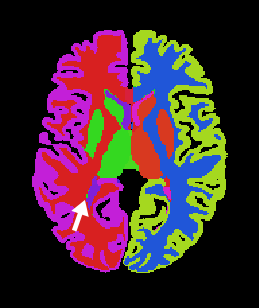} &
\includegraphics[width=.15\linewidth,valign=m]{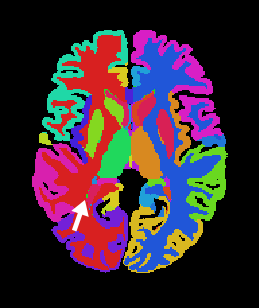} &
\includegraphics[width=.15\linewidth,valign=m]{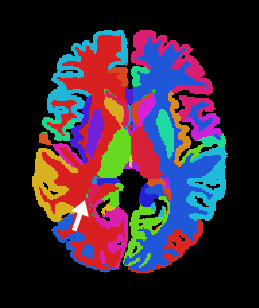} &
\includegraphics[width=.15\linewidth,valign=m]{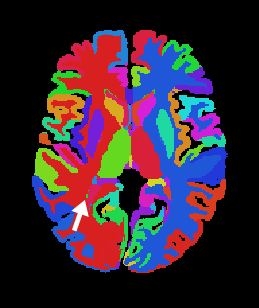}\\
\vspace{2pt}
\rotatebox[origin=c]{90}{$\loss_{wass+seg}$} &
\includegraphics[width=.15\linewidth,valign=m]{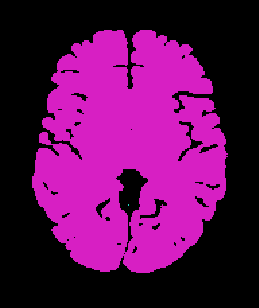} &
\includegraphics[width=.15\linewidth,valign=m]{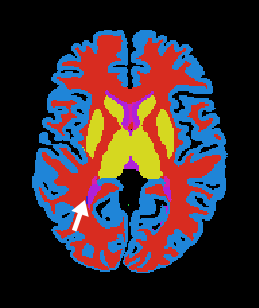} &
\includegraphics[width=.15\linewidth,valign=m]{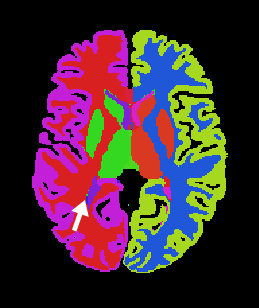} &
\includegraphics[width=.15\linewidth,valign=m]{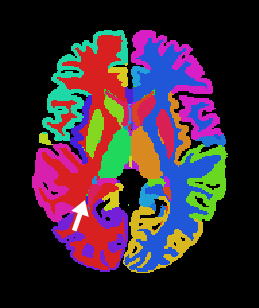} &
\includegraphics[width=.15\linewidth,valign=m]{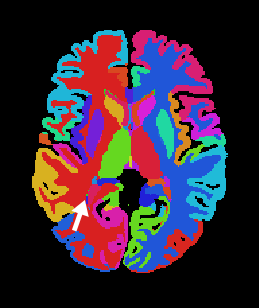} &
\includegraphics[width=.15\linewidth,valign=m]{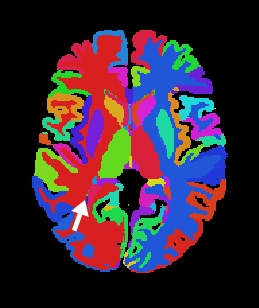}\\
\end{tabular}
\caption{Visual comparison of the baseline loss and the proposed Wasserstein-based loss on the AOMIC dataset. Each column shows the predicted segmentation masks at progressively finer levels of the label hierarchy. \revdel{The white arrow marks the challenging class \textit{non-WM hypointensities}, which the Wasserstein-based loss segments correctly, whereas the baseline fails to capture it.}\revmod{AOMIC uses a GIF-derived pseudo-ground-truth mask in this manuscript; the white arrow marks the challenging class \textit{non-WM hypointensities}, for which the Wasserstein-based loss better matches the pseudo-ground-truth mask whereas the baseline fails to capture it.}\robustrevref{1}{3}}
\label{fig:qualitative_result_bp}
\end{figure}

\begin{figure}[htbp]
\centering
\setlength\tabcolsep{0pt}
\footnotesize
\begin{tabular}{c c c c c c c c}
& \textbf{sRGB}
& \makecell[c]{\textbf{Sparsely}\\[-3pt]\textbf{ annotated}\\[-3pt]\textbf{ground truth}}
& \textbf{$\loss_{twce+seg}^{M_{\ell}}(\tau_{0})$}
& \textbf{$\loss_{twce+seg}^{M_{\ell}}$}
& \textbf{$\loss_{twce+seg}^{M_{t}}$}
& \textbf{$\loss_{wass+seg}^{M_{t}}$}
& \textbf{$\loss_{wass+seg}^{M_{h}}$}\\
& \includegraphics[height=1.32cm,valign=t]{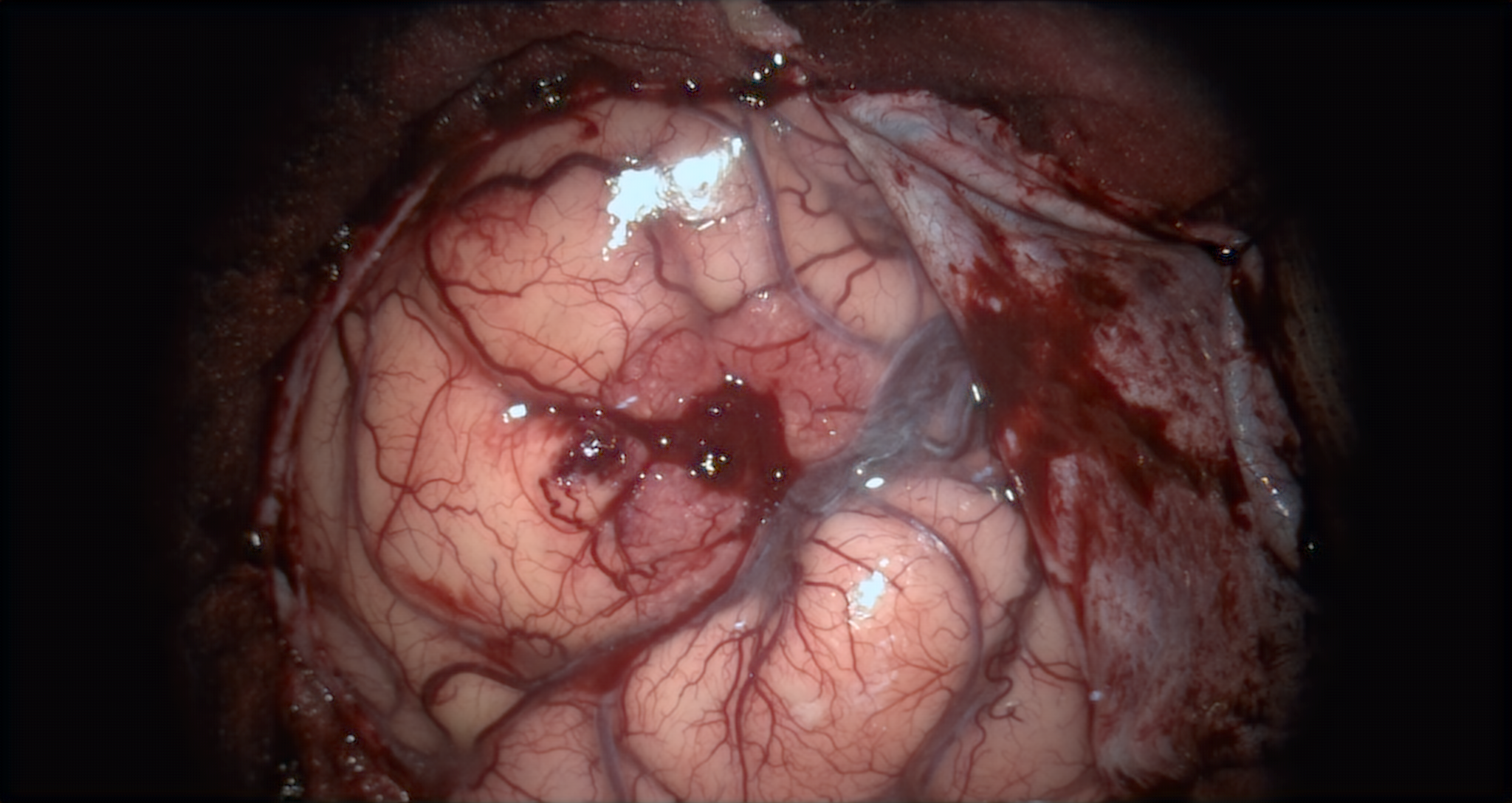}
& \includegraphics[height=1.32cm,valign=t]{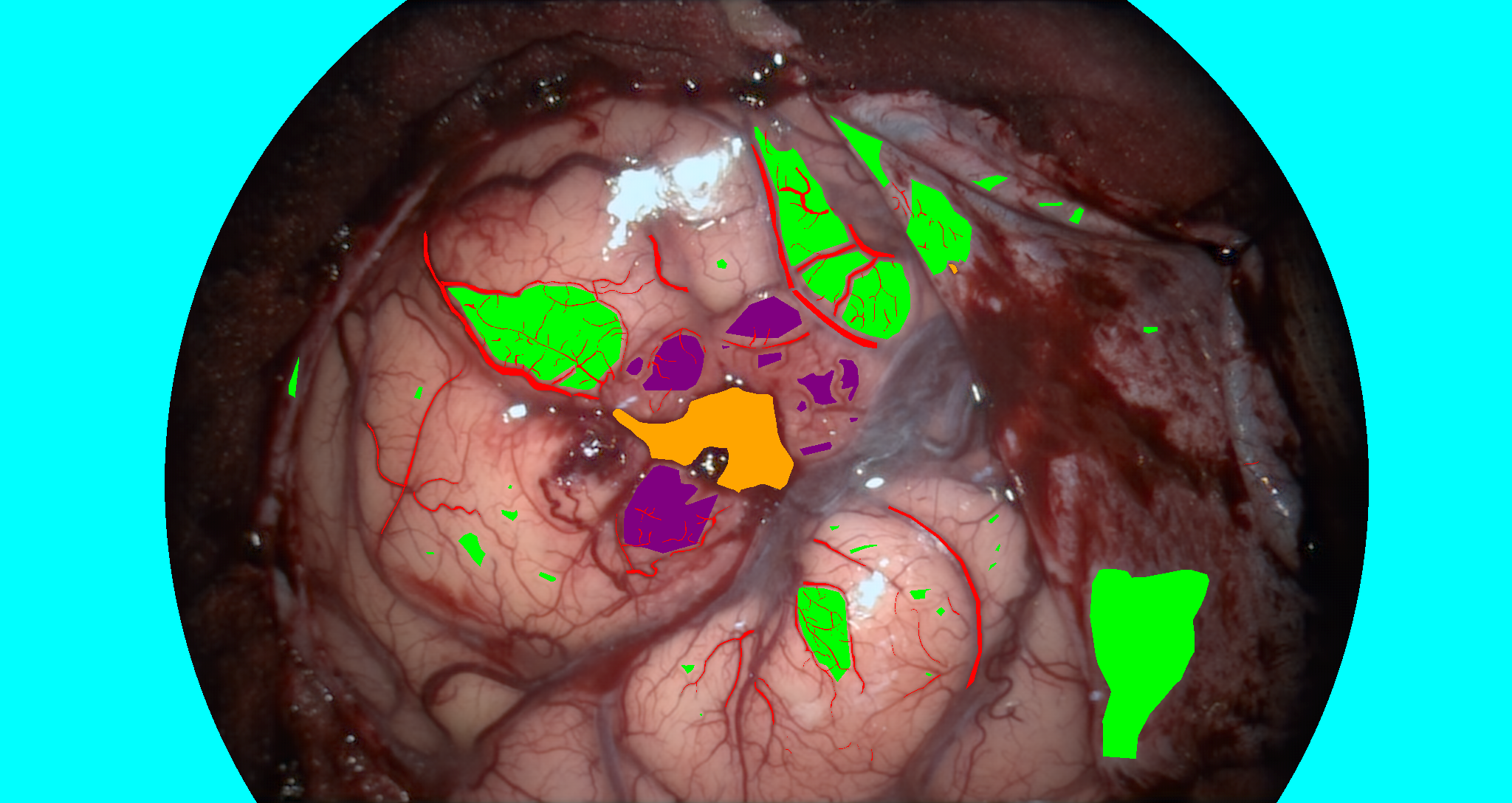}
& \includegraphics[height=1.32cm,valign=t]{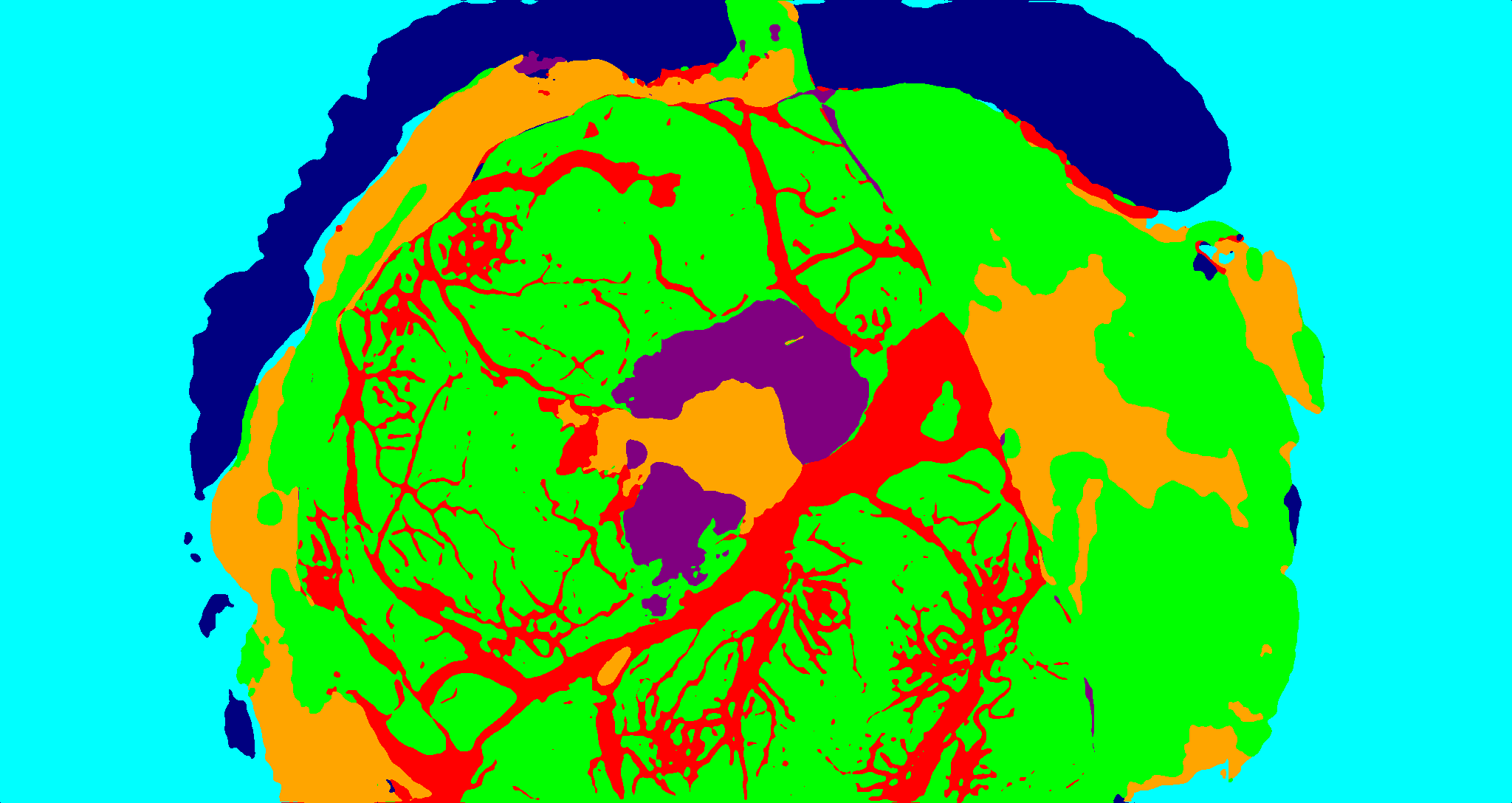}
& \includegraphics[height=1.32cm,valign=t]{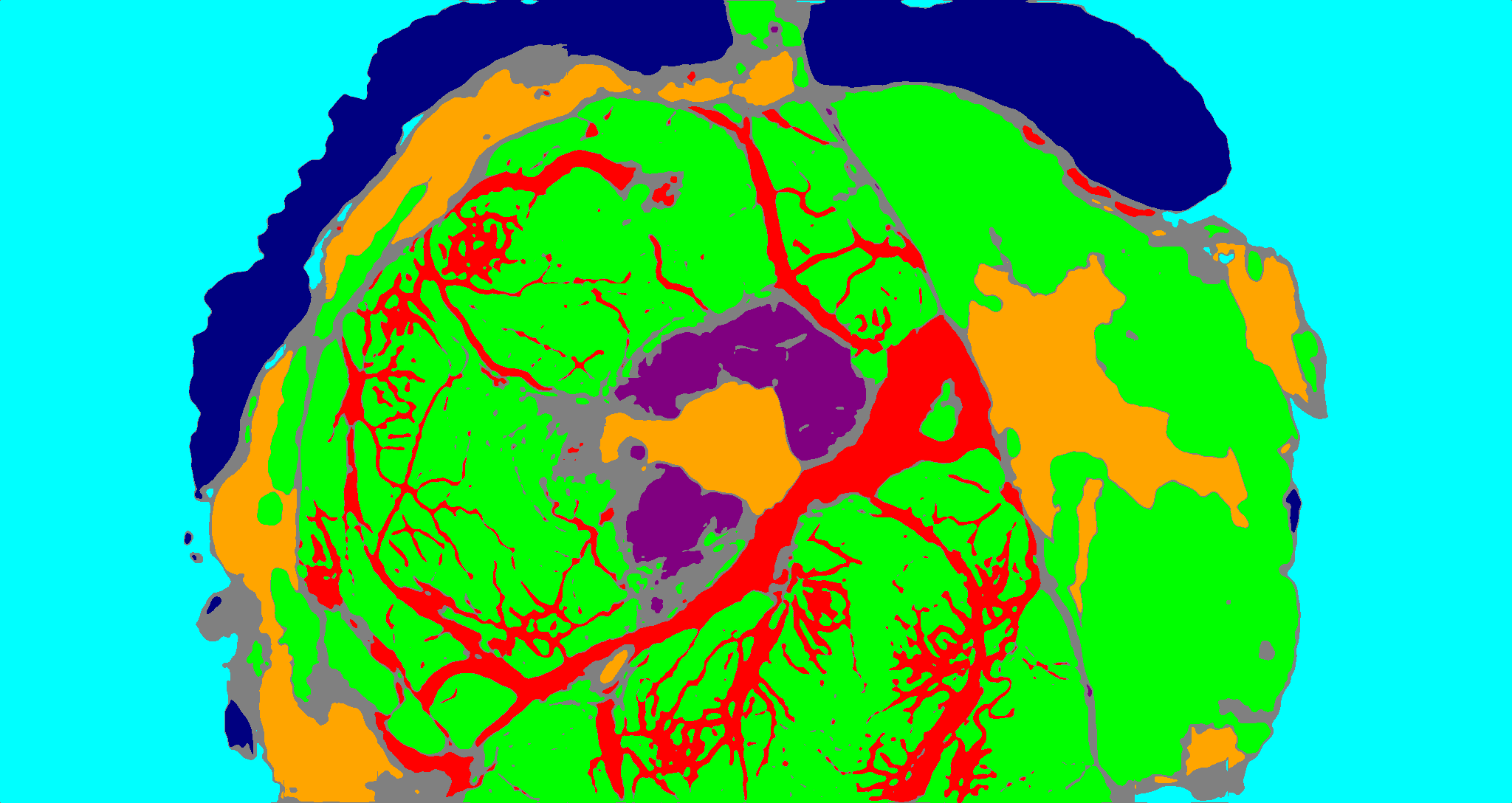}
& \includegraphics[height=1.32cm,valign=t]{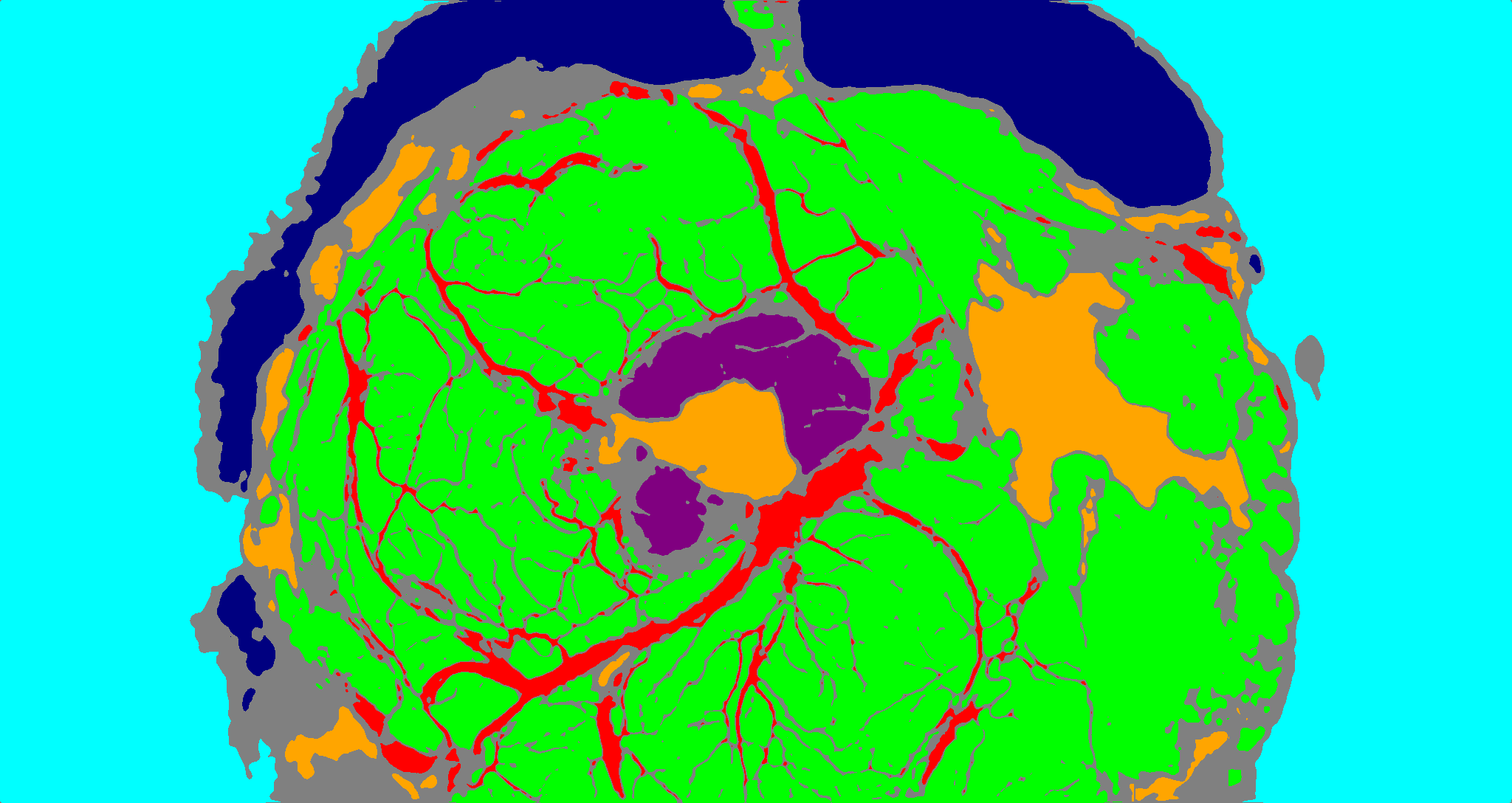}
& \includegraphics[height=1.32cm,valign=t]{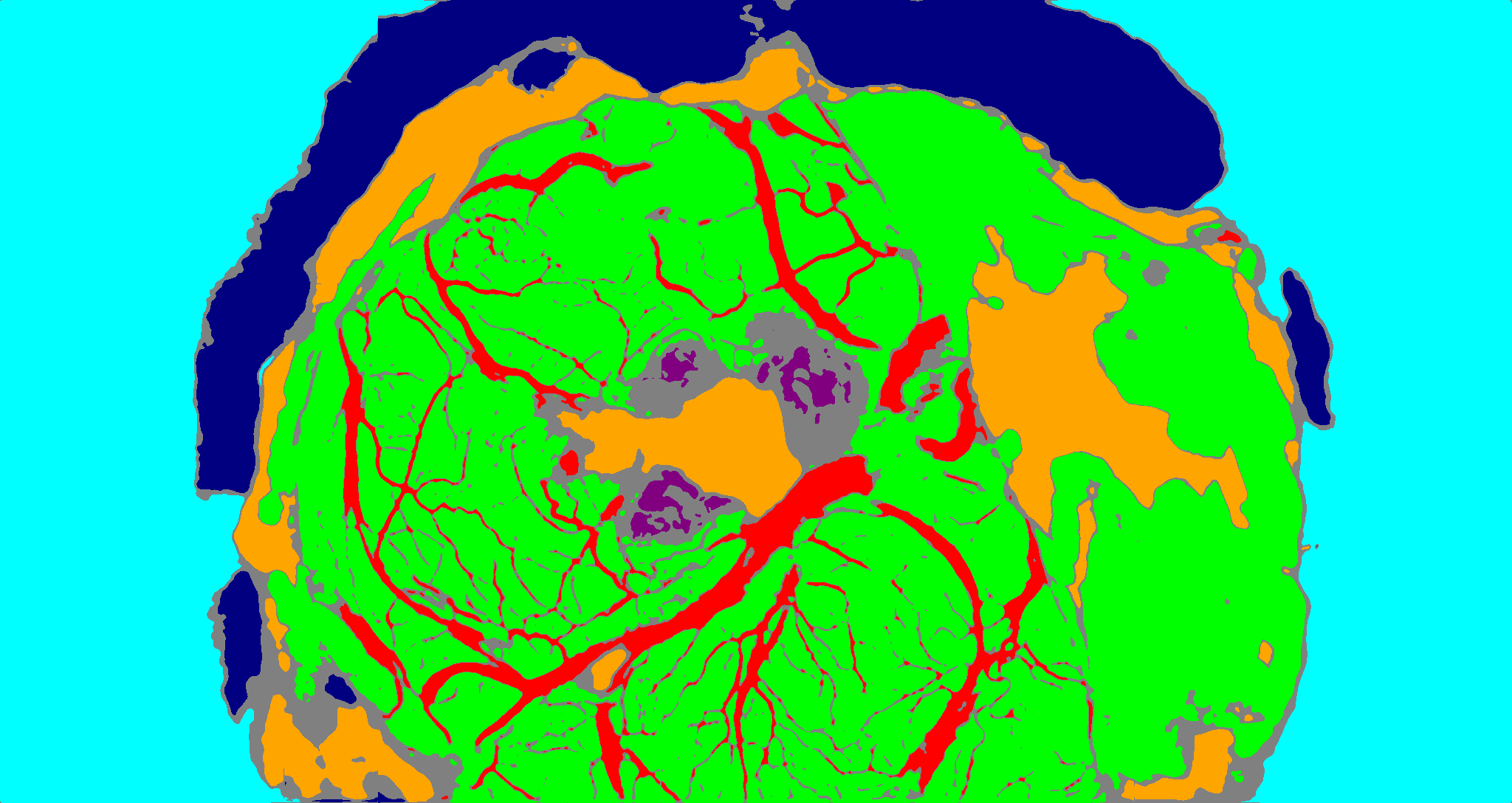}
& \includegraphics[height=1.32cm,valign=t]{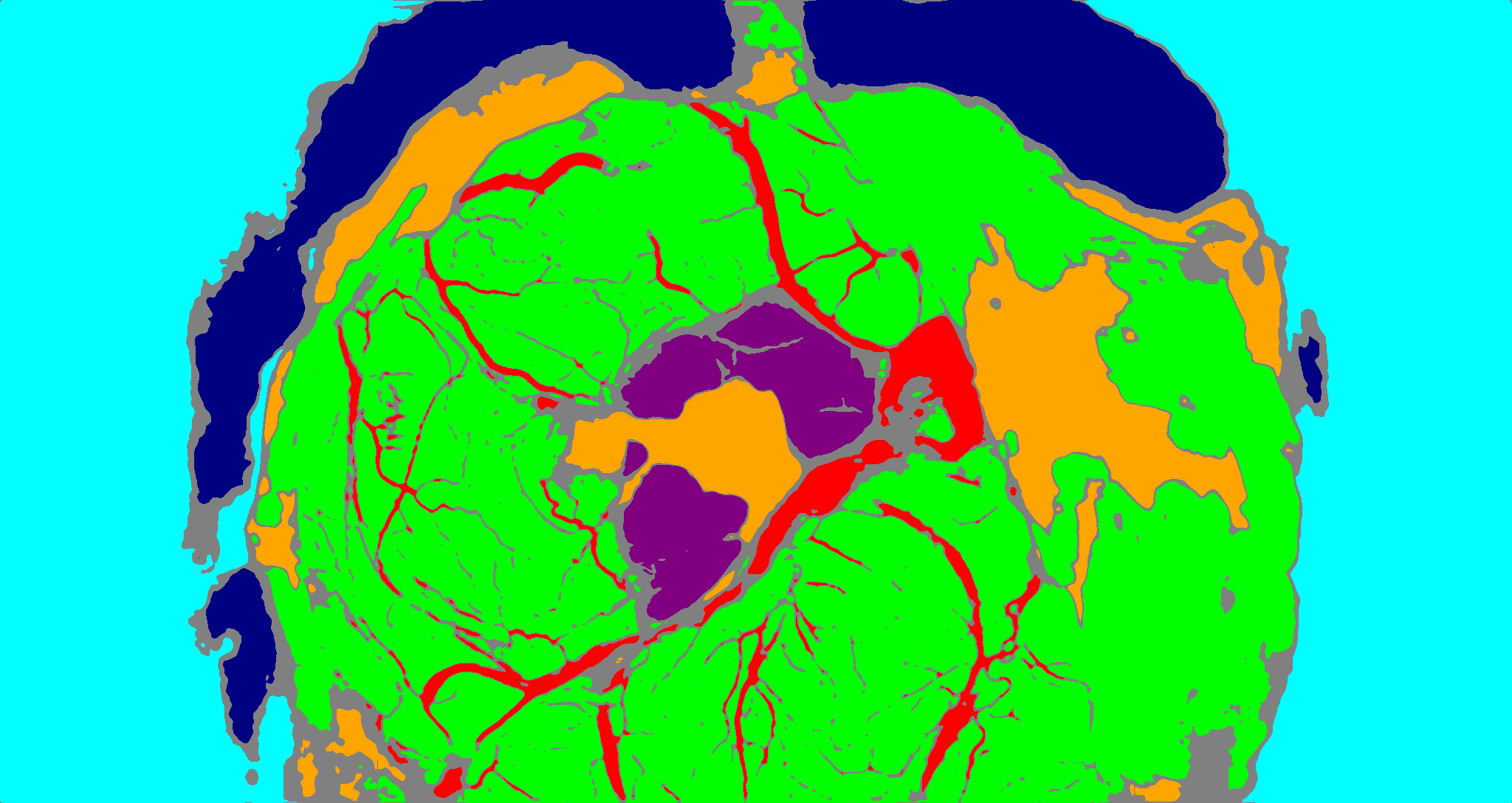}\\
& \includegraphics[height=1.32cm,valign=t]{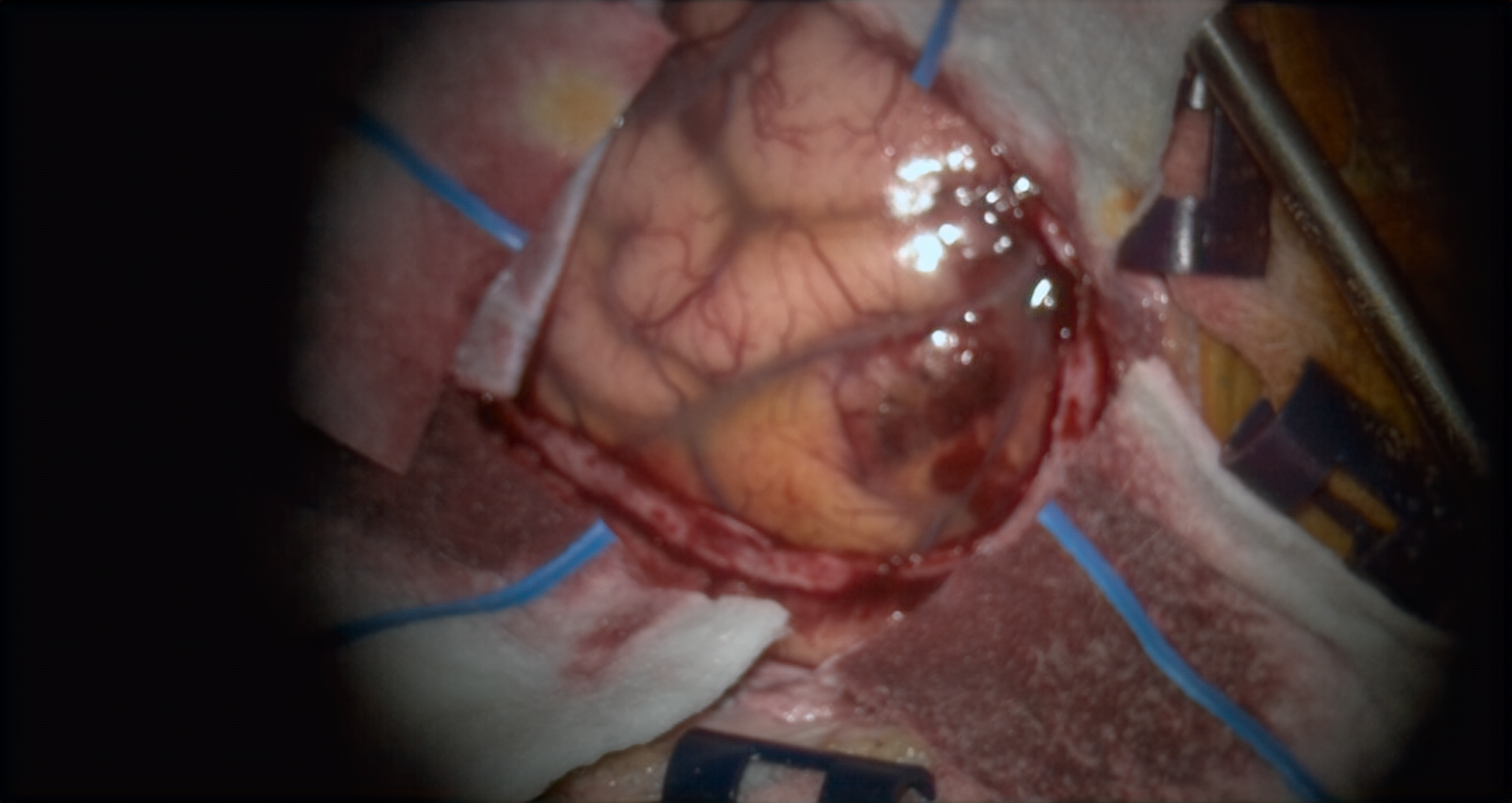}
& \includegraphics[height=1.32cm,valign=t]{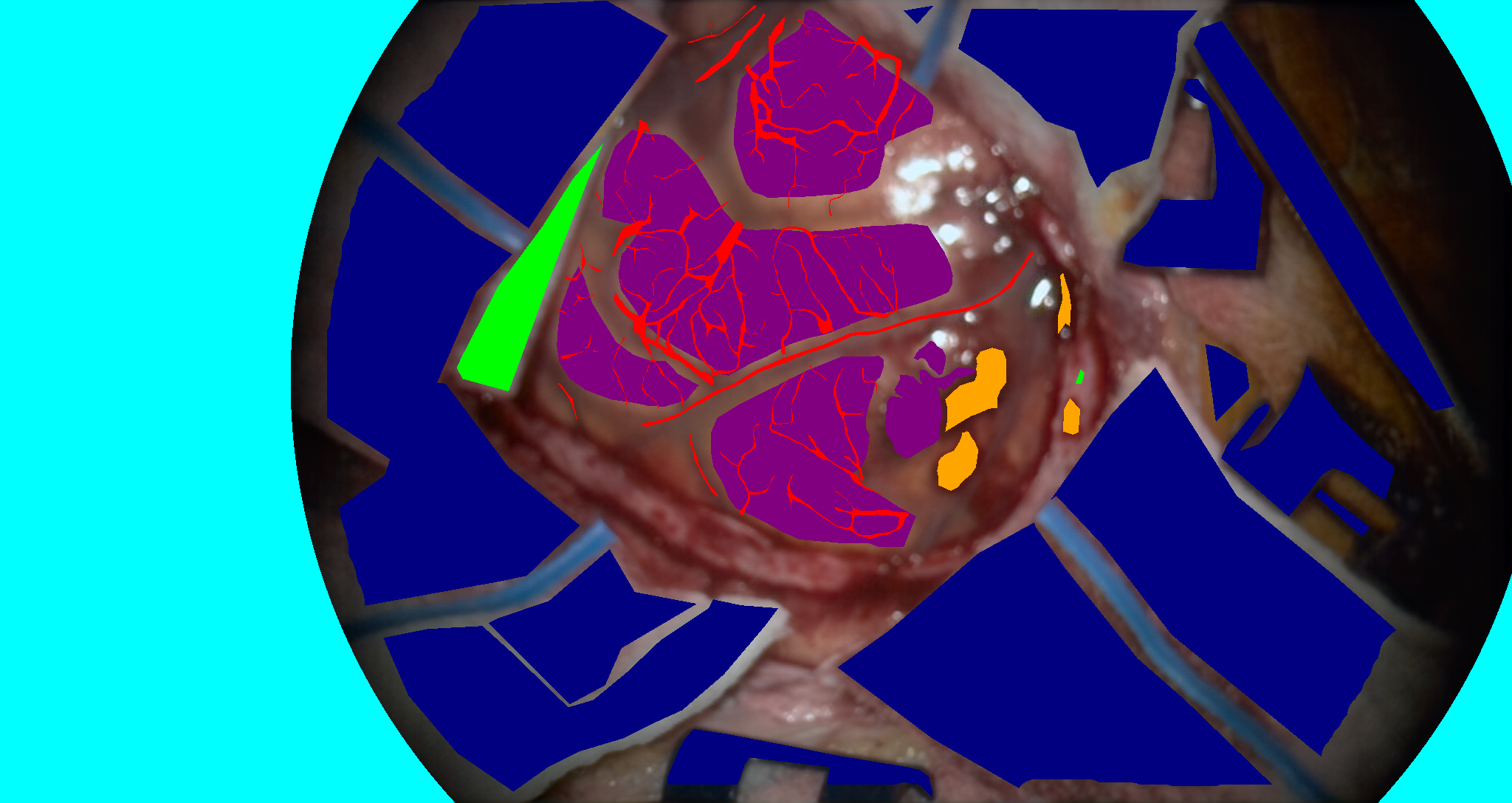}
& \includegraphics[height=1.32cm,valign=t]{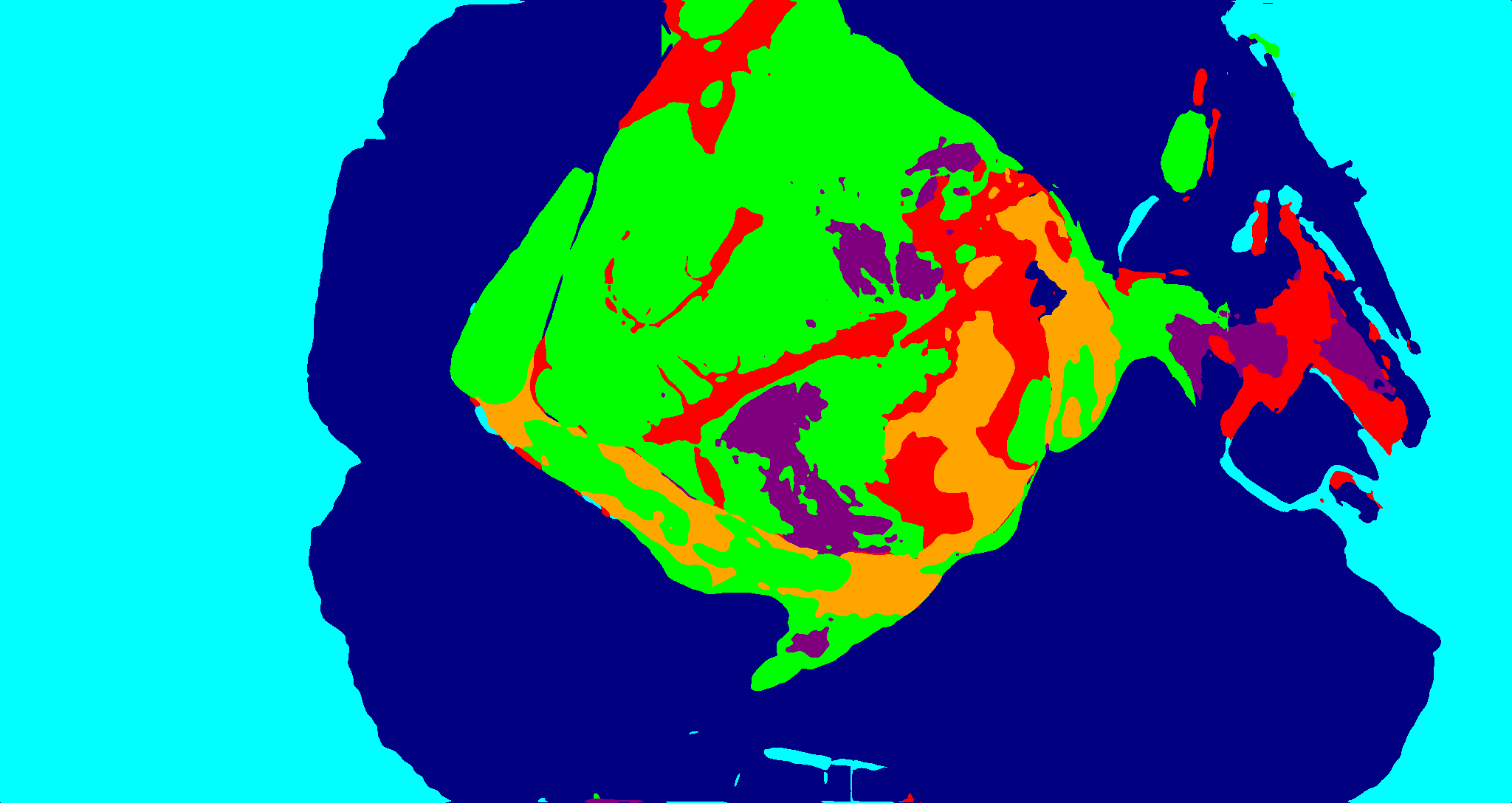}
& \includegraphics[height=1.32cm,valign=t]{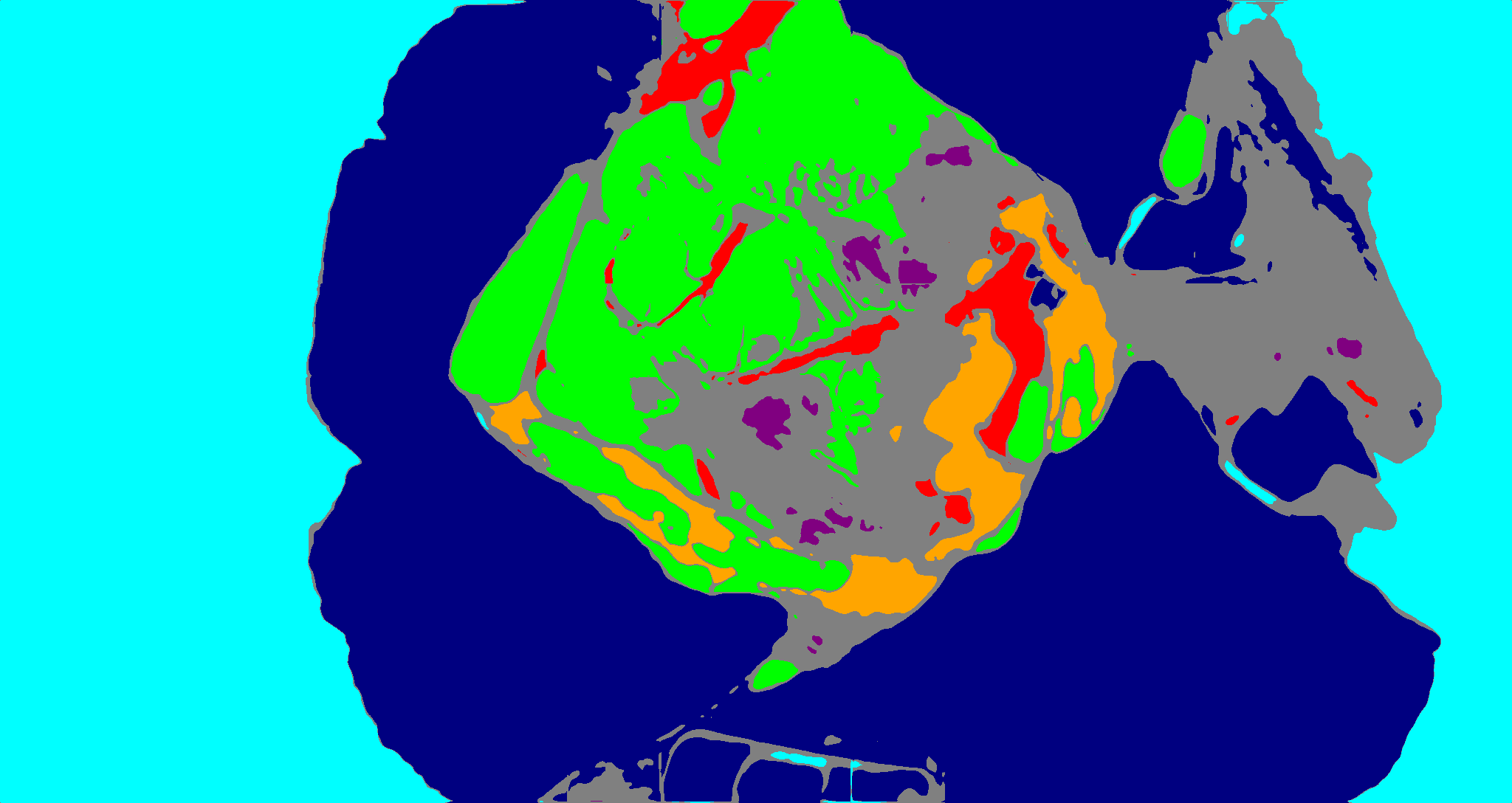}
& \includegraphics[height=1.32cm,valign=t]{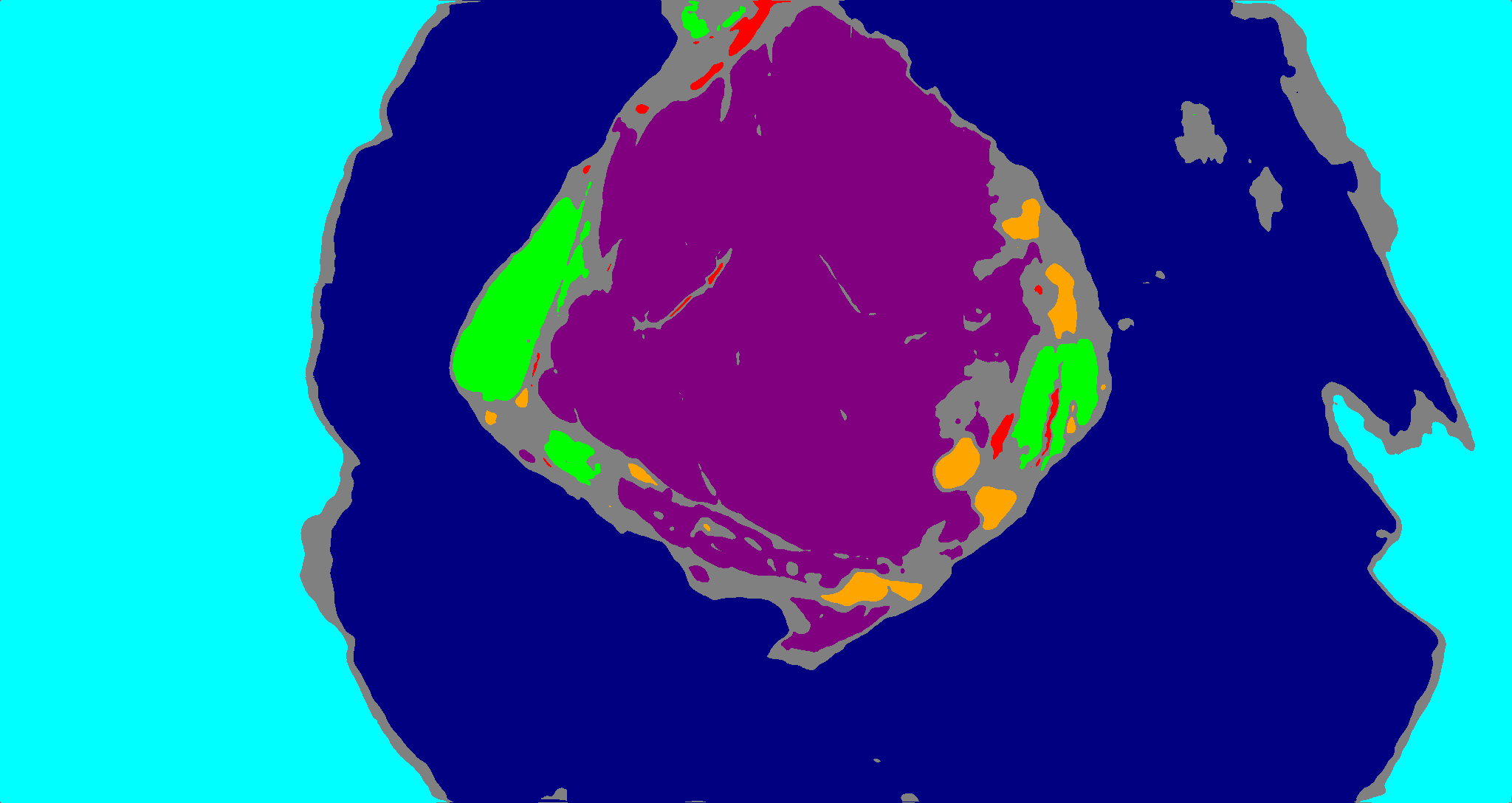}
& \includegraphics[height=1.32cm,valign=t]{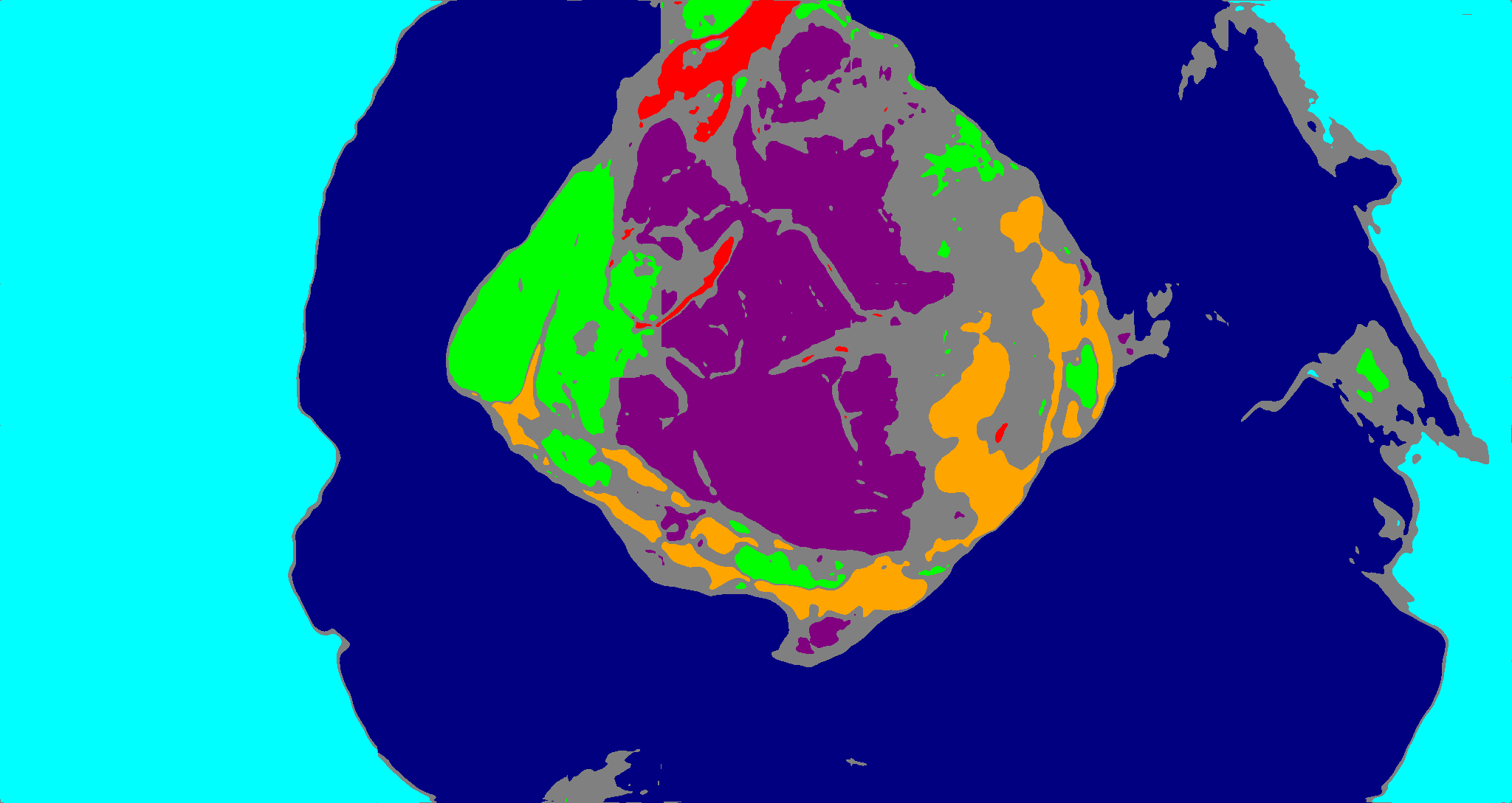}
& \includegraphics[height=1.32cm,valign=t]{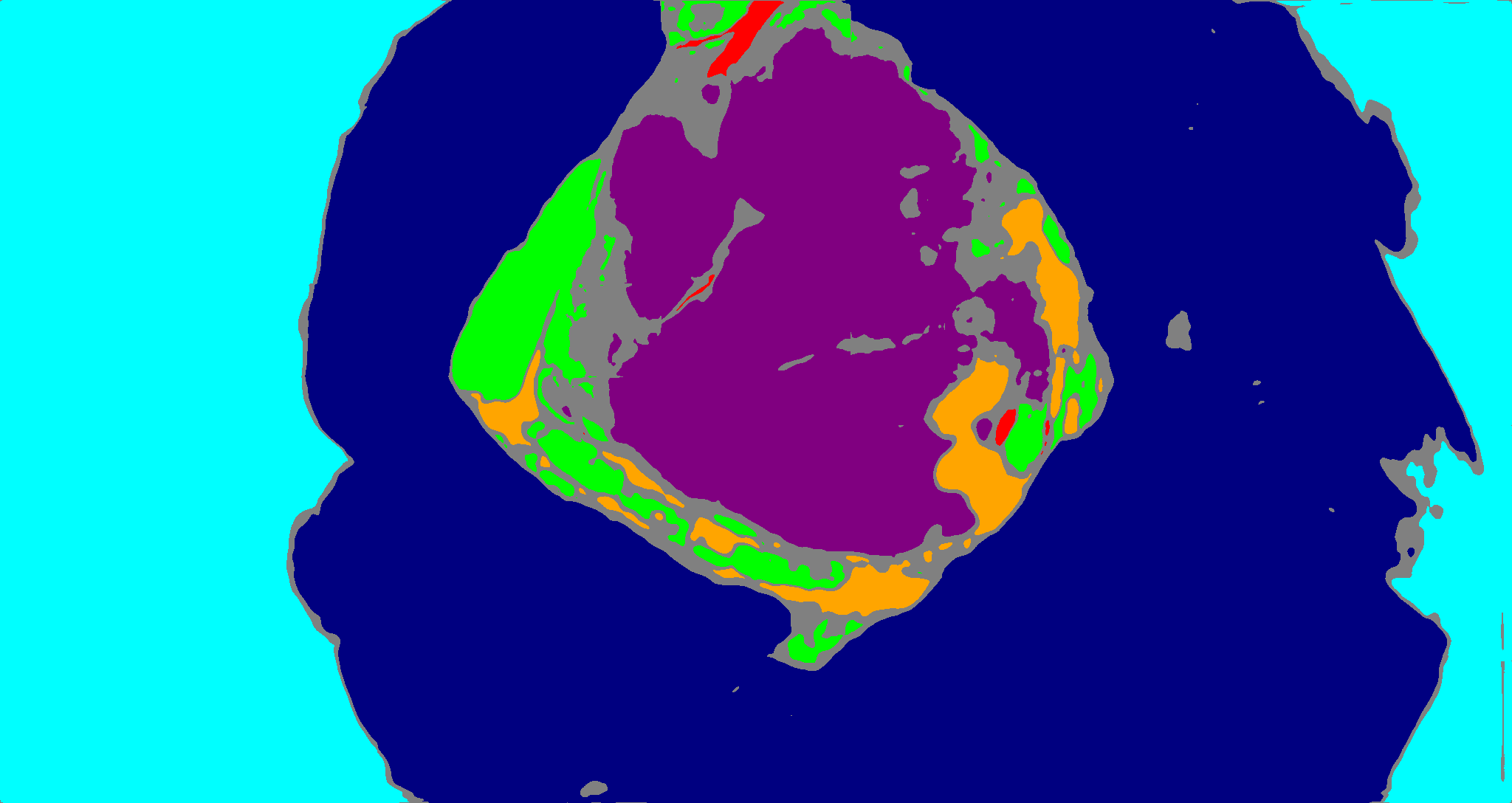}\\
& \includegraphics[height=1.32cm,valign=t]{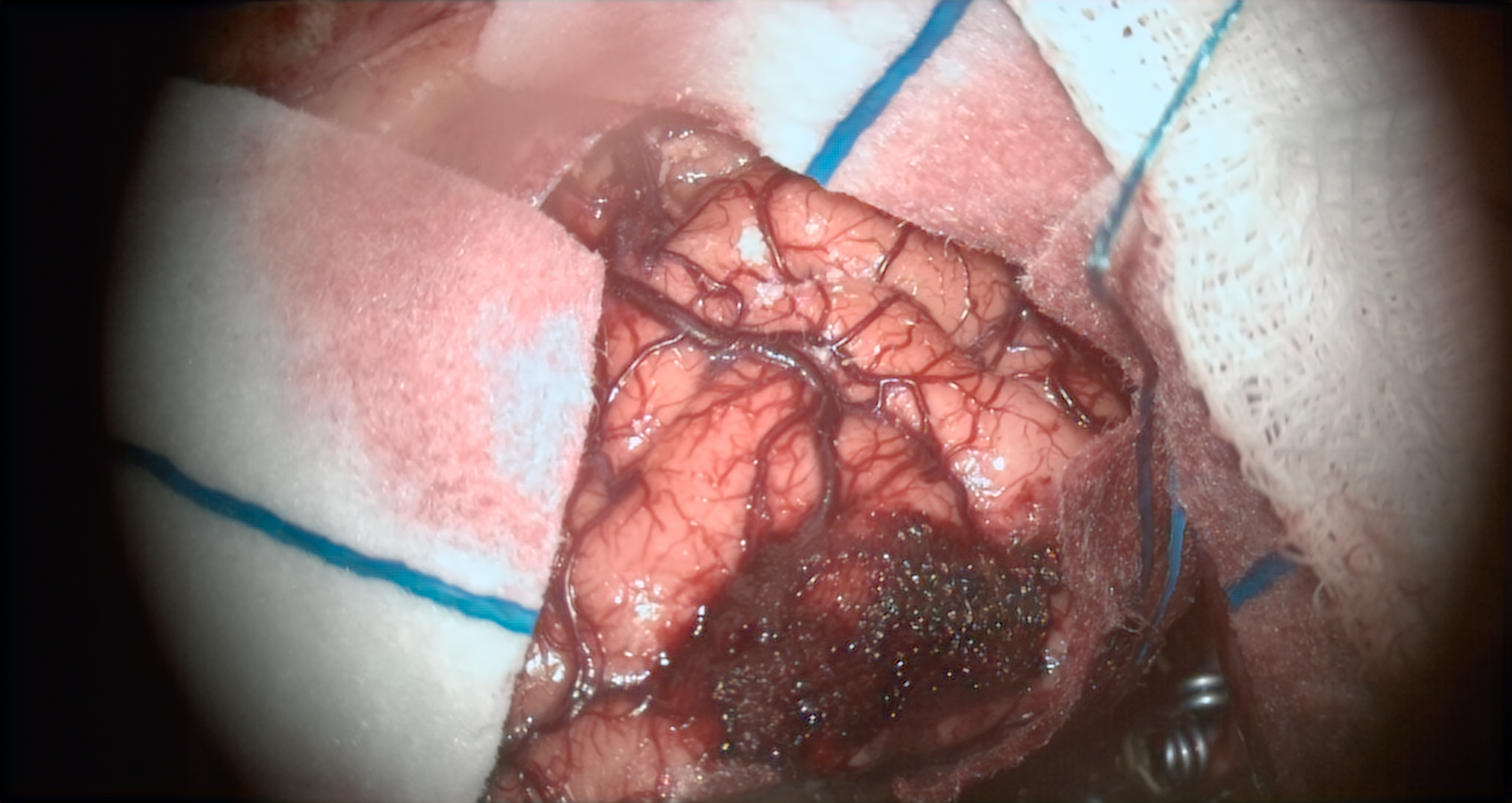}
& \includegraphics[height=1.32cm,valign=t]{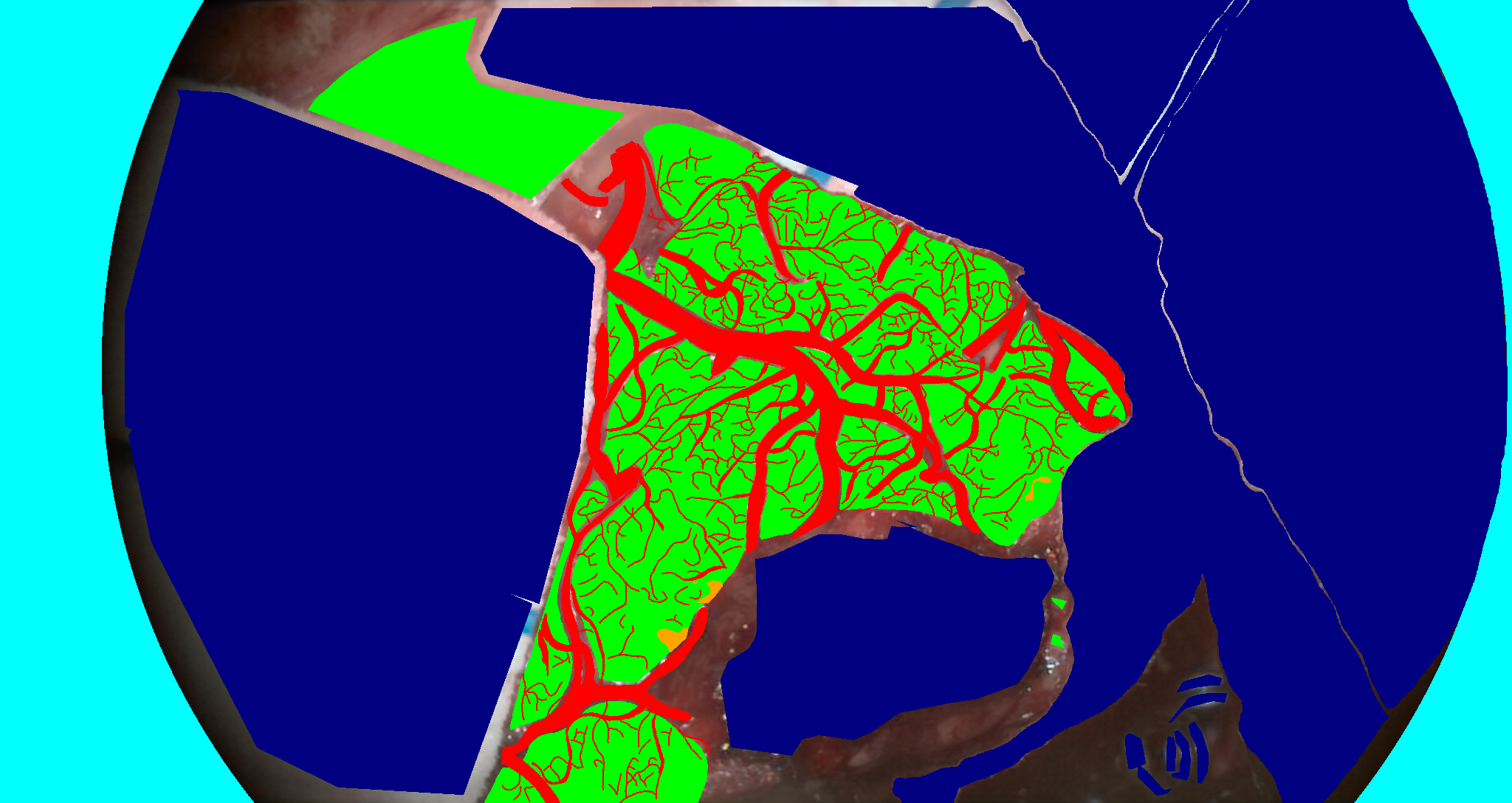}
& \includegraphics[height=1.32cm,valign=t]{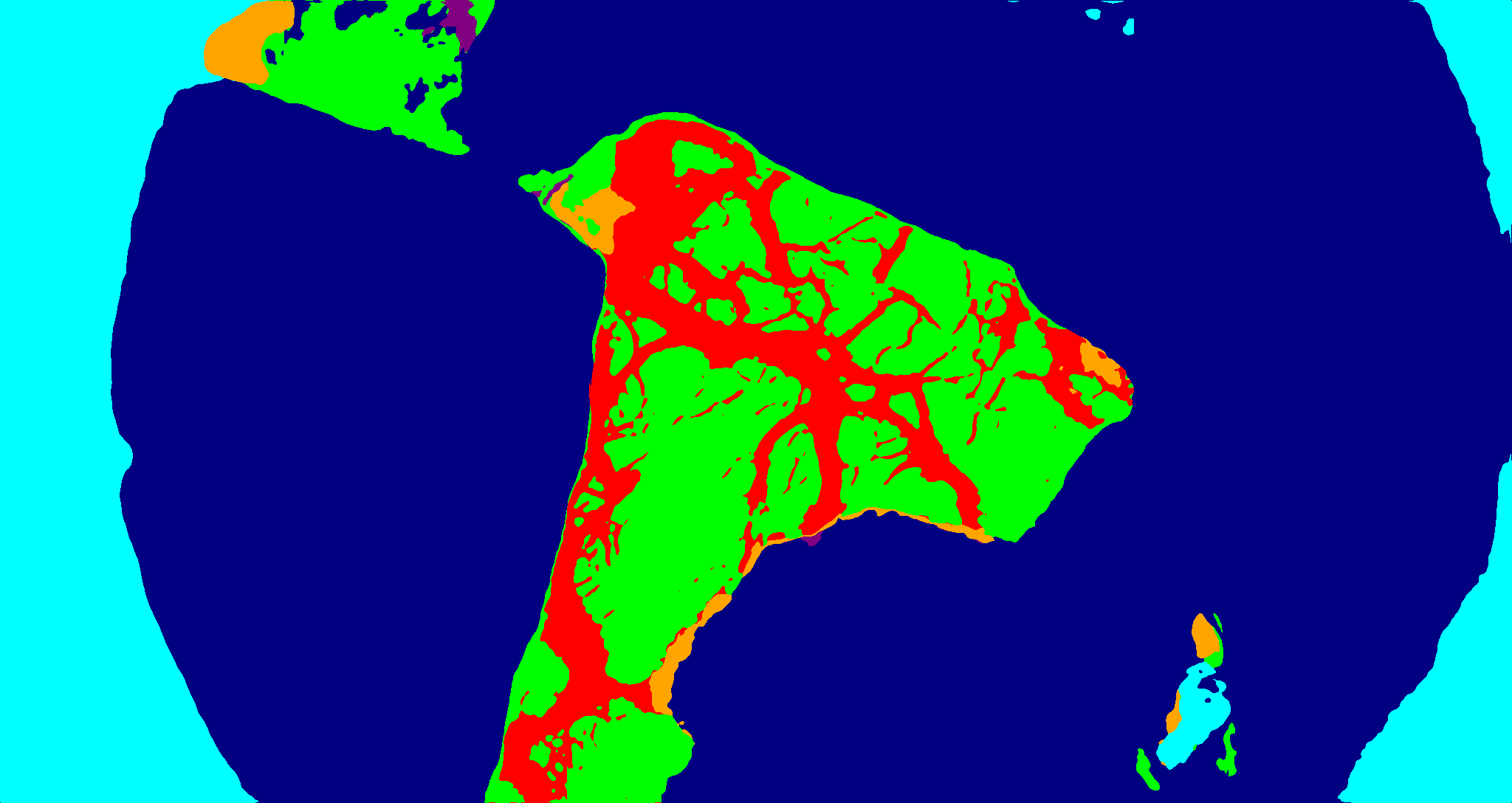}
& \includegraphics[height=1.32cm,valign=t]{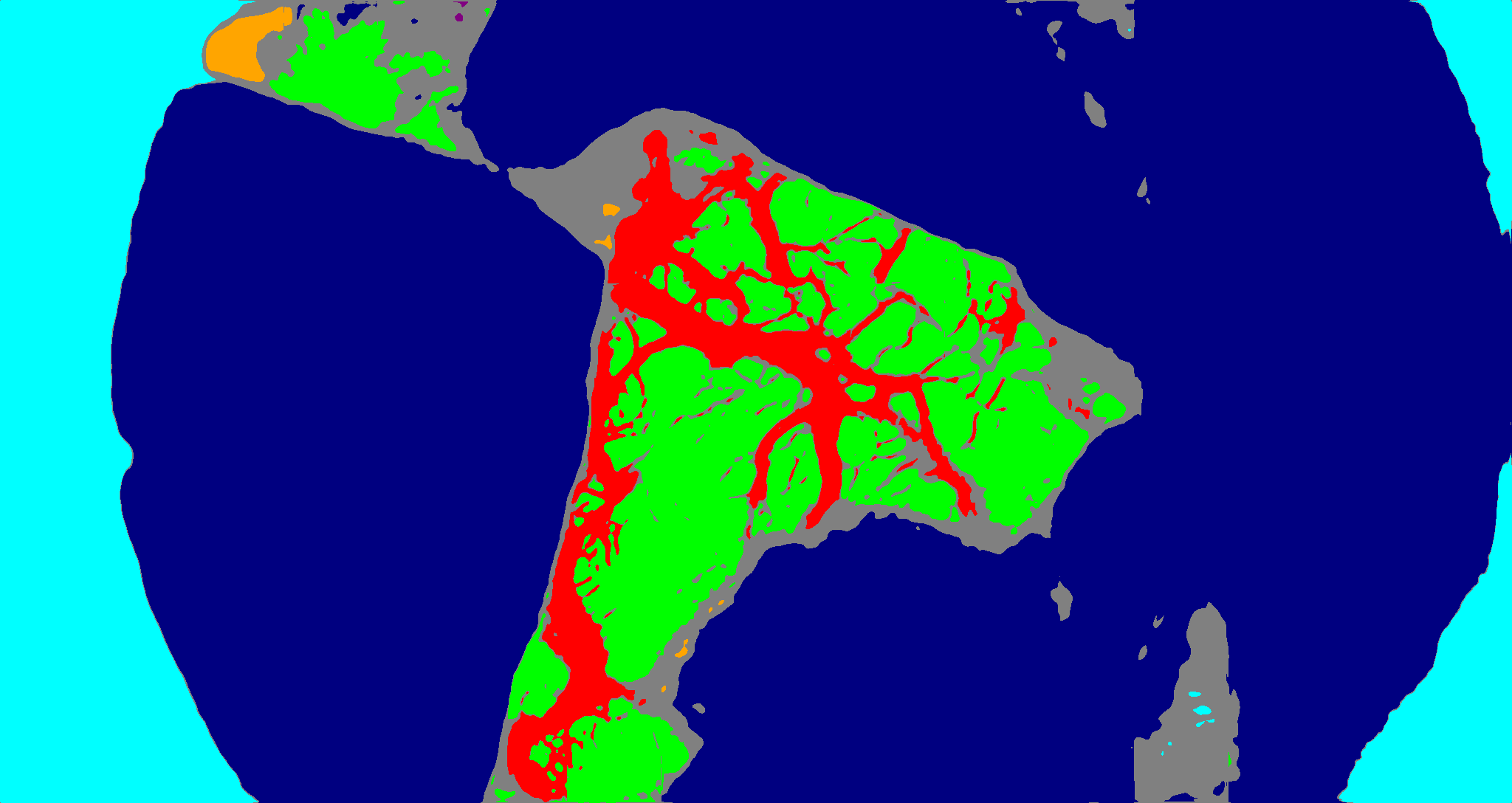}
& \includegraphics[height=1.32cm,valign=t]{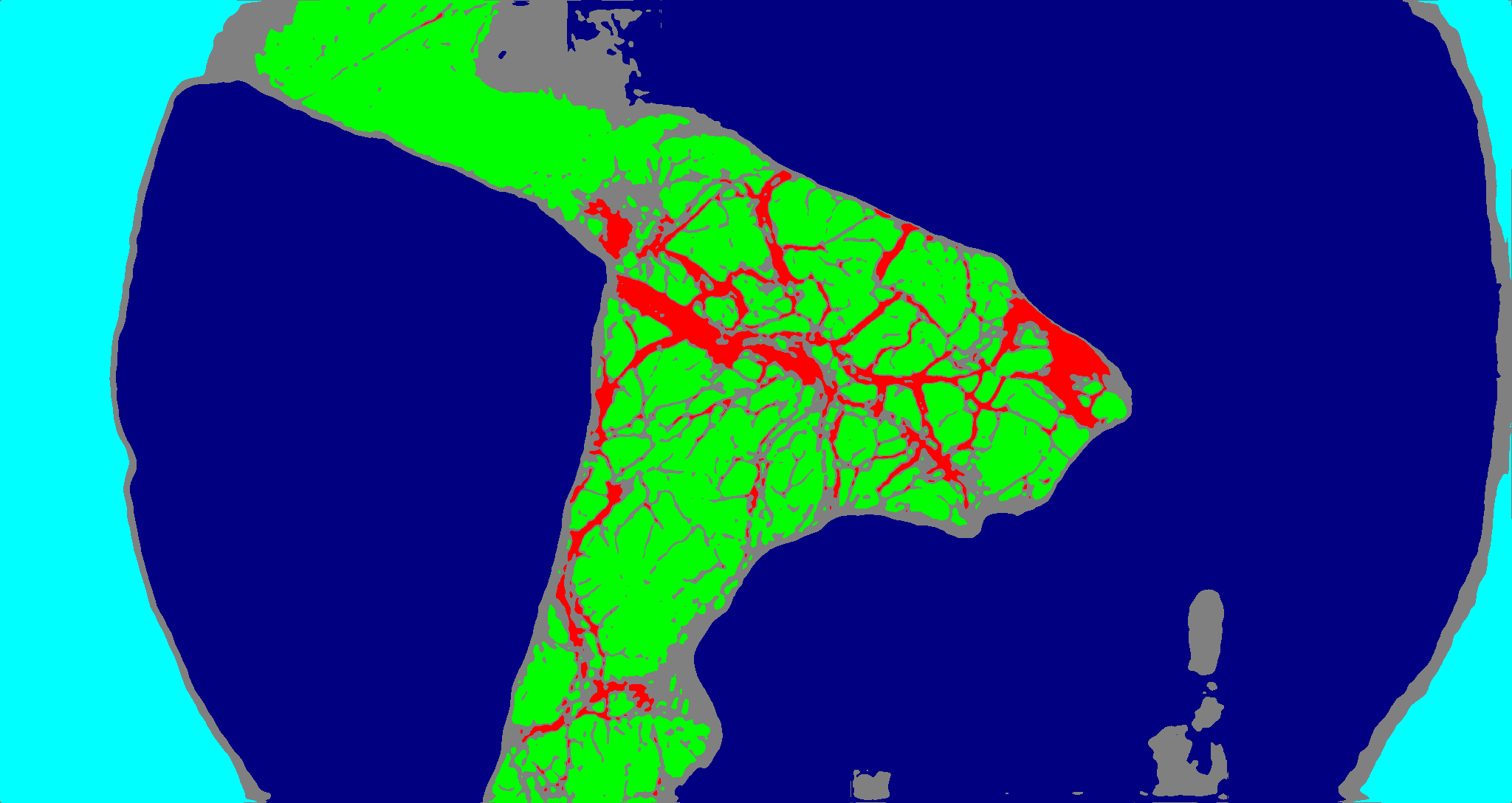}
& \includegraphics[height=1.32cm,valign=t]{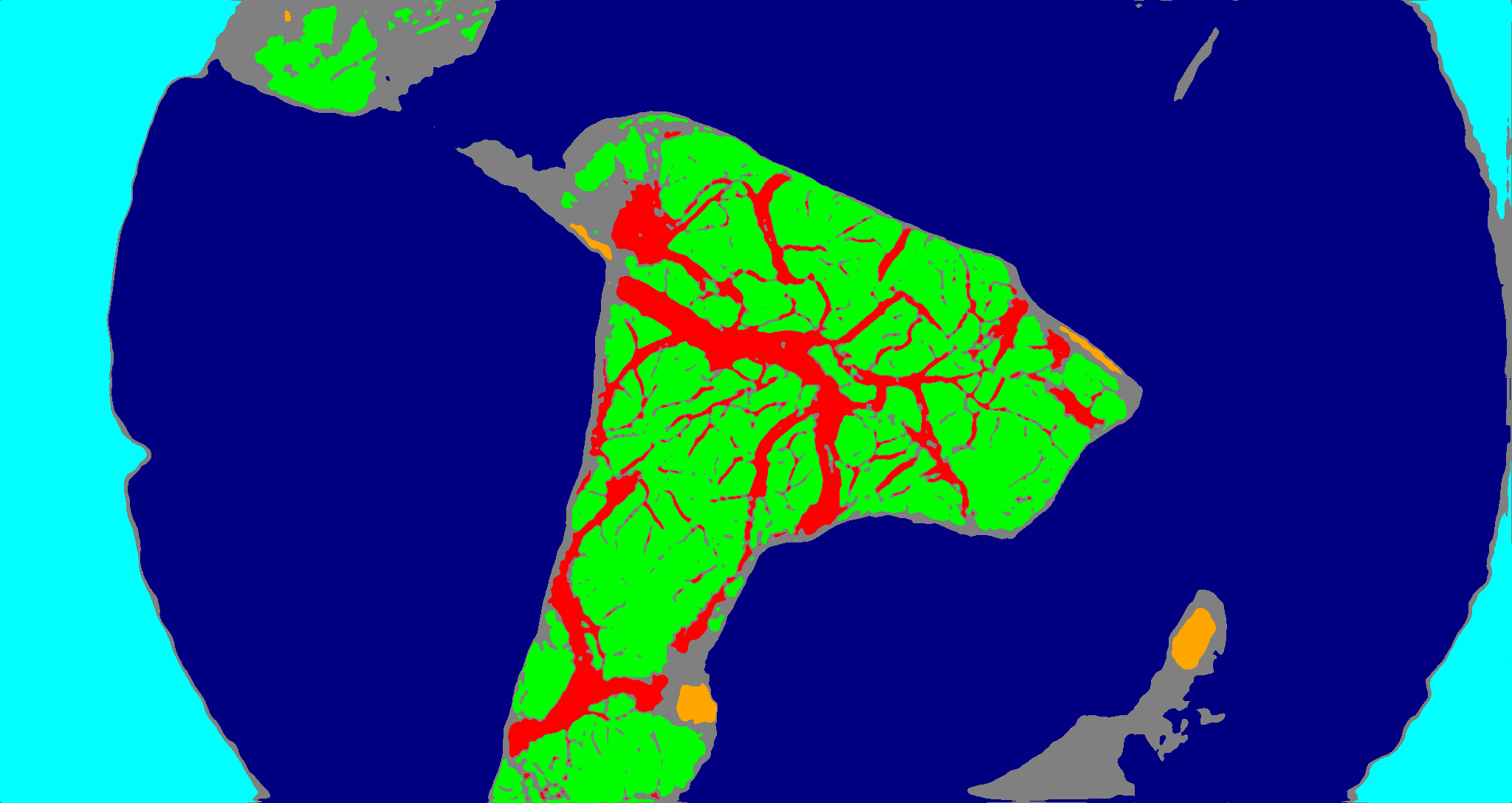}
& \includegraphics[height=1.32cm,valign=t]{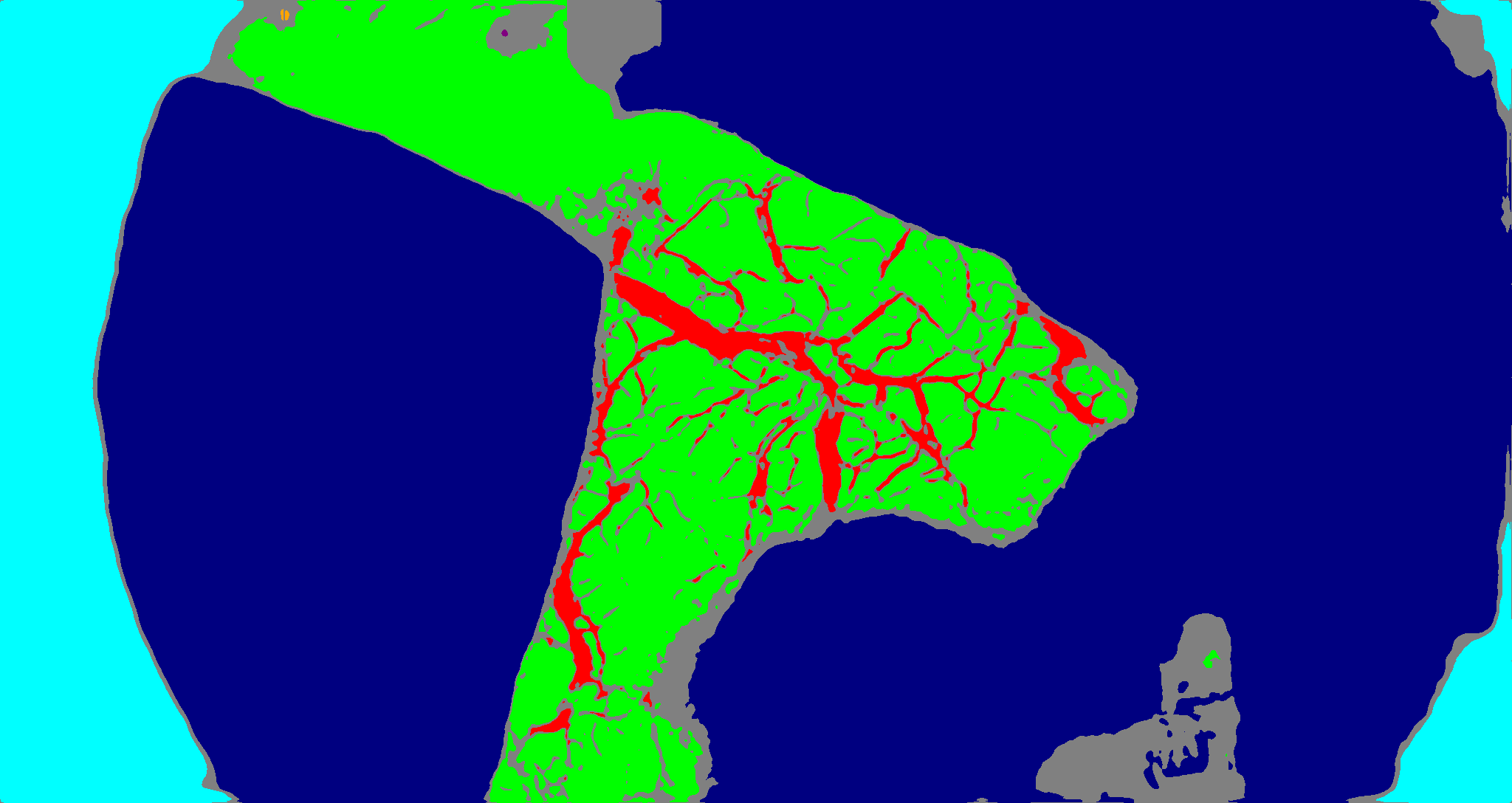}\\
& \includegraphics[height=1.32cm,valign=t]{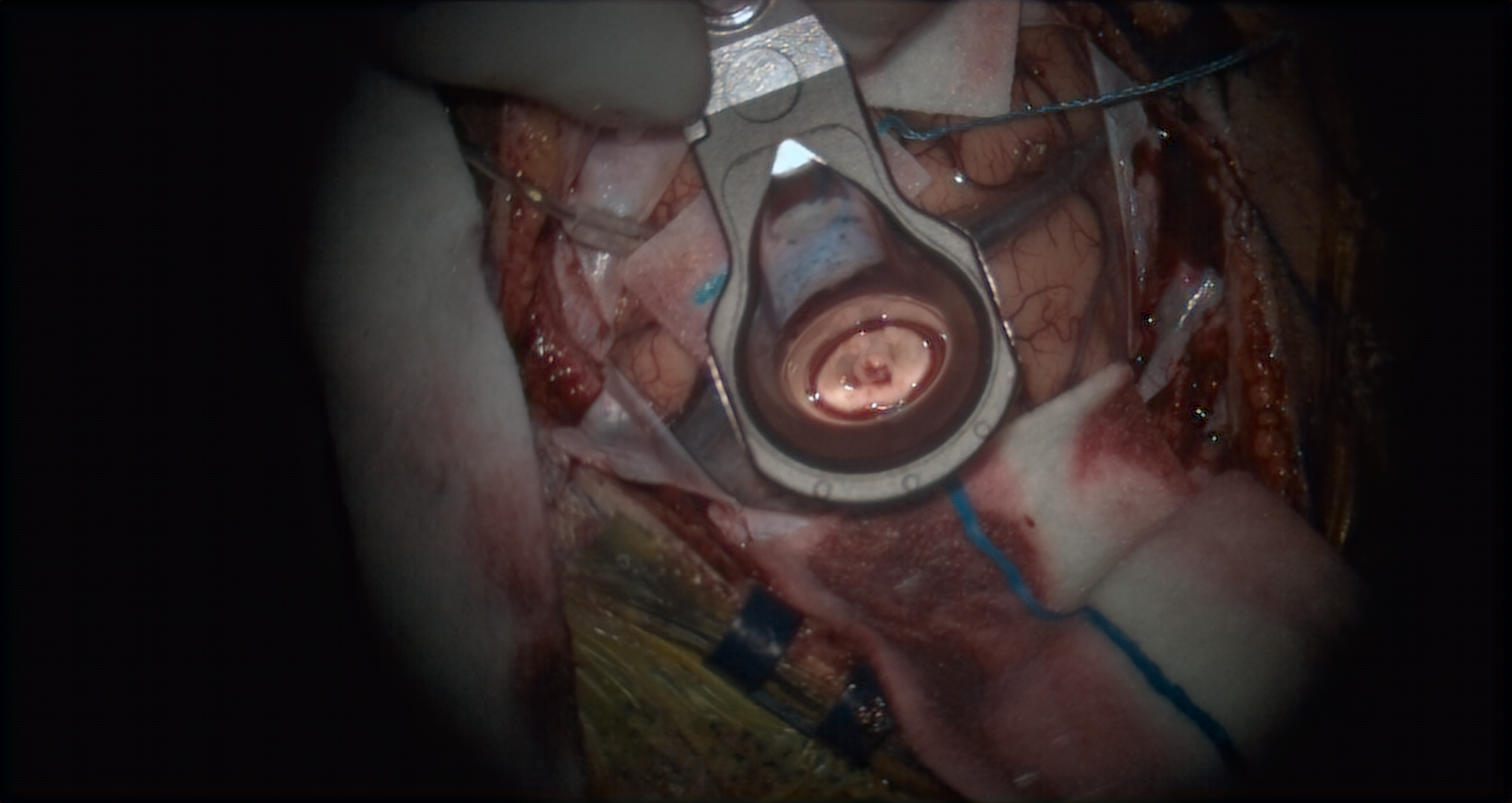}
& \includegraphics[height=1.32cm,valign=t]{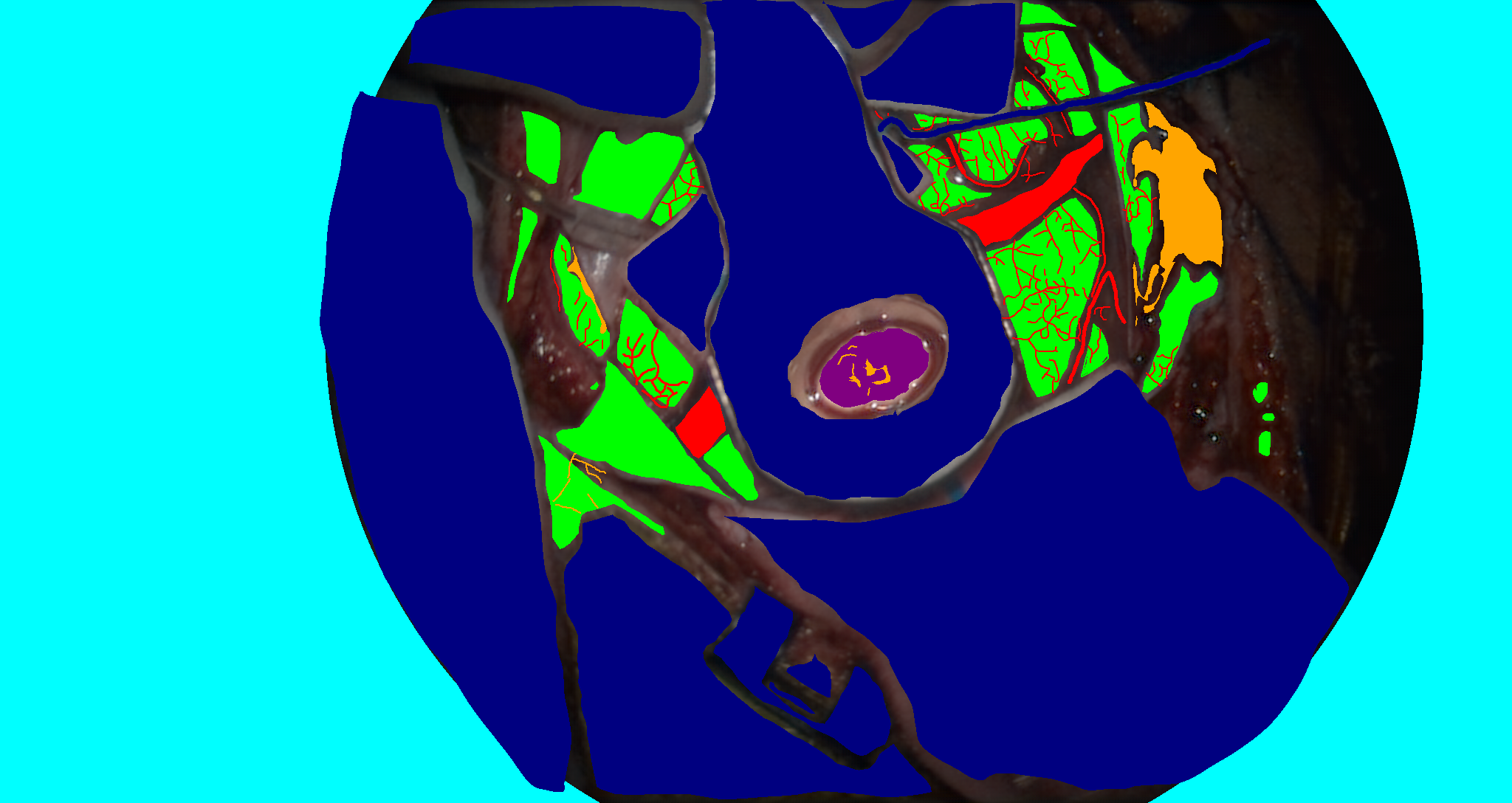}
& \includegraphics[height=1.32cm,valign=t]{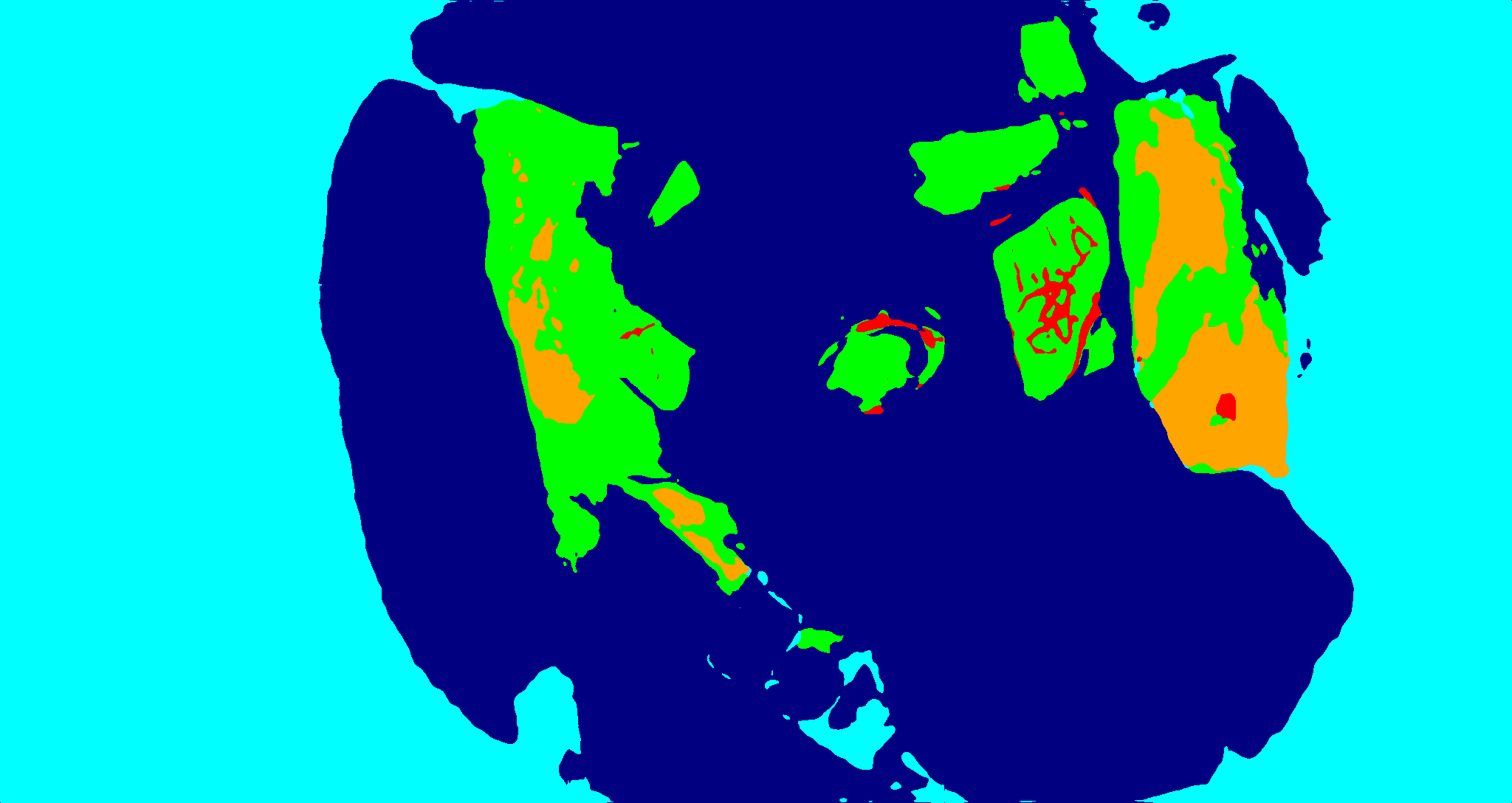}
& \includegraphics[height=1.32cm,valign=t]{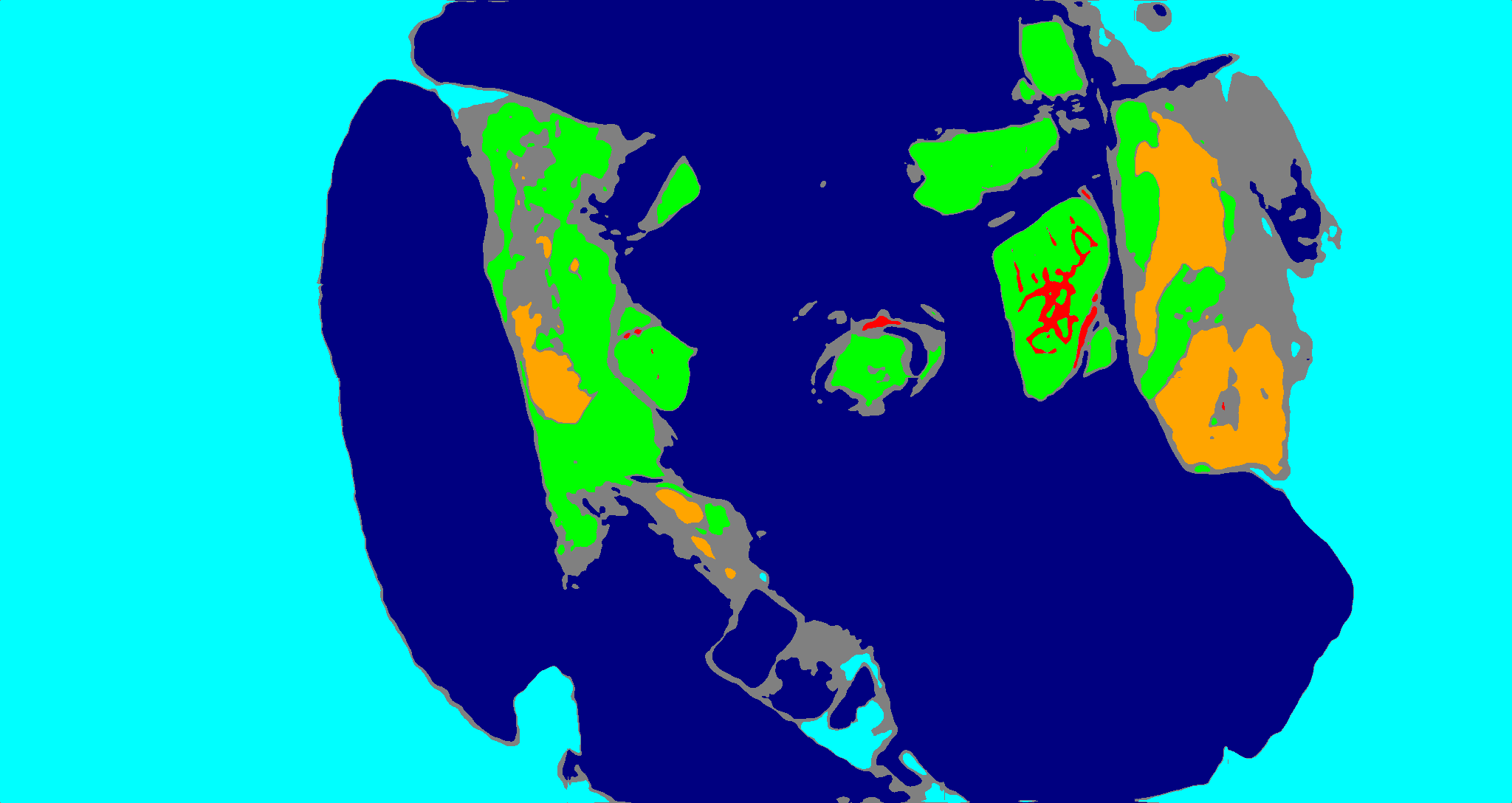}
& \includegraphics[height=1.32cm,valign=t]{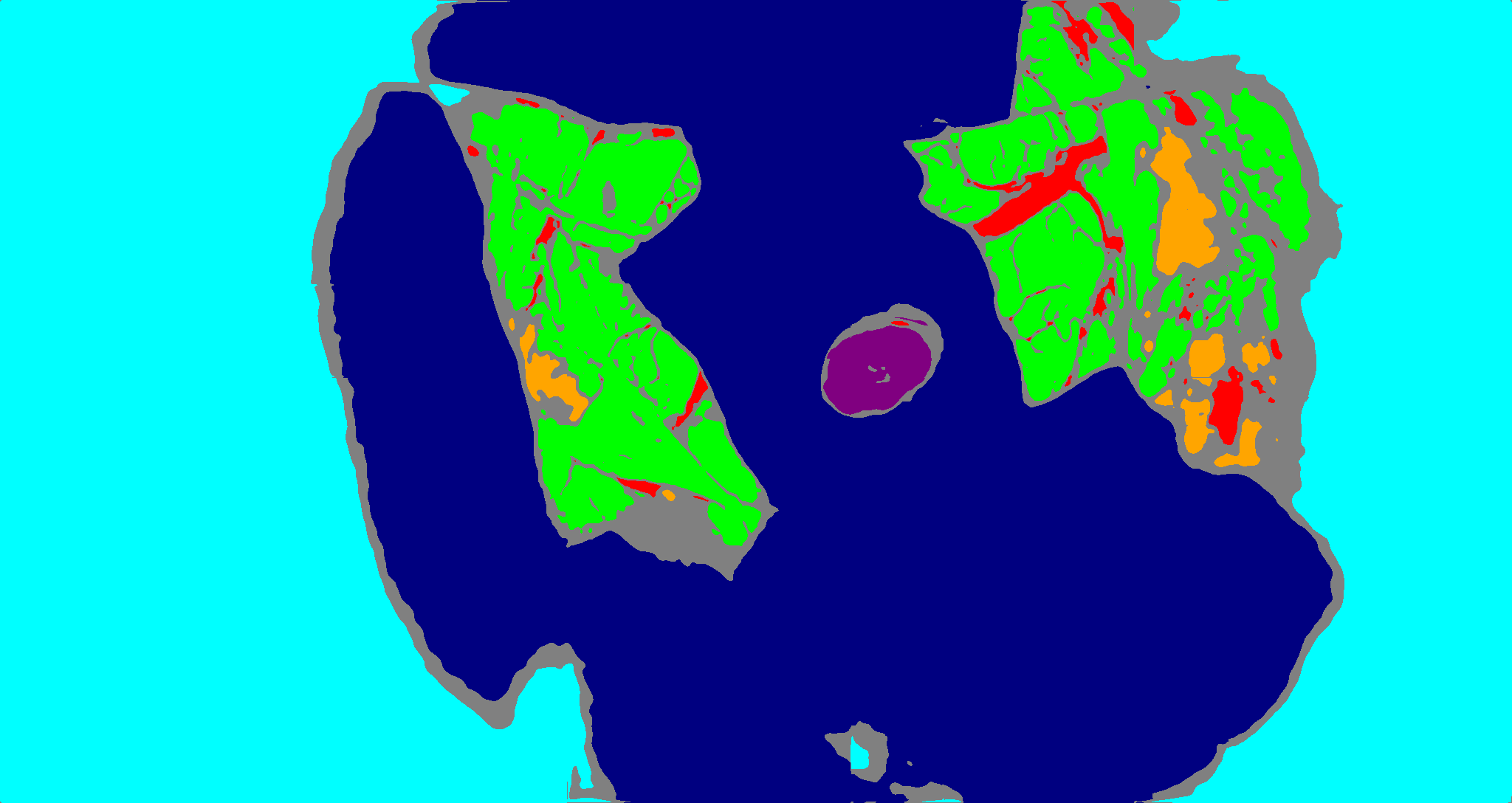}
& \includegraphics[height=1.32cm,valign=t]{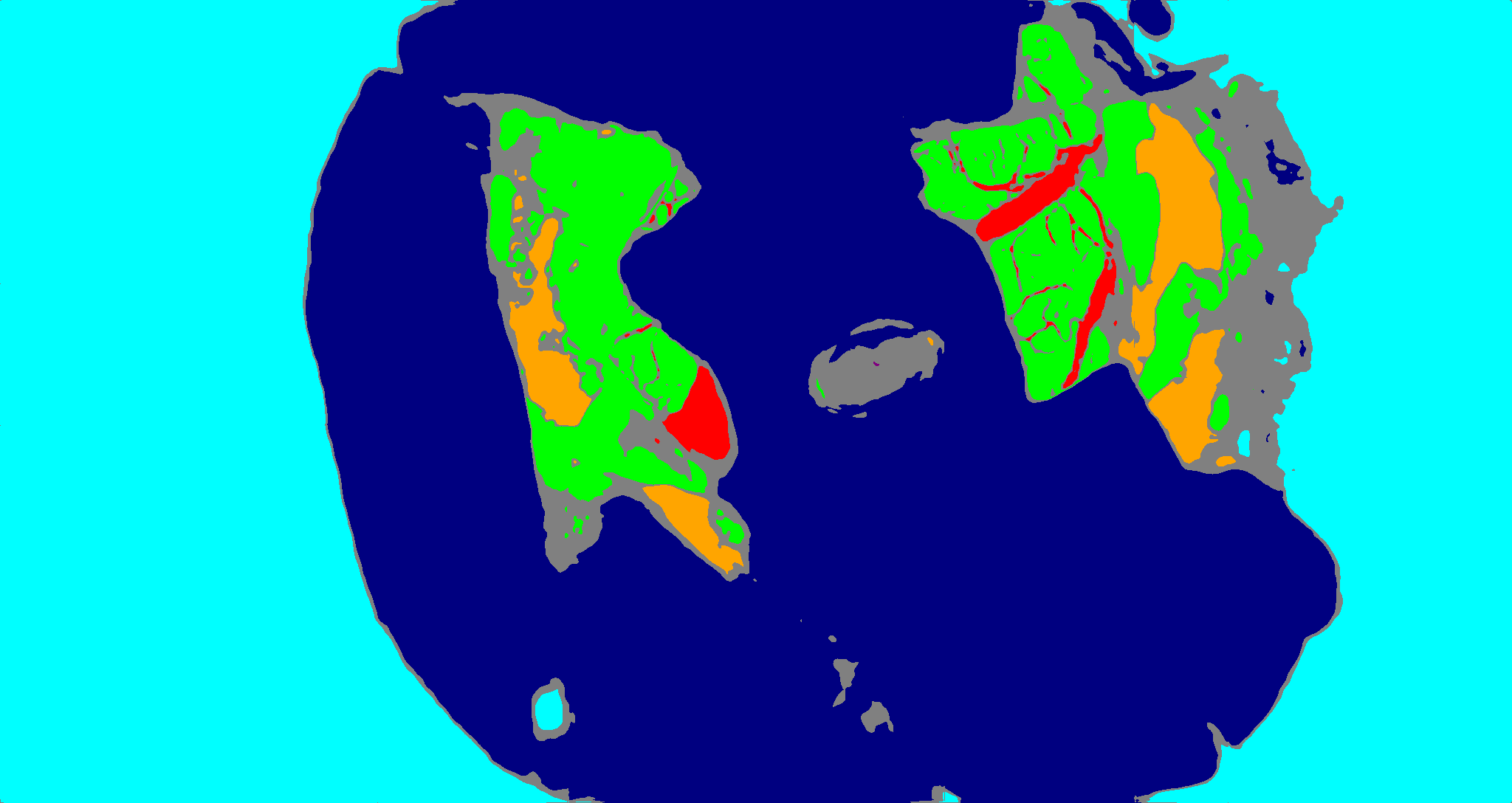}
& \includegraphics[height=1.32cm,valign=t]{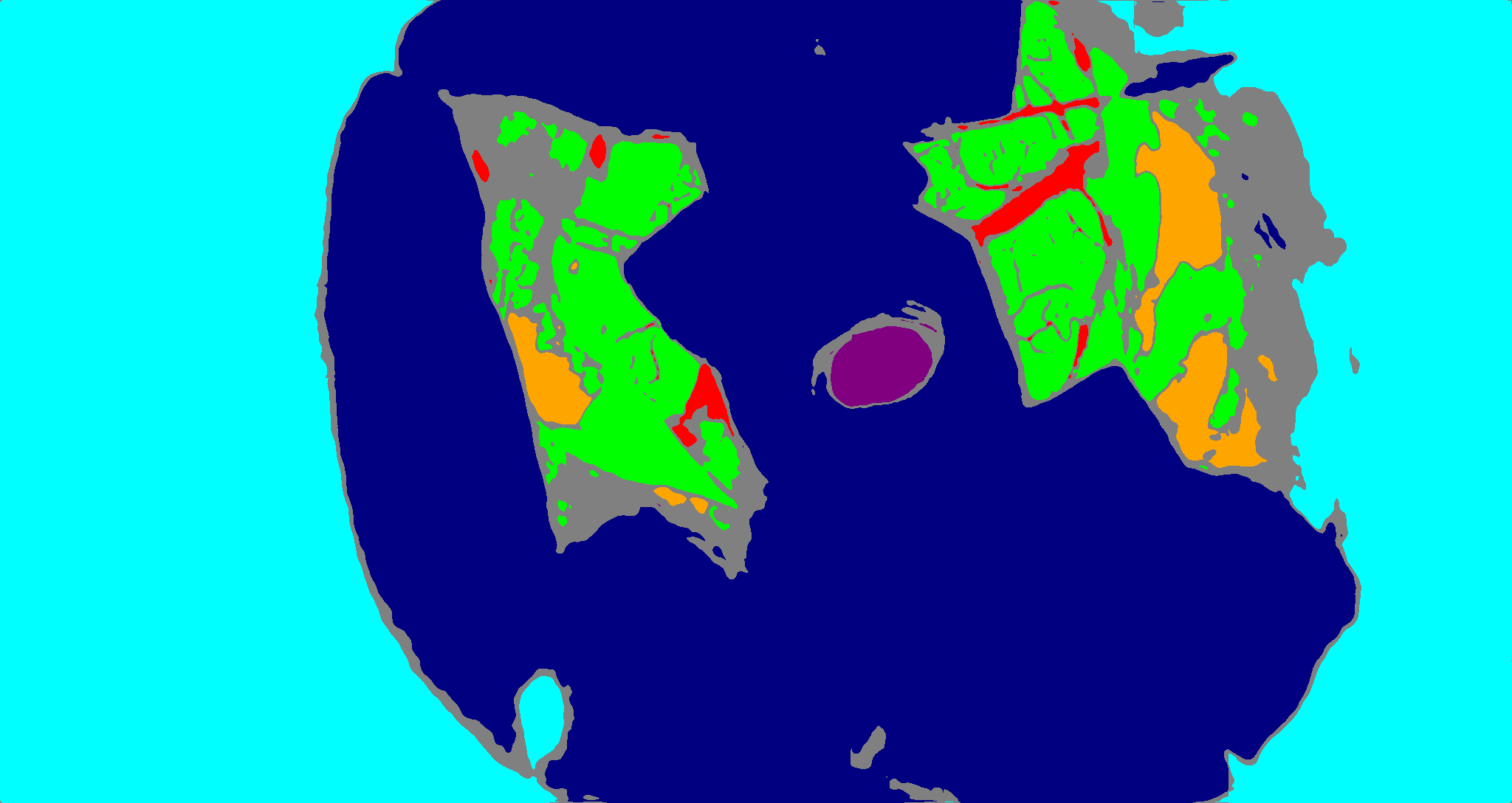}\\
& \includegraphics[height=1.32cm,valign=t]{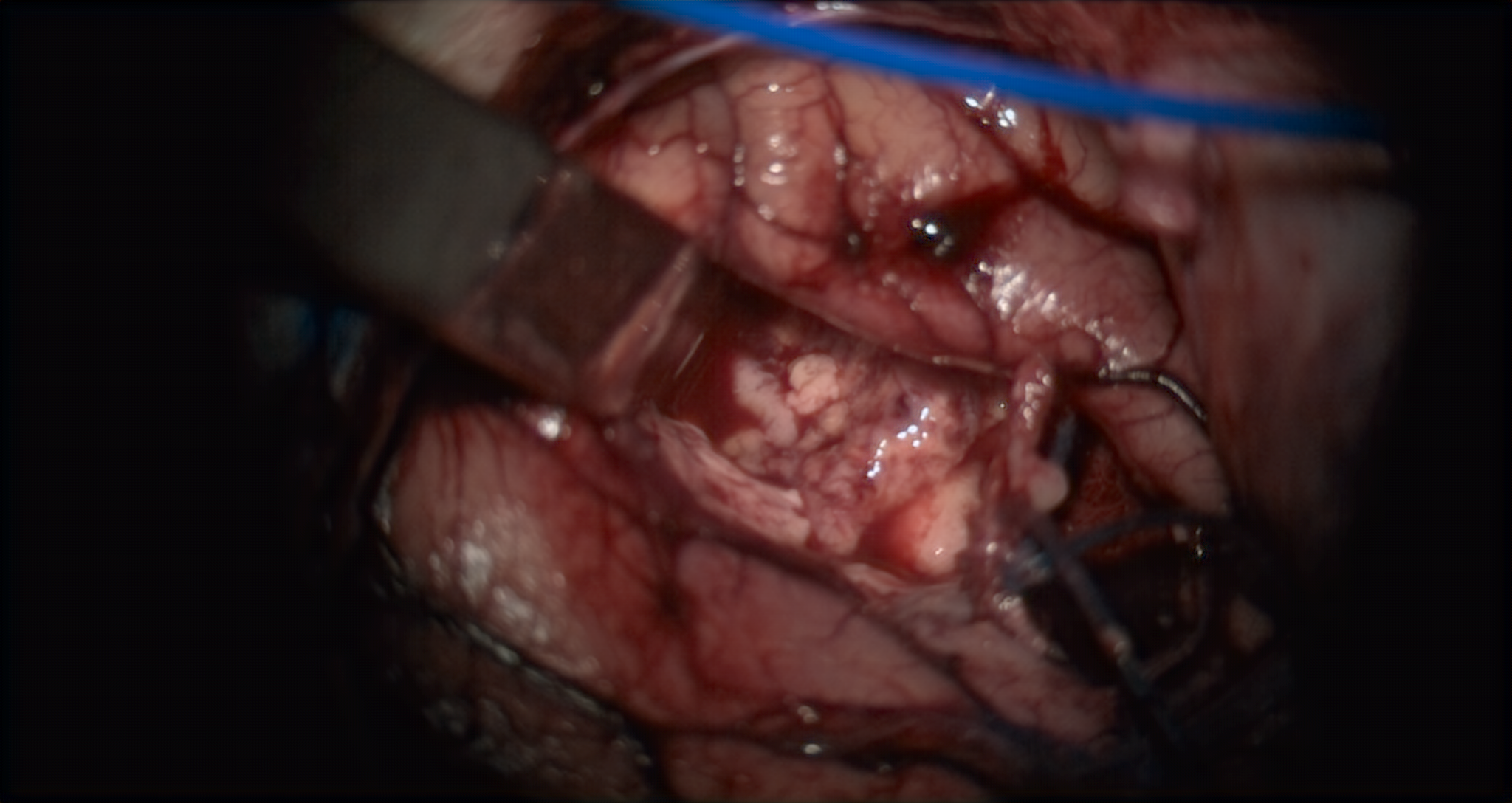}
& \includegraphics[height=1.32cm,valign=t]{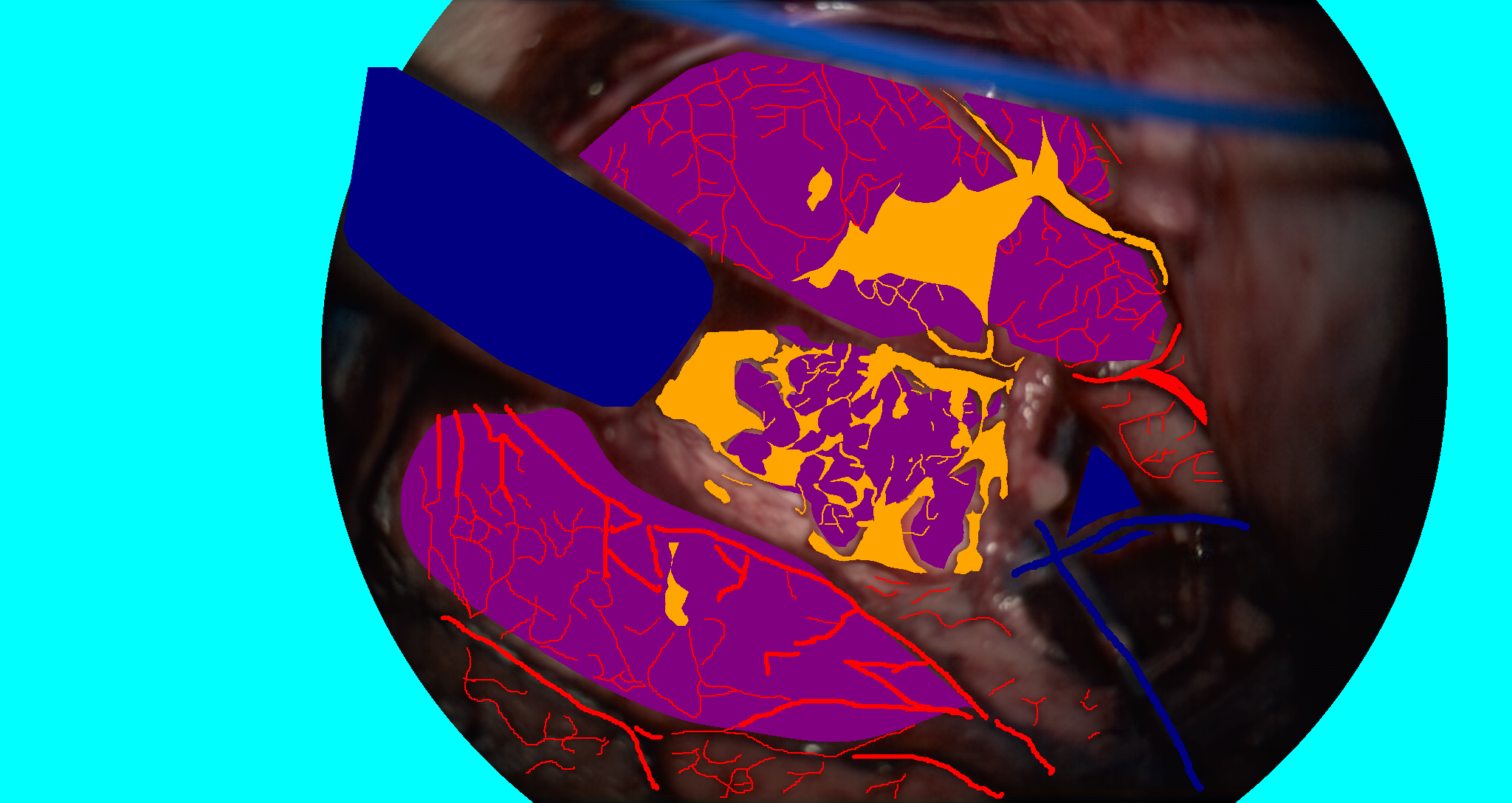}
& \includegraphics[height=1.32cm,valign=t]{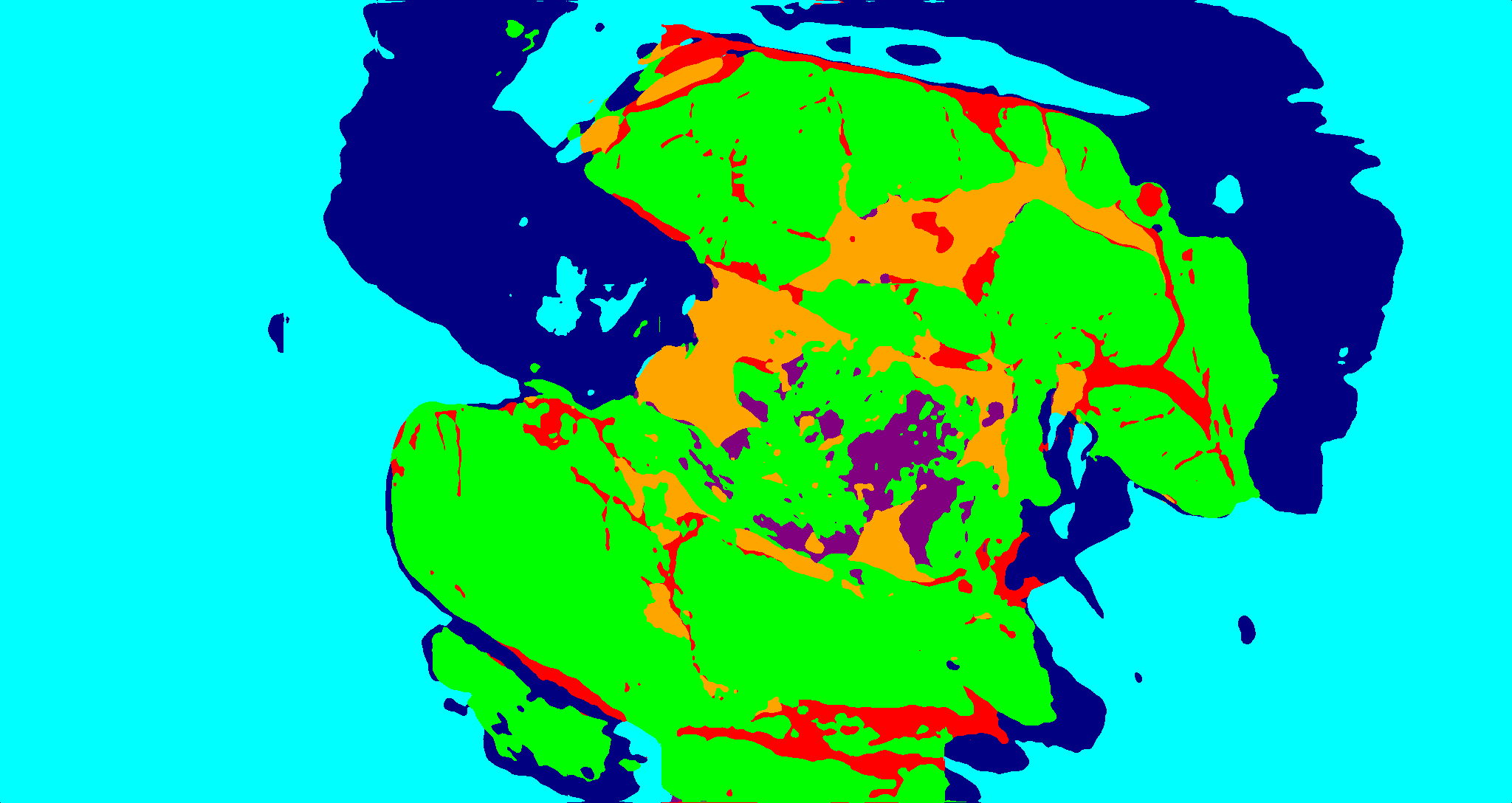}
& \includegraphics[height=1.32cm,valign=t]{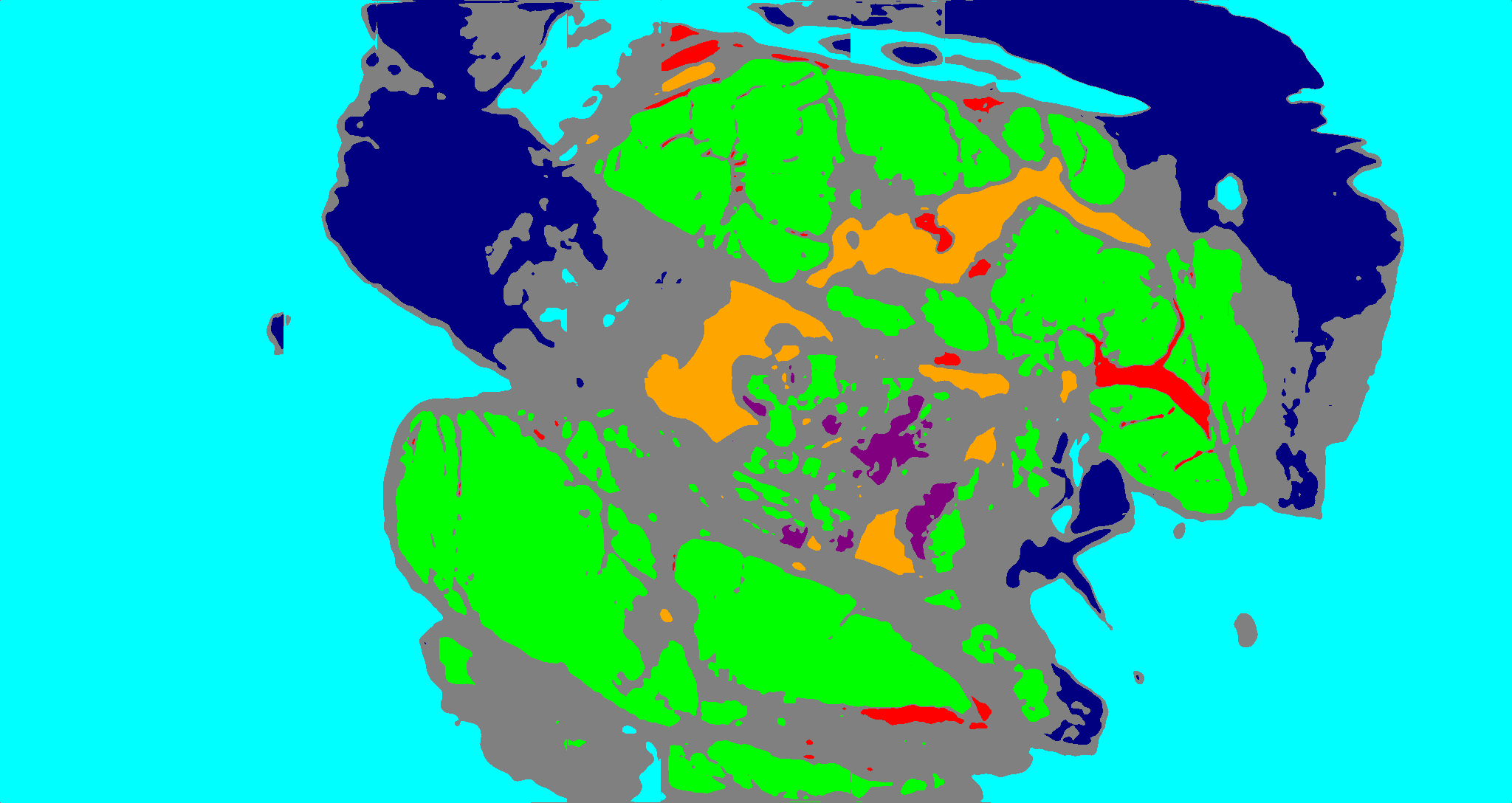}
& \includegraphics[height=1.32cm,valign=t]{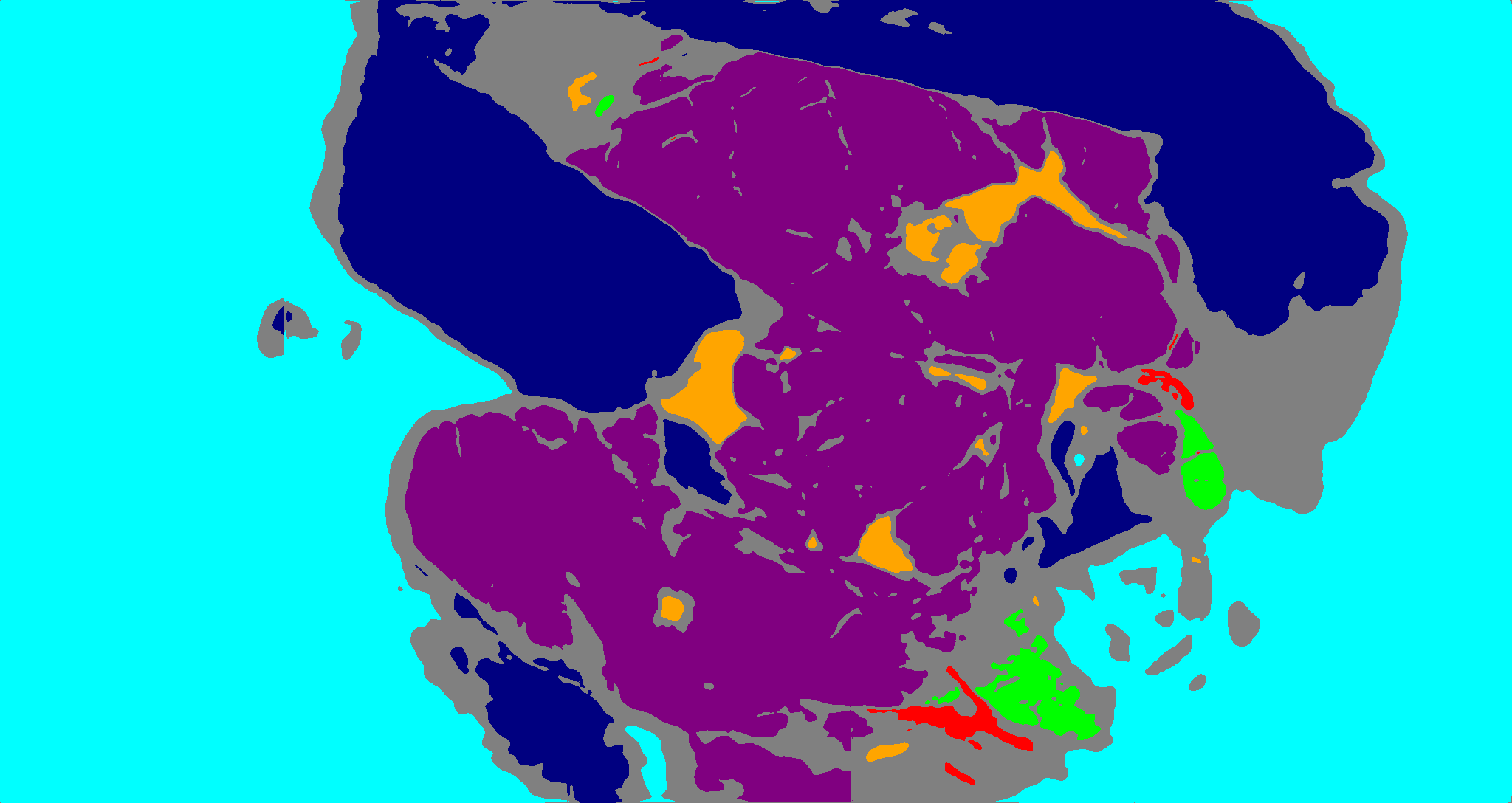}
& \includegraphics[height=1.32cm,valign=t]{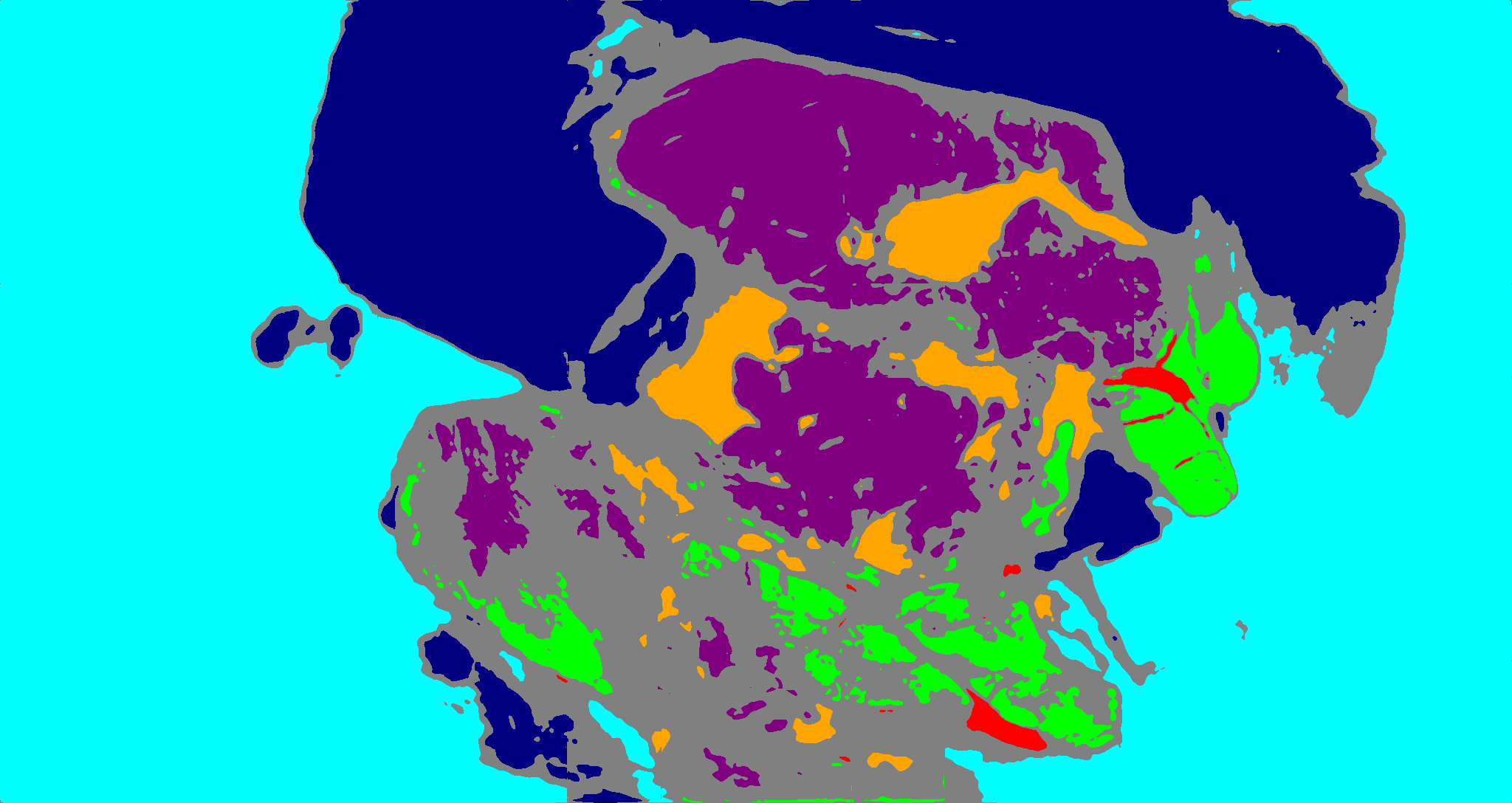}
& \includegraphics[height=1.32cm,valign=t]{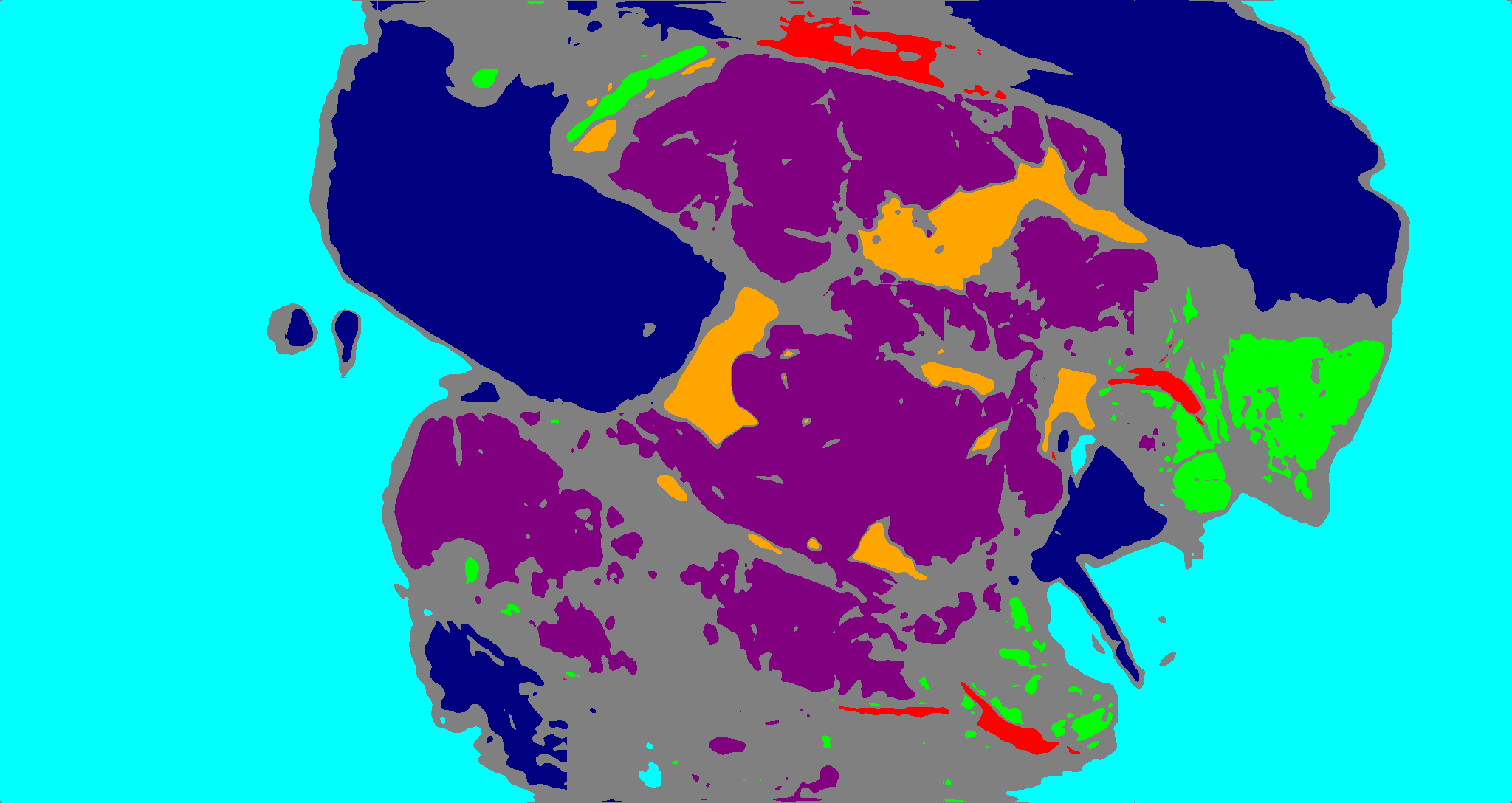}\\
& \multicolumn{7}{c}{\includegraphics[width=0.98\linewidth]{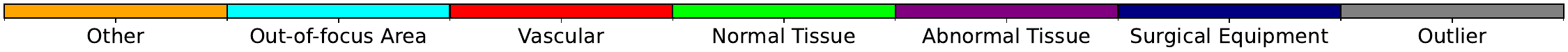}}
\end{tabular}
\caption{Qualitative result on top-level classes. We show the result of same image using different methods at confidence threshold $\tau_m$. Baseline results at $\tau_0=0$ are added to represent result without outlier detection.}
\label{fig:qualitative_results_all_classes}
\end{figure}

\subsubsection{WBP}
\figref{fig:qualitative_result_bp} qualitatively compares the baseline model trained with $\loss_{seg}$ (CE + Dice) against our Wasserstein compound loss $\loss_{wass+seg}$ on the AOMIC dataset. Each column displays the predicted mask at an increasingly fine level of the label hierarchy using the colour scheme and dendrogram from \figref{fig:hierarchy_mindboggle}. The white arrow marks the challenging class ``non-WM-hypointensities,'' which the baseline fails to detect but our method segments correctly across all hierarchical levels.

\subsubsection{HSI}
\figref{fig:qualitative_results_all_classes} presents qualitative results comparing different loss functions on some challenging cases. Although all models have the capacity to differentiate leaf node classes, for simplicity, we visualise predictions at the top-level nodes. We present results at $\tau_0 = 0$ to illustrate the outcome without background / OOD segmentation for the CE baseline ($\loss_{seg}$). For the Wasserstein-based loss, we include results using ground distance matrices $M_t$ and $M_h$ for comparison. Comparing against $M_{\ell}$ baseline, $\loss_{wass+seg}$ and $\loss_{twce}$ show qualitative results that are more semantically plausible in terms of differentiating normal and abnormal tissues. For other classes such as vascular ones, they also show improved segmentation performance by reducing false positive prediction.

\FloatBarrier

\subsection{Analysis of error types (confusion matrices)}\label{sec:error_analysis}
\begin{figure}[htbp]
\centering
\setlength\tabcolsep{1pt}
\begin{tabular}{ccccc}
& $\loss_{seg}$ WBP & $\loss_{wass+seg}$ WBP
& $\loss_{seg}$ HSI & $\loss_{wass+seg}$ HSI \\
\rotatebox[origin=c]{90}{True}
& \includegraphics[width=.24\linewidth,valign=m]{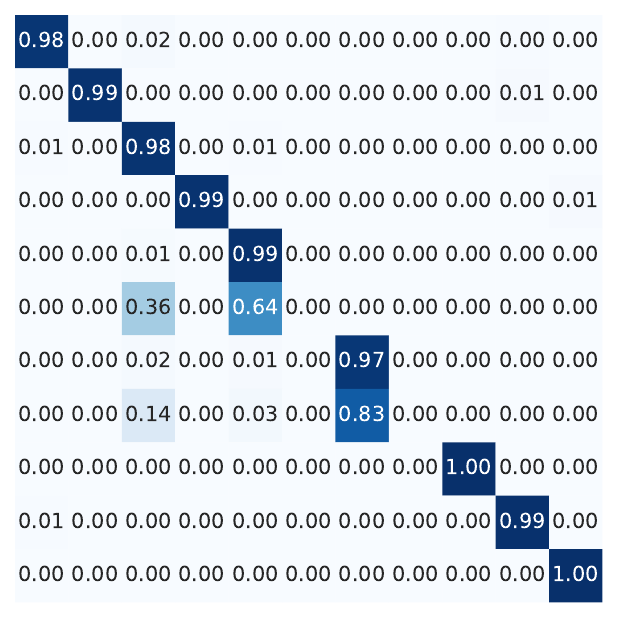}
& \includegraphics[width=.24\linewidth,valign=m]{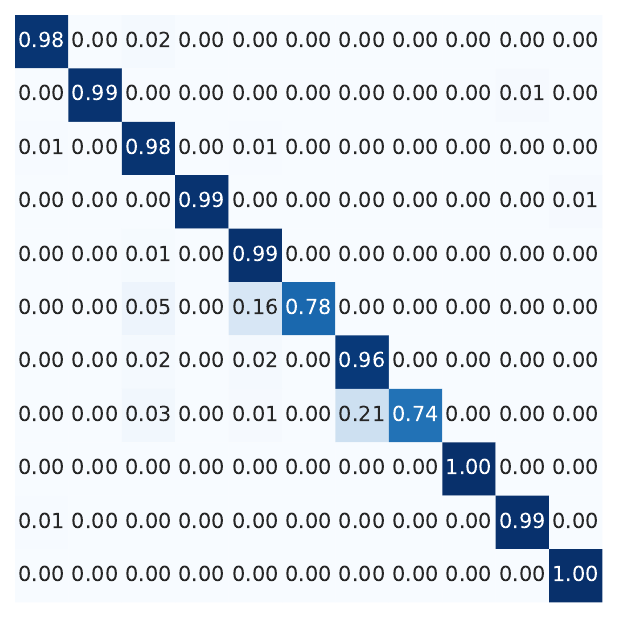}
& \includegraphics[width=.24\linewidth,valign=m]{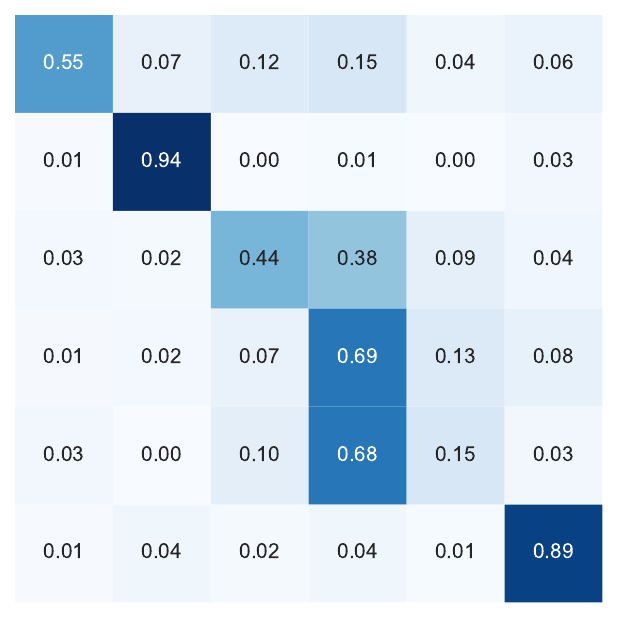}
& \includegraphics[width=.24\linewidth,valign=m]{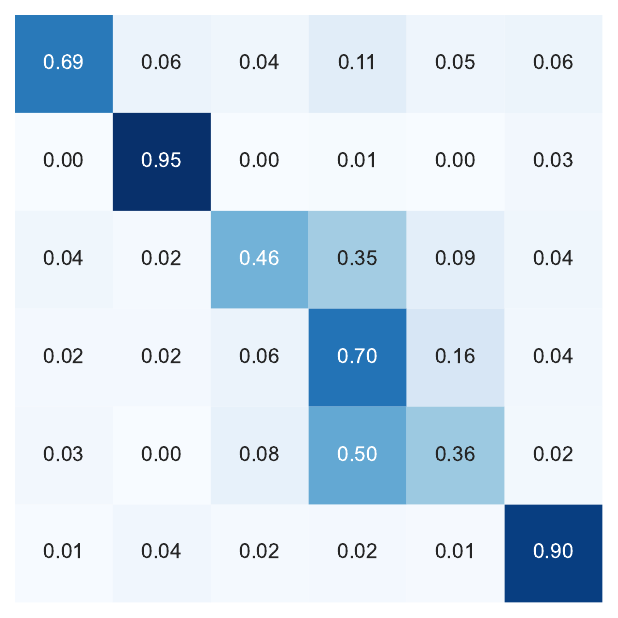}\\
& \multicolumn{4}{c}{Predicted} \\
\end{tabular}
\caption{Confusion matrices for the WBP and HSI tasks. For WBP, the evaluation is on 10 small classes of the AOMIC dataset. Class names from top-left to bottom-right: Left-Inf-Lat-Vent, Left-vessel, Left-choroid-plexus, Right-vessel, Right-choroid-plexus, 5th-Ventricle, WM-hypointensities, non-WM-hypointensities, Optic-Chiasm, ctx-lh-unknown, ctx-rh-unknown. For HSI, the evaluation is on top-level nodes. Class names from top-left to bottom-right: Other, Out-of-focus Area, Vascular, Normal Tissue, Abnormal Tissue and Surgical Equipment.}
\label{fig:confusion_matrix}
\end{figure}

Evaluation relying solely on overall segmentation performance is insufficient to capture how effectively a model infers relationships among different tissue types. It neglects the possibility that the model may choose incorrect but semantically meaningful labels. To explore these aspects, we plot a multi-class confusion matrix for the top-level nodes $p^{\dagger,K-1}$, as shown in \figref{fig:confusion_matrix}. Because class distributions vary across folds, we average the results of each cross-validation fold to ensure a fair comparison.

For CE baseline on HSI task, the model struggles to distinguish normal from abnormal tissues, which constitute a semantically important distinction. By contrast, the tree-semantic losses ($\loss_{wass+seg}$) exhibit a more meaningful confusion pattern between normal and abnormal tissues. Similarly for WBP task, baseline model struggle to differentiate between hypointensity and normal regions. These findings suggest that the proposed method successfully exploits hierarchical relationships within the label space.

\FloatBarrier

\subsection{Hyperparameter tuning result}
\begin{figure}[tbh!]
    \centering
    \setlength\tabcolsep{1pt}
    \includegraphics[width=0.4\linewidth,valign=m]{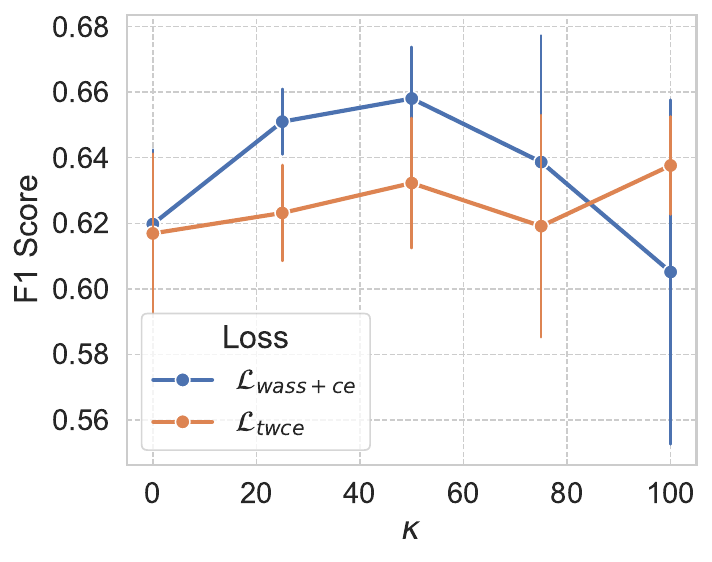}
    \caption{\revdel{Sensitivity analysis result by changing scaling parameter $\kappa$ on HSI dataset. On each trial model is evaluated on a four-fold cross-validation.}\revmod{Sensitivity analysis for the HSI task. The horizontal axis shows the hierarchy-distance scaling parameter $\kappa$ used in $M_h$, and the vertical axis shows the mean F1 score over four-fold cross-validation. Each point corresponds to one complete four-fold trial at the specified $\kappa$ value.}\robustrevref{1}{6}\label{fig:sensitivity_analysis_hsi}}
\end{figure}

Our method relies heavily on the distance matrix $M$, which encodes prior knowledge of inter-class relationships. In practice, exhaustively exploring all possible configurations of $M$ can be computationally expensive. To address this, we adopt the previous $M_h$ configuration, where the edge weight of a parent node is set to be $\kappa$ times larger than that of its child node. Under this $M_h$ configuration, optimizing $M$ reduces to finding the optimal scaling parameter $\kappa$.

To assess the influence of the hyperparameter on model performance, we run multiple trials by tuning the scaling parameter $\kappa$. For each trial, a complete four-fold cross-validation was performed. As shown in \figref{fig:sensitivity_analysis_hsi}, the model performance varies with increasing values of $\kappa$ on the brain HSI dataset. We observed that model performance improved over the baseline by increasing $\kappa$. For $\loss_{wass+ce}$, the model achieved its peak performance at $\kappa = 50$ with best result $0.66$ mean F1 score. Beyond which both the average performance and its consistency across folds decreases.

\FloatBarrier

\section{Conclusion and discussion}
We propose two semantically driven loss functions applicable for both sparse and dense supervised segmentation tasks, relying on a tree-structured label space defined by domain experts. Both the Wasserstein distance based segmentation loss and the tree-weighted semantic segmentation loss leverage prior knowledge of inter-class relationships. The former captures these relationships through a distance matrix in label space, while the latter extends the standard CE loss to incorporate weighted probabilities aggregated at each node in the tree. Additionally, we integrate these loss functions into a sparse positive-only learning framework for segmentation, which enables pixel-level background segmentation through an OOD detection approach.

\revnew{Our experiments were designed to evaluate the proposed losses under strong established segmentation backbones rather than to benchmark alternative architectures. In particular, the WBP experiments use nnU-Net as a widely adopted reference framework for 3D medical image segmentation \citep{isenseeNnUNetSelfconfiguringMethod2021,isensee2024nnunetrevisited}, while the HSI experiments use the established U-Net-based sparse positive-only segmentation pipeline from prior work \citep{wang2024oodsegoutofdistributiondetectionimage, wangTreebasedSemanticLosses2025}.}\revref{2}{2}

Regarding the optimal weighting of hierarchical levels, our experiments on four distance matrices reveal that top-level weights exert the greatest influence on performance when the evaluation is conducted at the corresponding level. Furthermore, we found that the hierarchical weighting scheme further enhances performance, \revdel{achieving state-of-the-art results on the dataset for both top-level and leaf node labels.}\revmod{improving over the evaluated baselines for both top-level and leaf node labels.}\revref{2}{3}\secondrevref{1}{4} Moreover, error analysis and qualitative evaluations demonstrate that these approaches offer improved tissue differentiation compared with standard baselines.

\section*{Conflict of Interest Statement}
TV and JS are co-founders and shareholders of Hypervision Surgical Ltd, London, UK. The authors have no other relevant interests to declare.

\section*{Funding}
This project received funding by the National Institute for Health and Care Research (NIHR) under its Invention for Innovation (i4i) Programme [NIHR202114]. The views expressed are those of the author(s) and not necessarily those of the NIHR or the Department of Health and Social Care. This work was supported by core funding from the Wellcome/EPSRC [WT203148/Z/16/Z; NS/A000049/1]. OM is funded by the EPSRC DTP [EP/T517963/1].

\section*{Acknowledgments}
For the purpose of open access, the authors have applied a CC BY public copyright license to any Author Accepted Manuscript version arising from this submission.

\bibliographystyle{Frontiers-Harvard}
\bibliography{ref}

\newpage
\onecolumn
\appendix
\section{Appendix}

\setcounter{figure}{0}
\renewcommand{\thefigure}{A\arabic{figure}}
\setcounter{table}{0}
\renewcommand{\thetable}{A\arabic{table}}

\begingroup
\centering
\scriptsize
\setlength{\tabcolsep}{2pt}
\renewcommand{\arraystretch}{0.85}
\begin{longtable}{llllrrrrr}
\caption{\revnew{Paired-sample WBP statistical tests on matched test subjects. 
$\Delta$: proposed method minus comparator.
Raw p-values come from two-sided paired t-tests; adjusted p-values use Holm-Bonferroni correction across all predefined WBP headline comparisons. The Raw $p$ and Holm columns report significance markers rather than numeric p-values.
Significance markers use raw p-values in the Raw $p$ column and Holm-adjusted p-values in the Holm column: * $p < 0.05$; ns, no significance.
}\robustrevref{1}{2}\label{tab:wbp_paired_sample_tests}}\\
\toprule
Comparison & Train & Test & Metric & $n$ & Mean $\Delta$ & Median $\Delta$ & Raw $p$ & Holm sig. \\
\midrule
\endfirsthead
\toprule
Comparison & Train & Test & Metric & $n$ & Mean $\Delta$ & Median $\Delta$ & Raw $p$ & Holm sig. \\
\midrule
\endhead
\input{table/wbp_paired_sample_tests.tex}
\end{longtable}
\vspace{0.5em}
\parbox{\linewidth}{\footnotesize \revnew{$\Delta$ is proposed method minus comparator. Significance markers use raw p-values in the Raw $p$ column and Holm-adjusted p-values in the Holm column: * $p < 0.05$; ns, no significance.}\revref{1}{2}}
\endgroup


\begin{figure}[tbh!]
\centering
\includegraphics[width=0.82\linewidth]{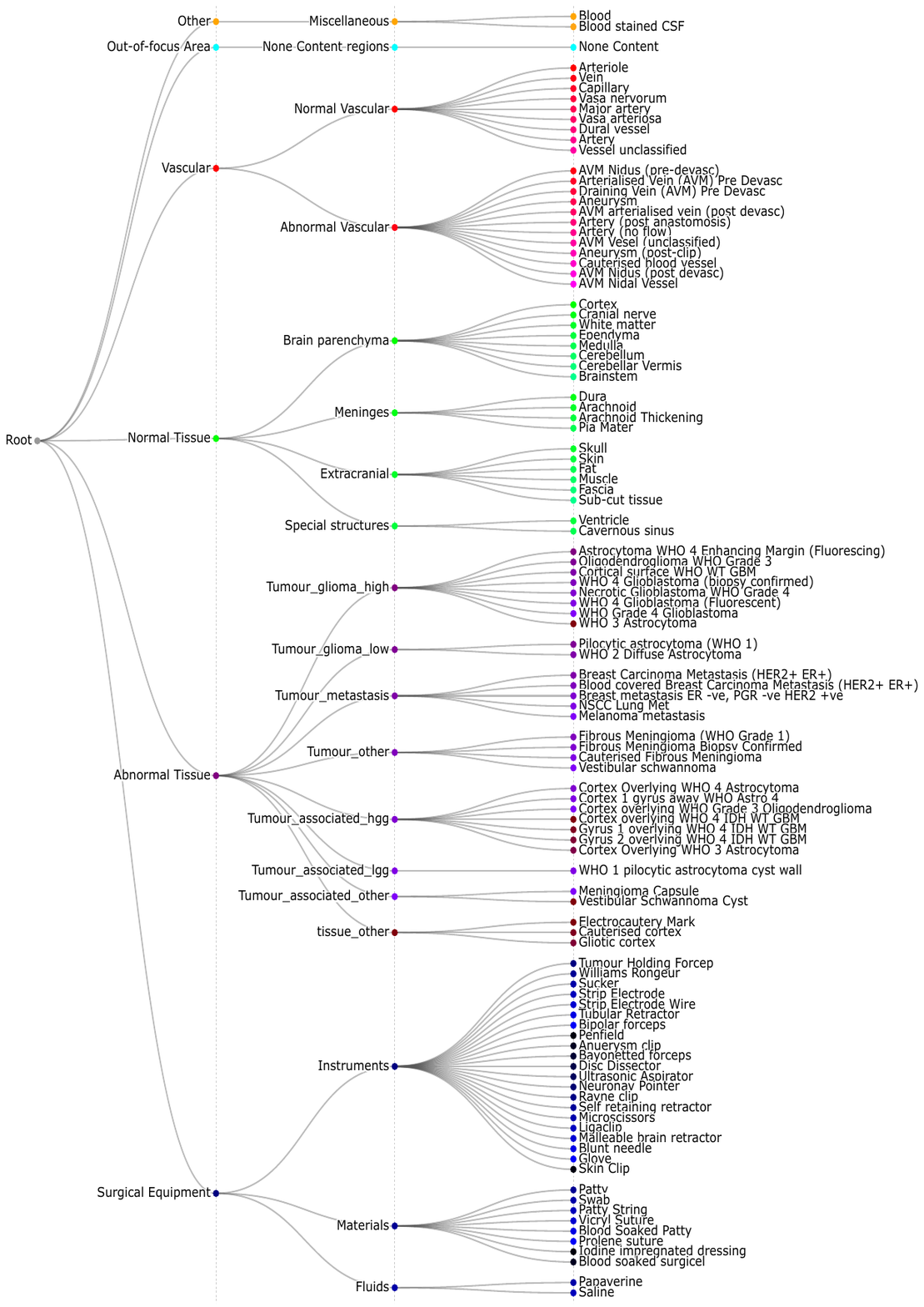}
\caption{Full tree-based label hierarchy of the surgical HSI dataset. From left to right, the hierarchy progresses from coarse object categories to specific classes. The colour coding matches the ground-truth mask at each level.\label{fig:hierarchy_phasetwo}}
\end{figure}

\end{document}

%% file: revision_commands.tex
\ifdefined\showrevision

\usepackage{xcolor}
\usepackage[normalem]{ulem}
\usepackage{marginnote}
\usepackage{xspace}

\newcommand{\revref}[2]{%
\marginnote{$R_{#1}C_{#2}$}
}
\newcommand{\secondrevref}[2]{%
\marginnote{$R_{#1}C_{#2}$}[0.3cm]
}
\DeclareRobustCommand{\robustrevref}[2]{%
\revref{#1}{#2}
}

\newcommand{\revmod}[1]{%
{\color{blue}#1}\xspace
}

\newcommand{\revnew}[1]{%
{\color{orange}#1}\xspace
}

\newcommand{\revdel}[1]{%
{\color{darkgray}\sout{#1}}\xspace
}

\else

\usepackage{xspace}

\newcommand{\marginnote}[1]{\ignorespaces}
\newcommand{\revref}[2]{\ignorespaces}
\newcommand{\secondrevref}[2]{\ignorespaces}
\newcommand{\robustrevref}[2]{\ignorespaces}
\newcommand{\revmod}[1]{#1\xspace}
\newcommand{\revnew}[1]{#1\xspace}
\newcommand{\revdel}[1]{\ignorespaces}

\fi

%% file: figures/tree_matrix_example.tex
\begin{center}
\centering
\revnew{
\begin{tabular}{@{}>{\centering\arraybackslash}m{0.40\linewidth}@{\hspace{0.04\linewidth}}>{\centering\arraybackslash}m{0.54\linewidth}@{}}
\begin{adjustbox}{width=0.78\linewidth, valign=c}
\includegraphics{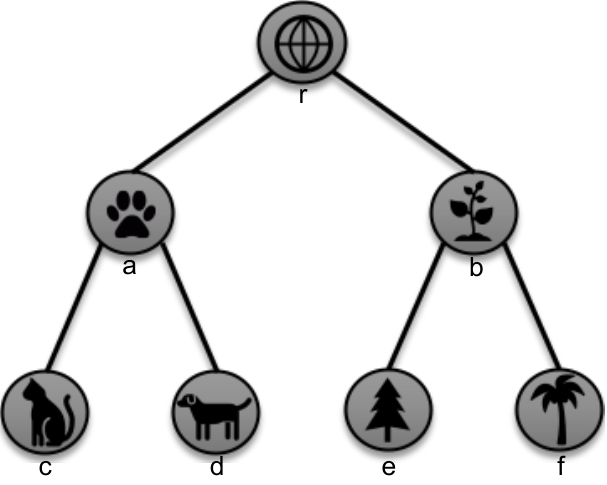}
\end{adjustbox}
&
\begin{adjustbox}{max width=\linewidth, valign=c}
{$\displaystyle
A =
\begin{pmatrix}
0 & 1 & 1 & 0 & 0 & 0 & 0\\
0 & 0 & 0 & 1 & 1 & 0 & 0\\
0 & 0 & 0 & 0 & 0 & 1 & 1\\
0 & 0 & 0 & 0 & 0 & 0 & 0\\
0 & 0 & 0 & 0 & 0 & 0 & 0\\
0 & 0 & 0 & 0 & 0 & 0 & 0\\
0 & 0 & 0 & 0 & 0 & 0 & 0
\end{pmatrix}.
$}
\end{adjustbox}
\end{tabular}
}
\captionsetup{hypcap=false}
\captionof{figure}{\revnew{Example label tree and its adjacency matrix. Each directed edge connects a child node to its parent, so $A_{u,v}=1$ denotes that node $u$ is the parent of node $v$.}\robustrevref{1}{7}}
\label{fig:tree_matrix_example}
\end{center}

%% file: table/wbp_results_tabular.tex
\begin{tabular}{c c cccc cccc cccc cccc}
\toprule
& &
\multicolumn{4}{c}{$\boldsymbol{\Dice}\uparrow$} &
\multicolumn{4}{c}{$\boldsymbol{\NSD}\uparrow$} &
\multicolumn{4}{c}{$\boldsymbol{\SDice}\uparrow$} &
\multicolumn{4}{c}{$\boldsymbol{\SNSD}\uparrow$}\\
\cmidrule(lr){3-6}
\cmidrule(lr){7-10}
\cmidrule(lr){11-14}
\cmidrule(lr){15-18}
\textbf{Train}
& \textbf{Loss} & MB42 & AOMIC & IXI & Avg. & MB42 & AOMIC & IXI & Avg. & MB42 & AOMIC & IXI & Avg. & MB42 & AOMIC & IXI & Avg.\\
\midrule
\multirow{4}{*}{MB59} & $\loss_{seg}$ & 79.5 & 81.1 & 78.9 & 79.8 & 93.3 & 95.2 & 94.3 & 94.3 & 58.2 & 58.5 & 56.9 & 57.9 & 77.0 & 77.7 & 77.6 & 77.4 \\
& \revnew{$\loss_{gwdl}$} & \revnew{79.4} & \revnew{80.7} & \revnew{78.4} & \revnew{79.5} & \revnew{93.1} & \revnew{95.0} & \revnew{94.2} & \revnew{94.1} & \revnew{57.5} & \revnew{57.1} & \revnew{54.2} & \revnew{56.3} & \revnew{75.2} & \revnew{76.0} & \revnew{76.1} & \revnew{75.8} \\
& $\loss_{twce+seg}$ & 79.3\revnew{\textsuperscript{*}} & 81.3\revnew{\textsuperscript{*}} & 79.0\revnew{\textsuperscript{*}} & 79.9 & 93.0\revnew{\textsuperscript{*}} & 95.2 & 94.2 & 94.2 & 58.2 & 58.5 & 56.8 & 57.8 & 76.9 & 77.9 & 77.5 & 77.4 \\
& $\loss_{wass+seg}$ & \textbf{79.8}\revnew{\textsuperscript{*}} & \textbf{81.6}\revnew{\textsuperscript{*}} & \textbf{79.2}\revnew{\textsuperscript{*}} & \textbf{80.2} & \textbf{94.2}\revnew{\textsuperscript{*}} & \textbf{96.1}\revnew{\textsuperscript{*}} & \textbf{95.2}\revnew{\textsuperscript{*}} & \textbf{95.2} & \textbf{60.7}\revnew{\textsuperscript{*}} & \textbf{62.7}\revnew{\textsuperscript{*}} & \textbf{59.4}\revnew{\textsuperscript{*}} & \textbf{60.9} & \textbf{86.4}\revnew{\textsuperscript{*}} & \textbf{87.3}\revnew{\textsuperscript{*}} & \textbf{87.2}\revnew{\textsuperscript{*}} & \textbf{87.0} \\
\midrule
\multirow{4}{*}{AOMIC} & $\loss_{seg}$ & \textbf{74.8} & 88.3 & 83.8 & 82.3 & 91.6 & 97.0 & 95.6 & 94.7 & 52.8 & 68.0 & 63.1 & 61.3 & 74.2 & 77.8 & 77.5 & 76.5 \\
& \revnew{$\loss_{gwdl}$} & \revnew{74.6} & \revnew{88.2} & \revnew{83.2} & \revnew{82.0} & \revnew{91.3} & \revnew{97.0} & \revnew{95.4} & \revnew{94.6} & \revnew{52.3} & \revnew{66.9} & \revnew{59.4} & \revnew{59.5} & \revnew{73.1} & \revnew{77.7} & \revnew{75.2} & \revnew{75.4} \\
& $\loss_{twce+seg}$ & 74.6 & 88.5 & 83.9 & 82.3 & 91.5 & 97.5\revnew{\textsuperscript{*}} & 95.9\revnew{\textsuperscript{*}} & 95.0 & 53.1 & 70.8\revnew{\textsuperscript{*}} & 66.2\revnew{\textsuperscript{*}} & 63.4 & 75.5 & 84.4\revnew{\textsuperscript{*}} & 84.1\revnew{\textsuperscript{*}} & 81.4 \\
& $\loss_{wass+seg}$ & \textbf{74.8} & \textbf{89.2}\revnew{\textsuperscript{*}} & \textbf{84.0} & \textbf{82.7} & \textbf{92.4}\revnew{\textsuperscript{*}} & \textbf{98.5}\revnew{\textsuperscript{*}} & \textbf{96.8}\revnew{\textsuperscript{*}} & \textbf{95.9} & \textbf{55.2}\revnew{\textsuperscript{*}} & \textbf{77.0}\revnew{\textsuperscript{*}} & \textbf{70.5}\revnew{\textsuperscript{*}} & \textbf{67.6} & \textbf{84.9}\revnew{\textsuperscript{*}} & \textbf{94.7}\revnew{\textsuperscript{*}} & \textbf{93.2}\revnew{\textsuperscript{*}} & \textbf{90.9} \\
\midrule
\multirow{4}{*}{IXI} & $\loss_{seg}$ & 74.6 & 86.2 & 90.0 & 83.6 & 91.9 & 97.5 & 98.1 & 95.8 & 52.4 & 68.8 & 74.2 & 65.1 & 76.5 & 85.3 & 87.3 & 83.0 \\
& \revnew{$\loss_{gwdl}$} & \revnew{74.5} & \revnew{85.7} & \revnew{89.3} & \revnew{83.1} & \revnew{91.6} & \revnew{96.7} & \revnew{97.3} & \revnew{95.2} & \revnew{51.8} & \revnew{64.4} & \revnew{67.8} & \revnew{61.3} & \revnew{73.8} & \revnew{77.4} & \revnew{78.9} & \revnew{76.7} \\
& $\loss_{twce+seg}$ & 74.7 & 86.0\revnew{\textsuperscript{*}} & 89.9\revnew{\textsuperscript{*}} & 83.5 & 91.7\revnew{\textsuperscript{*}} & 97.3\revnew{\textsuperscript{*}} & 98.0\revnew{\textsuperscript{*}} & 95.7 & 52.6 & 68.9 & 74.1 & 65.2 & 75.7\revnew{\textsuperscript{*}} & 85.0 & 87.0\revnew{\textsuperscript{*}} & 82.6 \\
& $\loss_{wass+seg}$ & \textbf{75.0}\revnew{\textsuperscript{*}} & \textbf{86.5} & \textbf{90.6}\revnew{\textsuperscript{*}} & \textbf{84.0} & \textbf{92.8}\revnew{\textsuperscript{*}} & \textbf{98.4}\revnew{\textsuperscript{*}} & \textbf{99.0}\revnew{\textsuperscript{*}} & \textbf{96.7} & \textbf{54.7}\revnew{\textsuperscript{*}} & \textbf{74.1}\revnew{\textsuperscript{*}} & \textbf{80.4}\revnew{\textsuperscript{*}} & \textbf{69.8} & \textbf{85.1}\revnew{\textsuperscript{*}} & \textbf{94.7}\revnew{\textsuperscript{*}} & \textbf{97.2}\revnew{\textsuperscript{*}} & \textbf{92.3} \\
\bottomrule
\end{tabular}

%% file: table/wbp_paired_sample_tests.tex
$\loss_{wass+seg}$ vs $\loss_{seg}$ & MB59 & MB42 & Dice & 42 & 0.0020 & 0.0021 & * & * \\
$\loss_{wass+seg}$ vs $\loss_{seg}$ & MB59 & MB42 & NSD & 42 & 0.0055 & 0.0066 & * & * \\
$\loss_{wass+seg}$ vs $\loss_{seg}$ & MB59 & MB42 & Dice\_small & 42 & 0.0132 & 0.0096 & * & * \\
$\loss_{wass+seg}$ vs $\loss_{seg}$ & MB59 & MB42 & NSD\_small & 42 & 0.0563 & 0.0869 & * & * \\
$\loss_{wass+seg}$ vs $\loss_{seg}$ & MB59 & AOMIC & Dice & 46 & 0.0035 & 0.0029 & * & * \\
$\loss_{wass+seg}$ vs $\loss_{seg}$ & MB59 & AOMIC & NSD & 46 & 0.0065 & 0.0087 & * & * \\
$\loss_{wass+seg}$ vs $\loss_{seg}$ & MB59 & AOMIC & Dice\_small & 46 & 0.0283 & 0.0298 & * & * \\
$\loss_{wass+seg}$ vs $\loss_{seg}$ & MB59 & AOMIC & NSD\_small & 46 & 0.0670 & 0.0975 & * & * \\
$\loss_{wass+seg}$ vs $\loss_{seg}$ & MB59 & IXI & Dice & 117 & 0.0032 & 0.0032 & * & * \\
$\loss_{wass+seg}$ vs $\loss_{seg}$ & MB59 & IXI & NSD & 117 & 0.0071 & 0.0082 & * & * \\
$\loss_{wass+seg}$ vs $\loss_{seg}$ & MB59 & IXI & Dice\_small & 117 & 0.0181 & 0.0150 & * & * \\
$\loss_{wass+seg}$ vs $\loss_{seg}$ & MB59 & IXI & NSD\_small & 117 & 0.0721 & 0.0942 & * & * \\
$\loss_{wass+seg}$ vs $\loss_{seg}$ & AOMIC & MB42 & Dice & 42 & -0.0002 & -0.0004 & ns & ns \\
$\loss_{wass+seg}$ vs $\loss_{seg}$ & AOMIC & MB42 & NSD & 42 & 0.0045 & 0.0057 & * & * \\
$\loss_{wass+seg}$ vs $\loss_{seg}$ & AOMIC & MB42 & Dice\_small & 42 & 0.0150 & 0.0043 & * & * \\
$\loss_{wass+seg}$ vs $\loss_{seg}$ & AOMIC & MB42 & NSD\_small & 42 & 0.0707 & 0.0905 & * & * \\
$\loss_{wass+seg}$ vs $\loss_{seg}$ & AOMIC & AOMIC & Dice & 46 & 0.0063 & 0.0064 & * & * \\
$\loss_{wass+seg}$ vs $\loss_{seg}$ & AOMIC & AOMIC & NSD & 46 & 0.0127 & 0.0151 & * & * \\
$\loss_{wass+seg}$ vs $\loss_{seg}$ & AOMIC & AOMIC & Dice\_small & 46 & 0.0705 & 0.0784 & * & * \\
$\loss_{wass+seg}$ vs $\loss_{seg}$ & AOMIC & AOMIC & NSD\_small & 46 & 0.1390 & 0.1693 & * & * \\
$\loss_{wass+seg}$ vs $\loss_{seg}$ & AOMIC & IXI & Dice & 117 & 0.0004 & 0.0013 & ns & ns \\
$\loss_{wass+seg}$ vs $\loss_{seg}$ & AOMIC & IXI & NSD & 117 & 0.0096 & 0.0116 & * & * \\
$\loss_{wass+seg}$ vs $\loss_{seg}$ & AOMIC & IXI & Dice\_small & 117 & 0.0600 & 0.0586 & * & * \\
$\loss_{wass+seg}$ vs $\loss_{seg}$ & AOMIC & IXI & NSD\_small & 117 & 0.1355 & 0.1514 & * & * \\
$\loss_{wass+seg}$ vs $\loss_{seg}$ & IXI & MB42 & Dice & 42 & 0.0030 & 0.0026 & * & * \\
$\loss_{wass+seg}$ vs $\loss_{seg}$ & IXI & MB42 & NSD & 42 & 0.0054 & 0.0072 & * & * \\
$\loss_{wass+seg}$ vs $\loss_{seg}$ & IXI & MB42 & Dice\_small & 42 & 0.0148 & 0.0054 & * & * \\
$\loss_{wass+seg}$ vs $\loss_{seg}$ & IXI & MB42 & NSD\_small & 42 & 0.0514 & 0.0801 & * & * \\
$\loss_{wass+seg}$ vs $\loss_{seg}$ & IXI & AOMIC & Dice & 46 & 0.0015 & 0.0016 & * & ns \\
$\loss_{wass+seg}$ vs $\loss_{seg}$ & IXI & AOMIC & NSD & 46 & 0.0055 & 0.0079 & * & * \\
$\loss_{wass+seg}$ vs $\loss_{seg}$ & IXI & AOMIC & Dice\_small & 46 & 0.0353 & 0.0357 & * & * \\
$\loss_{wass+seg}$ vs $\loss_{seg}$ & IXI & AOMIC & NSD\_small & 46 & 0.0643 & 0.0957 & * & * \\
$\loss_{wass+seg}$ vs $\loss_{seg}$ & IXI & IXI & Dice & 117 & 0.0037 & 0.0046 & * & * \\
$\loss_{wass+seg}$ vs $\loss_{seg}$ & IXI & IXI & NSD & 117 & 0.0065 & 0.0087 & * & * \\
$\loss_{wass+seg}$ vs $\loss_{seg}$ & IXI & IXI & Dice\_small & 117 & 0.0454 & 0.0547 & * & * \\
$\loss_{wass+seg}$ vs $\loss_{seg}$ & IXI & IXI & NSD\_small & 117 & 0.0726 & 0.0969 & * & * \\
$\loss_{twce+seg}$ vs $\loss_{seg}$ & MB59 & MB42 & Dice & 42 & -0.0020 & -0.0014 & * & * \\
$\loss_{twce+seg}$ vs $\loss_{seg}$ & MB59 & MB42 & NSD & 42 & -0.0025 & -0.0025 & * & * \\
$\loss_{twce+seg}$ vs $\loss_{seg}$ & MB59 & MB42 & Dice\_small & 42 & -0.0003 & 0.0014 & ns & ns \\
$\loss_{twce+seg}$ vs $\loss_{seg}$ & MB59 & MB42 & NSD\_small & 42 & -0.0008 & -0.0028 & ns & ns \\
$\loss_{twce+seg}$ vs $\loss_{seg}$ & MB59 & AOMIC & Dice & 46 & 0.0019 & 0.0018 & * & * \\
$\loss_{twce+seg}$ vs $\loss_{seg}$ & MB59 & AOMIC & NSD & 46 & 0.0003 & 0.0004 & ns & ns \\
$\loss_{twce+seg}$ vs $\loss_{seg}$ & MB59 & AOMIC & Dice\_small & 46 & 0.0007 & 0.0006 & ns & ns \\
$\loss_{twce+seg}$ vs $\loss_{seg}$ & MB59 & AOMIC & NSD\_small & 46 & 0.0022 & 0.0008 & ns & ns \\
$\loss_{twce+seg}$ vs $\loss_{seg}$ & MB59 & IXI & Dice & 117 & 0.0014 & 0.0016 & * & * \\
$\loss_{twce+seg}$ vs $\loss_{seg}$ & MB59 & IXI & NSD & 117 & -0.0006 & -0.0005 & * & ns \\
$\loss_{twce+seg}$ vs $\loss_{seg}$ & MB59 & IXI & Dice\_small & 117 & -0.0012 & -0.0005 & ns & ns \\
$\loss_{twce+seg}$ vs $\loss_{seg}$ & MB59 & IXI & NSD\_small & 117 & -0.0005 & 0.0010 & ns & ns \\
$\loss_{twce+seg}$ vs $\loss_{seg}$ & AOMIC & MB42 & Dice & 42 & -0.0011 & -0.0009 & * & ns \\
$\loss_{twce+seg}$ vs $\loss_{seg}$ & AOMIC & MB42 & NSD & 42 & -0.0000 & -0.0003 & ns & ns \\
$\loss_{twce+seg}$ vs $\loss_{seg}$ & AOMIC & MB42 & Dice\_small & 42 & -0.0025 & -0.0029 & ns & ns \\
$\loss_{twce+seg}$ vs $\loss_{seg}$ & AOMIC & MB42 & NSD\_small & 42 & 0.0126 & 0.0067 & * & ns \\
$\loss_{twce+seg}$ vs $\loss_{seg}$ & AOMIC & AOMIC & Dice & 46 & 0.0013 & 0.0014 & * & ns \\
$\loss_{twce+seg}$ vs $\loss_{seg}$ & AOMIC & AOMIC & NSD & 46 & 0.0054 & 0.0052 & * & * \\
$\loss_{twce+seg}$ vs $\loss_{seg}$ & AOMIC & AOMIC & Dice\_small & 46 & 0.0270 & 0.0289 & * & * \\
$\loss_{twce+seg}$ vs $\loss_{seg}$ & AOMIC & AOMIC & NSD\_small & 46 & 0.0678 & 0.0656 & * & * \\
$\loss_{twce+seg}$ vs $\loss_{seg}$ & AOMIC & IXI & Dice & 117 & -0.0014 & 0.0001 & ns & ns \\
$\loss_{twce+seg}$ vs $\loss_{seg}$ & AOMIC & IXI & NSD & 117 & 0.0025 & 0.0048 & * & * \\
$\loss_{twce+seg}$ vs $\loss_{seg}$ & AOMIC & IXI & Dice\_small & 117 & 0.0309 & 0.0333 & * & * \\
$\loss_{twce+seg}$ vs $\loss_{seg}$ & AOMIC & IXI & NSD\_small & 117 & 0.0694 & 0.0756 & * & * \\
$\loss_{twce+seg}$ vs $\loss_{seg}$ & IXI & MB42 & Dice & 42 & 0.0005 & -0.0001 & ns & ns \\
$\loss_{twce+seg}$ vs $\loss_{seg}$ & IXI & MB42 & NSD & 42 & -0.0018 & -0.0014 & * & * \\
$\loss_{twce+seg}$ vs $\loss_{seg}$ & IXI & MB42 & Dice\_small & 42 & 0.0021 & 0.0020 & ns & ns \\
$\loss_{twce+seg}$ vs $\loss_{seg}$ & IXI & MB42 & NSD\_small & 42 & -0.0080 & -0.0046 & * & * \\
$\loss_{twce+seg}$ vs $\loss_{seg}$ & IXI & AOMIC & Dice & 46 & -0.0022 & -0.0025 & * & * \\
$\loss_{twce+seg}$ vs $\loss_{seg}$ & IXI & AOMIC & NSD & 46 & -0.0017 & -0.0017 & * & * \\
$\loss_{twce+seg}$ vs $\loss_{seg}$ & IXI & AOMIC & Dice\_small & 46 & 0.0010 & 0.0023 & ns & ns \\
$\loss_{twce+seg}$ vs $\loss_{seg}$ & IXI & AOMIC & NSD\_small & 46 & -0.0033 & -0.0027 & * & ns \\
$\loss_{twce+seg}$ vs $\loss_{seg}$ & IXI & IXI & Dice & 117 & -0.0015 & -0.0017 & * & * \\
$\loss_{twce+seg}$ vs $\loss_{seg}$ & IXI & IXI & NSD & 117 & -0.0011 & -0.0012 & * & * \\
$\loss_{twce+seg}$ vs $\loss_{seg}$ & IXI & IXI & Dice\_small & 117 & -0.0012 & -0.0002 & ns & ns \\
$\loss_{twce+seg}$ vs $\loss_{seg}$ & IXI & IXI & NSD\_small & 117 & -0.0028 & -0.0025 & * & * \\
\bottomrule